\documentclass{tlp}

\usepackage{url}
\usepackage{latexsym}
\usepackage{alltt}
\usepackage{epsfig}
\usepackage{epic}
\usepackage{eepic}
\usepackage{xspace}
\usepackage{amstext,amssymb,amsbsy}
\newif\ifmakebbl

\newcommand{\GR}[2]{\ensuremath{{\Gamma}_{#1}^{#2}}} 
\renewcommand{\GR}[2]{\ensuremath{{\mathit{GR}} (#1,#2)}}
\renewcommand{\GR}[2]{\ensuremath{{\mathit{GR}} (#2,#1)}}

\newcommand{\ext}[1]{\hat{#1}}
\newcommand{\extp}[1]{\overline{#1}}
\renewcommand{\ext}[1]{\lceil{#1}\rceil}
\newcommand{\occurs}{{\it occurs}}
\newcommand{\joins}{{\it joins}}
\newcommand{\query}{{\it query}}
\newcommand{\lowquerycand}{{\it query\_cand}}
\newcommand{\querycand}{{\it iquery\_cand}}
\newcommand{\subquery}{{\it sub\_query}}
\newcommand{\source}{{\it source}}
\newcommand{\dbname}{{\it db\_name}}
\newcommand{\whereRef}{{\it whereRef}}
\newcommand{\subpath}{{\it subpath}}
\newcommand{\whereRefCmp}{{\it whereRefCmp}}
\newcommand{\whereCmp}{{\it whereCmp}}
\newcommand{\consRef}{{\it consRef}}
\newcommand{\hassource}{{\it has\_source}}
\newcommand{\issubquery}{{\it is\_sub\_query}}
\newcommand{\topquery}{{\it top\_query}}
\newcommand{\synonym}{{\it synonym}}
\newcommand{\cov}{{\it covers}}

\newcommand{\PLP}[0]{\textrm{plp}\xspace}
\newcommand{\impact}{\textsl{IMPACT}}

\newcommand{\SI}{{\it Sel}}
\newcommand{\prgm}[1]{\Pi_{\it{#1}}}
\newcommand{\prgmsup}[2]{\Pi_{\it #1}^{\it #2}}
\newcommand{\qs}{{\it query\_source}}
\newcommand{\selects}{{\it selects}}
\newcommand{\constructs}{{\it constructs}}
\newcommand{\charge}{{\it charge}}
\newcommand{\downtime}{{\it avg\_down\_time}}
\newcommand{\lastupd}{{\it last\_update}}
\newcommand{\acc}{{\it accurate}}
\newcommand{\freq}{{\it update\_frequency}}
\newcommand{\spec}{{\it specialized}}
\newcommand{\relevant}{{\it relevant}}
\newcommand{\loadtime}{{\it avg\_download\_time}}
\newcommand{\type}{{\it source\_type}}
\newcommand{\sitelang}{{\it source\_language}}
\newcommand{\format}{{\it data\_format}}
\newcommand{\reliable}{{\it reliable}}
\newcommand{\site}{{\it source}}
\newcommand{\siteup}{{\it up}}
\newcommand{\occ}{{\it access}}
\renewcommand{\occ}{{\it cref}}
\newcommand{\defobj}{{\it default\_class}}
\newcommand{\defpath}{{\it default\_path}}
\newcommand{\ob}{{\it class}}
\newcommand{\oattr}{{\it class\_att}}

\newcommand{\attr}{{\it att\_val}}
\newcommand{\isa}{{\it is\_a}}
\newcommand{\inst}{{\it instance}}

\newcommand{\namemap}{\ensuremath{n}}
\newcommand{\nameset}{\ensuremath{N}}
\newcommand{\namef}[1]{\namemap(#1)}
\newcommand{\nameo}[0]{{\mathit{n}}}
\newcommand{\name}[1]{\nameo_{#1}}

\newcommand{\commadots}[0]{,\ldots ,}

\newcommand{\DLV}[0]{\textrm{DLV}\xspace}

\newcommand{\iec}[0]{i.e.,\ }
\newcommand{\egc}[0]{e.g.,\ }

\newcommand{\head}[1]{H(#1)}

\newcommand{\body}[1]{B(#1)}
\newcommand{\bodyp}[1]{B^+(#1)}
\newcommand{\bodyn}[1]{B^-(#1)}

\newcommand{\la}{\leftarrow}
\newcommand{\naf}{{\it not}\,}

\newcommand{\ground}[1]{{\mathit{ground}}(#1)}
\renewcommand{\ground}[1]{{\mathcal{G}}(#1)}

\newcommand{\NP}{\mbox{${\rm N\!P}$}}
\renewcommand{\NP}{\mbox{\rm NP}}

\newcommand{\SigmaP}[1]{{\Sigma}_{#1}^{P}}

\newcommand{\at}{{\mathit At}}
\renewcommand{\at}{{\mathcal A}}

\newcommand{\Lit}{{\mathit Lit}}

\newcommand{\tuple}[1]{\langle#1\rangle}
\renewcommand{\tuple}[1]{(#1)}

\newcommand{\nop}[1]{}

\newcommand{\seqo}{\mathit{seq}}
\newcommand{\seq}[1]{\seqo(#1)}

\newtheorem{definition}{Definition}
\newtheorem{proposition}{Proposition}
\newtheorem{theorem}{Theorem}

\newtheorem{example}{Example}

\newcounter{myenumctr}

\newcommand{\querysite}{{\it query\_source}}

\begin{document}

\title[A Knowledge-Based Approach for Selecting Information Sources]
{A Knowledge-Based Approach for
 \\ 
Selecting Information 
Sources\thanks{Part of the material in this paper has appeared, in
preliminary form, in the Proceedings of the Eighth International Conference 
on Principles of Knowledge
Representation and Reasoning (KR '02), pp.\ 49-60, April 22-25,
Toulouse, France, 2002.}
%\\[-1ex]
%~
}

\author[Thomas Eiter,\ Michael Fink,\ and\ Hans Tompits]
{THOMAS EITER,\  MICHAEL FINK,\ and\ HANS TOMPITS\\
Technische Universit\"{a}t Wien, \\
Institut
f{\"u}r Informationssysteme\\
Favoritenstra{\ss}e 9-11, A-1040 Vienna, Austria\\
email: {\{eiter,michael,tompits\}@kr.tuwien.ac.at}
}

\submitted{16 April 2004}
\revised{13 August 2005}
\accepted{20 April 2006}

\pagerange{\pageref{firstpage}--\pageref{lastpage}}
\volume{\textbf{Vol.} (Nr.):}
\jdate{Month 2006}
\setcounter{page}{1}
\pubyear{2006}

\maketitle

\label{firstpage}

\begin{abstract}
Through the Internet and the World-Wide Web, a vast number of
information sources has become available, which offer information on
various subjects by different providers, often in heterogeneous
formats. This calls for tools and methods for building an advanced
information-processing infrastructure. 
One issue in this area is the selection of suitable information
sources in query answering. In this paper, we present a
knowledge-based approach to this problem, in the setting where one among
a set of information sources (prototypically, data repositories)
should be selected for evaluating a user query. We use extended logic
programs (ELPs) to represent rich descriptions of the information
sources, an underlying domain theory, and user queries in a formal
query language (here, XML-QL, but other languages can be handled as
well). Moreover, we use ELPs for declarative query analysis and
generation of a query description.  Central to our approach are declarative
\emph{source-selection programs}, for which we define syntax and
semantics. Due to the structured nature of the considered data items,
the semantics of such programs must carefully respect implicit context
information in source-selection rules, and furthermore combine it with
possible user preferences. A prototype implementation of our approach
has been realized exploiting the {\DLV} KR system and its {\PLP}
front-end for prioritized ELPs. We describe a representative
example involving specific movie databases, and report about
experimental results.
\end{abstract}

\begin{keywords}
knowledge representation, nonmonotonic reasoning, logic programming, 
answer-set programming, information-source selection, data repositories, 
preference handling.
\end{keywords}

\section{Introduction} 
\label{sec:intro}

Through the Internet and the World-Wide Web (WWW), a wealth of
information has become available to a large group of users. A huge
number of documents, files, and data re\-positories on a range of
subjects are offered by different providers, which may be non-profit
individuals, organizations, or companies. Such data repositories are
currently in heterogeneous formats, but the trend is that XML becomes
a future de-facto standard for releasing data on the Web, since this
eases data exchange. Nonetheless, the quality of their
contents may differ significantly with respect to aspects such as their accuracy,
coverage of certain topics and completeness for them, or refresh cycle,
to mention just a few.  

Accessing and processing data on the Web calls for developing tools
and methods for an advanced information-processing infrastructure.
{\em Mediators}~\cite{wied-93} and special {\em information agents}
(``{middle agents}''~\cite{deck-etal-97}), which provide various
services including finding, selecting, and querying relevant
information sources, play an important role here.  The potential of
knowledge-based approaches---and in particular of logic
pro\-gram\-ming---for developing reasoning components for intelligent
information agents is recognized in the AI community and outlined, \egc by
\citeN{dimo-kaka}, \citeN{eite-etal-00j}, and \citeN{sadr-toni-99+}.

In this paper, we pursue this issue further and present a declarative
approach for information-source selection in the following
setting. Given a query by a user in some formal query language and a
suite of information sources over which this query might be
evaluated, which of these sources
is the best to answer the query, i.e., 
such that the utility of the answer, measured by the quality of the
result and other criteria (e.g., costs), is as large as possible for
the user?  Note that this problem is in fact not bound to information
sources on the Web but is of interest in any context where different
candidate information sources (e.g., scientific databases, newspaper
archives, stock exchange predictions, etc.) are available and one of
them should be selected. Selection of a single source may be desired
because of (high) cost associated with accessing each source, for
instance.  Furthermore, problems arising by integrating data from different
sources (like inconsistencies between sources) can be circumvented
this way.

For a concrete example, consider the following scenario to illustrate our ideas.  

\begin{example} \label{exa:movie}{\rm 
Assume that some agent has access to XML information sources, $s_1$, $s_2$,
and $s_3$, about movies. Furthermore, suppose that the following XML-QL\footnote{For
details about XML-QL, cf.~Section~\ref{XML-QL}.}
query is handed to the agent, which informally asks a
source for the titles of all movies directed by Alfred 
Hitchcock:

{\small
\begin{verbatim}
    FUNCTION HitchcockMovies($MovieDB:"Movie.dtd") {
      CONSTRUCT <MovieList> {
        WHERE <MovieDB> <Movie>
                 <Title> $t </Title>
                 <Director> <Personalia>
                     <FirstName> "Alfred" </FirstName>
                     <LastName> "Hitchcock" </LastName>
                 </Personalia> </Director>
              </Movie> </MovieDB> 
        IN source($MovieDB)     
        CONSTRUCT <Movie> $t </Movie>
     } </MovieList> }  
\end{verbatim} 
}

\noindent
Here, {\tt \$t} is a variable into which the value of attribute
{\tt Title} is selected, for usage in the resulting construction.
Suppose the agent knows that $s_1$ is a very good source for information about
directors, while $s_2$ has usually good coverage about person data; all
that is known about $s_3$, however, is that it is not very reliable. In this situation,
we would expect that the agent selects $s_1$ for querying.}
\end{example}

Obviously, a sensible solution to this problem is nontrivial
and involves various aspects such as taking basic properties of the
information sources, knowledge about their contents, and knowledge
about the particular application domain into account. These aspects have to be
suitably combined, and reasoning may be needed to elicit implicit
knowledge. 
We stress that the general problem considered here is distinct from a
simple keyword-based search as realized by Web engines
like Google,\footnote{Google's homepage is found at \url{http://www.google.com}.}
% ~\cite{google},
and consequently we do not propose a method
for competing with these tools here.%
\footnote{Note that Google  
does not index XML files or databases underlying Web query interfaces, 
and hence cannot be readily applied for the purposes considered here.}
In fact, we are concerned with {\em qualitative} selection from different alternatives, 
based on rich meta-knowledge and a formal semantics, thereby respecting 
preference and context information which involves heuristic defaults.

\medskip
Our approach, which incorporates aspects mentioned above, makes several
contributions, which are briefly summarized as follows. 

\medskip

(1)  
We base our method on the {\em answer-set programming}
paradigm, in which problems are encoded in terms of
nonmonotonic logic programs and solutions are extracted from the models
of these programs (%.\footnote{
cf.\ \citeN{bara-03} for a comprehensive treatise on answer-set programming).
More precisely, we use \emph{extended logic programs}
(ELPs) under the answer-set semantics~\cite{gelf-lifs-91}, augmented
with \emph{priorities} (cf., e.g., \citeN{brew-eite-99}, \citeN{delg-etal-02}, or 
\citeN{inou-saka-00b} for work about priorities in answer-set programming)
and \emph{weak constraints}~\cite{bucc-etal-99b,leon-etal-2002-dlv}, to
represent rich descriptions of the information sources, an underlying
domain theory, and queries in a formal language. We perform query
analysis by ELPs and compute {\em query descriptions}. Here, we
consider XML-QL~\cite{deut-etal-99}, but our approach is not
committed to semi-structured data and XML per se, and other formal query
languages can be handled as well (e.g., \citeN{schi-02} adopts our query-analysis method for the
ubiquitous SQL language for relational databases).

(2) 
At the heart, a declarative {\em source-selection program} represents
both \emph{qualitative} and \emph{quantitative} criteria for source
selection, in terms of rules and soft 
constraints.  The
rules may access information supplied by other programs, including
object and value occurrences in the query.  For example, a rule $r_1$
may state that a query about a person Alfred Hitchcock 
should be posed to source $s_1$. Furthermore, ordinal rule priorities
can be employed in order to specify source-selection preference.
For
example, a priority may state that a certain rule mentioning a last
name in the query is preferred over another rule mentioning the
concept $\mathit{person}$ only. Rules and priorities are of qualitative
nature and are taken into account for singling out a coherent decision for
the source selection in model-theoretic terms. Quantitative criteria 
(like, e.g., cost) are used to
discriminate between different such options by means of an objective
function which is optimized. To this end, conditions in terms of
conjunctions of literals can be stated whose violation is penalized
(e.g., the selection of a certain source might be penalized but not
strictly forbidden), and total penalization is minimized. Such a two
step approach seems to be natural and provides the user with a range
of possibilities to express his or her knowledge and selection desires in convenient
form.

(3) We consider the interesting and, to the best of our knowledge, novel issue of {\em
contexts} in nonmonotonic logic programs, which is similar to
preference based on specificity~\cite{delg-scha-94,geer-verm-93,geer-verm-95}.
Structured data items
require a careful definition of the selection semantics, since an
attribute might be referenced following a path of indirections, starting from a root
object and passing through other objects. In
Example~\ref{exa:movie}, 
for instance,
the attribute $\it FirstName$
is referenced with the path $\it Movie/Director/Personalia/$ $\it FirstName$,
which starts at an object of type $\it Movie$ and passes through
objects of type $\it Director$ and $\it Personalia$. Each of these
objects opens a context in which $\it FirstName$ is referenced along
the remaining path. Intuitively, a context is less specific the closer
we are at the end of the path. Thus, for example, the reference from
$\it Personalia$ is less specific than from $\it Movie$, and the
latter should have higher priority. Note that such priority is not
based on \emph{inheritance} (which is tailored for ``flat''
objects). Therefore, inheritance-based approaches such as those by 
\citeN{laen-verm-90a} or \citeN{bucc-etal-96} do not apply here.
Furthermore, implicit priorities derived from context information as
above must be combined with explicit user preferences from the
selection policy, and arising conflicts must be resolved.

(4) We have implemented a prototype, based on the 
 KR system {\DLV}~\cite{leon-etal-2002-dlv} and its front-end
 {\PLP}~\cite{delg-etal-01}
 for prioritized ELPs,  which we used to 
build a model application involving movie
information sources. It comprises several XML databases, wrapped from movie
databases on the Web, and handles queries in XML-QL. Experiments that
we have conducted showed that the system behaved as expected on a number
of natural queries, some of which require reasoning from the
background knowledge to identify the proper selection.

\medskip
The reason to use a knowledge-based approach---and in particular an
answer-set programming approach---for source selection rather than a standard
decision-theoretic approach based on utility functions is motivated by
the following advantages: 

\begin{itemize}
  \item Source-selection programs, which are special kinds of extended logic
programs, are declarative and have a well-defined formal semantics,
both under qualitative as well as under quantitative criteria.

\item The formalism is capable of handling incomplete information and
performing nonmonotonic inferences, which, arguably, is an inherent
feature of the problem domain under consideration.

\item Changes in the specification of the source-selection process
are easily incorporated by modifying or adding suitable rules or
constraints, without the need for re-designing the given program, as
may be the case, \egc in procedural languages. 

\item Finally, the
declarative nature of the answer-set semantics formalism permits also
a coupling with sophisticated ontology tools, as well as with reasoning
engines for them, providing advanced features for the domain
knowledge. In particular, the approach of \citeANP{eite-etal-04} 
\shortcite{eite-etal-04,eite-etal-05}, providing a
declarative coupling of logic programs under the answer-set semantics with description-logic
knowledge bases, can be integrated into our framework. 
\end{itemize}

We note that while we focus here on selecting a \emph{single source}, our
approach can be easily extended to select \emph{multiple
information sources}, as well as to perform ranked selections (cf.\ Section~\ref{sec:extensions}).

The rest of this paper is organized as follows. The next section
contains the necessary prerequisites from answer-set programming and XML-QL, and
Section~\ref{sec:system} gives a brief outline of our approach.  In
Section~\ref{sec:query-abstract}, we 
consider the generation of an internal query representation, while
Section~\ref{sec:site-descr} addresses the modeling of sources.
Section~\ref{sec:ss}, then, is devoted to source-selection programs and 
includes a discussion of some of their properties.
The implementation and the movie
application, as well as experimental results, are the topics of Section~\ref{sec:appl}. 
Section~\ref{sec:rel-conc}
addresses related work, and Section~\ref{sec:concl} concludes the main part of the
paper with a brief summary and open research issues.
Certain technical details and additional properties of our approach are relegated to an appendix.

\section{Preliminaries}\label{sec:elps}

\subsection{Answer-set programming}
\label{sec:ASP}

We recall the basic concepts of answer-set programming. 
Let $\cal L$ be a function-free first-order language.
Throughout this paper, we denote variables by 
alphanumeric strings starting with an upper-case letter, anonymous variables by `\_', 
and constants by alphanumeric strings starting with a lower-case letter or by 
a string in double quotes.

An
\emph{extended logic program} (ELP)~\cite{gelf-lifs-91} is a finite
set of rules over
$\cal L$ of form
\begin{equation}\label{rule:1}
L_0 \la L_1\commadots L_m, \naf L_{m+1}\commadots \naf L_n,
\end{equation}
where each $L_i$, $0 \leq i \leq n$, 
is a {\em literal}, i.e., an
atom $A$ or a negated atom $\neg A$, and ``\naf'' denotes \emph{negation as
failure}, or \emph{default negation}. 
Intuitively, a rule of form~(\ref{rule:1}) states that we can conclude
$L_0$ if (i) $L_1\commadots L_m$ are known and (ii) $L_{m+1}\commadots
L_n$ are \emph{not} known.
For a rule $r$ as above, we call the literal $L_0$ the \emph{head} of 
$r$ (denoted $\head{r}$) and the set 
$\{L_1\commadots L_m,\naf L_{m+1}$ $\commadots \naf L_n\}$
the \emph{body} of $r$ (denoted $\body{r}$). 
Furthermore,  we define 
$\bodyp{r}=\{L_1\commadots L_m\}$ and $\bodyn{r}=\{L_{m+1}\commadots L_n\}$.
If $\body{r}=\emptyset$, then $r$ is called a \emph{fact}.
We write
$r(V_1\commadots V_n)$ to indicate that rule $r$ has 
variables $V_1\commadots V_n$.
To ease notation,  for any program $\Pi$
and any set $S$ of literals, $\Pi\cup S$ stands for the program 
$\Pi\cup\{L\la\mid L\in S\}$.
Finally, for a literal $L$, we write $\neg L$ to denote its
complementary literal, \iec $\neg L=A$ if $L=\neg A$, and $\neg L=\neg
A$ if $L=A$, for any atom $A$.

The semantics of an ELP $\Pi$ is given in terms of the semantics of
its ground instantiation, $\ground{\Pi}$, over the Herbrand universe
$U_{\cal L}$ of $\cal L$, which is the 
language generated by $\Pi$. The program $\ground{\Pi}$ contains all
instances of rules from $\Pi$, \iec where the variables are (uniformly) replaced with
arbitrary terms from $U_{\cal L}$. Recall that a literal, rule,
program, etc., is {\em ground} iff it contains no variables.  In what
follows, we assume that all such objects are ground.

An \emph{interpretation}, $X$, is a consistent set of (ground) literals,
i.e., $X$ does not contain a complementary pair $A$, $\neg A$ of
literals. A literal, $L$, is \emph{true} in $X$
if $L\in X$, and \emph{false} otherwise. The body,
$\body{r}$, of a rule $r$ is true in $X$ 
iff (i)~each $L\in\bodyp{r}$ is true in $X$ and (ii) each
$L\in\bodyn{r}$ is false in $X$. 
Rule $r$ is true in $X$ 
iff $\head{r}$ is true in $X$
whenever $\body{r}$ is true in $X$. 
Finally, a program, $\Pi$, is true in $X$, or $X$ is a \emph{model} of $\Pi$, 
iff all rules in $\Pi$ are true in $X$.
We write $X\models \alpha$ to indicate that an object $\alpha$, which may be 
either  a literal, the body of a rule, a rule, or a program, 
is true in $X$.
 
Let $X$ be a set of literals and $\Pi$ a program.
The \emph{Gelfond-Lifschitz reduct}, or simply \emph{reduct}, $\Pi^X$, of 
$\Pi$ \emph{relative to} $X$ is given by 
\[
\Pi^X
=
\{ \head{r} \la \bodyp{r}\mid r\in \Pi \mbox{ and } \bodyn{r}\cap X=\emptyset \,\}.
\]
We call $X$ an \emph{answer set} of $\Pi$
iff $X$ is a minimal model of $\Pi^X$ with respect to set inclusion.
Observe that any answer set of $\Pi$ is \emph{a fortiori} a model of $\Pi$.
The set of all \emph{generating rules} of an answer set $X$ with respect to 
$\Pi$ is given by
\[
\GR{\Pi}{X}
=
\{r\in \Pi\mid X\models \body{r}\}\mbox{.}
\]

\begin{example} \label{example:kb-tv}{\rm 
Let $\Pi= \{\,s \la \naf t; \ n \la \ ; \  t\la n,\naf s; \ w\la t\,\}$. For the
interpretation $X_1 = \{ n, t, w\}$, we have 
$\Pi^{X_1} = \{ n \la \ ; \  t\la n; \ w\la t \}$. Clearly, 
$X_1$ is a minimal model of $\Pi^{X_1}$, and thus $X_1$ is an answer set of
$\Pi$. Note that $X_2=\{ s, n \}$ is another answer set of $\Pi$. 
}
\end{example}

A (possibly non-ground) program $\Pi$ is \emph{locally stratified}~\cite{przy-88} iff there
exists a mapping $\lambda$ assigning each literal occurring in
$\ground{\Pi}$ a natural number such that, for each rule $r\in \ground{\Pi}$,
it holds that
(i)~$\lambda(\head{r})\geq\max_{L\in\bodyp{r}}\lambda(L)$ and
(ii)~$\lambda(\head{r})>\max_{L\in\bodyn{r}}\lambda(L)$.
Note that $\Pi$ is
(\emph{globally}) {\em stratified}~\cite{apt-etal-88} if, additionally,
for all positive (resp., negative)
literals $L$ and $L'$ with the same predicate, $\lambda(L)=\lambda(L')$ holds.
It is well-known that
if a program is locally stratified, then it has at most one answer
set.

A refinement of the answer-set semantics is the admission of preferences 
among the rules of a given ELP, yielding the class of 
\emph{prioritized logic programs}.
Several approaches in this respect have been introduced 
in the literature, like, \egc those by \citeN{brew-eite-99} or \citeN{inou-saka-00b}; 
here, we use a preference approach based on a method  
due to \citeN{delg-etal-02}, defined as follows.

Let $\Pi$ be an ELP and $<$ a strict partial order between the
elements of $\Pi$ (i.e., $<$ is an irreflexive and transitive relation).
Informally, for rules $r_1,r_2\in \Pi$, the relation $r_1<r_2$  
expresses that $r_2$ has preference over $r_1$.
Define the relation $<_{\mathcal{G}}$ over the ground instantiation 
$\ground{\Pi}$ of $\Pi$ by setting 
$\hat{r}_1<_{\mathcal{G}} \hat{r}_2$ iff $r_1<r_2$, for 
$\hat{r}_1,\hat{r}_2\in \ground{\Pi}$.
Then, the pair $(\Pi,<)$ is called a 
\emph{prioritized extended logic program}, or simply 
a \emph{prioritized logic program}, if the relation 
$<_{\mathcal{G}}$ is a strict partial order.

The semantics of prioritized programs is as follows.
Let $(\Pi,<)$ be a prioritized logic program where
$\Pi$ is ground, and let $X$ be an answer set of $\Pi$.
We call $X$ a \emph{preferred answer set} of $(\Pi,<)$
iff there exists an enumeration
  \(
  \langle r_i\rangle_{i\in I}
  \)
  of\/ $\GR{\Pi}{X}$
 such that, for every $i,j\in I$, we have that:
\begin{enumerate}
  \item[$(P_1)$]\label{def:order:preserving:zero}
    $\bodyp{r_i}\subseteq\{\head{r_k}\mid k<i\}$;
  \item[$(P_2)$]\label{def:order:preserving:one}
    if $r_i<r_j$, then $j<i$;
    and
  \item[$(P_3)$]\label{def:order:preserving:two}
    if
    $r_i<r'$
    and 
    \(
    r'\in {\Pi\setminus\GR{\Pi}{X}},
    \)
    then
      $\bodyp{r'}\not\subseteq X$
    or 
    $\bodyn{r'}\cap\{\head{r_k}\mid k<i\}\neq\emptyset$.
  \end{enumerate}

Conditions $(P_1)$--$(P_3)$ realize a strongly ``prescriptive'' 
interpretation of  preference, in the sense that, whenever 
$r_1<r_2$ holds,  it is ensured that $r_2$ is known to be applied 
or blocked ahead of $r_1$ (with respect to the order of rule 
application).
More specifically, $(P_2)$ guarantees that all generating 
rules are applied according to the given order, whilst 
$(P_3)$ assures that any preferred yet inapplicable rule 
is either blocked due to the non-derivability of its 
prerequisites or because it is defeated by higher-ranked or unrelated rules.
As shown by \citeN{delg-etal-02}, 
the selection of preferred answer sets can be encoded by means of 
a suitable translation from prioritized logic 
programs into standard ELPs.

Preferred answer sets of a prioritized program $(\Pi,<)$ 
where $\Pi$ is non-ground are given by the preferred 
answer sets of the prioritized program 
$(\ground{\Pi},<_{\mathcal{G}})$, where 
$<_{\mathcal{G}}$ is as above.
Note that the concept of prioritization realizes a 
\emph{filtering} of the answer sets of a given 
program $\Pi$, as every preferred answer set of 
$(\Pi,<)$ is an answer set of $\Pi$, but not vice versa.

Besides imposing \emph{qualitative selection criteria}, 
like assigning preferences between different rules, 
another refinement of the answer-set semantics are 
\emph{weak constraints} \cite{bucc-etal-99b,leon-etal-2002-dlv}, 
representing a \emph{quantitative filtering of answer sets}.  
Formally, a weak constraint is an expression of form
\begin{equation}\label{eq:weak}
\Leftarrow \; L_1\commadots L_m,\naf L_{m+1}\commadots \naf L_n \;[w:l],
\end{equation}
where each $L_i$, $1\leq i\leq n$, is a literal (not necessarily
ground) and $w,l\geq 1$ are natural numbers.%
\footnote{The part ``$[w:l]$'' is convenient syntactic sugar for the
original definition by \citeN{bucc-etal-99b}, which merely provided a
partitioning of the weak constraints in priority levels.}
 The number $w$ is the
\emph{weight} and $l$ is the \emph{priority level} of the weak
constraint (\ref{eq:weak}). Given an interpretation $X$, the weight of
a ground weak constraint $c$ of the above form with respect to a level
$l'\geq 1$, $\mathit{weight}_{c,l'}(X)$, is $w$, if $X\models L_i$, $1\leq i \leq
m$, $X\not\models L_{m+j}$, $1\leq j \leq n$, and $l'=l$, and 0
otherwise; the weight of a non-ground weak constraint $c$ with
respect to a level $l$, $\mathit{weight}_{c,l}(X)$, is given by
$\sum_{c'\in \ground{c}} \mathit{weight}_{c',l}(X)$, where
$\ground{c}$ denotes the set of all ground instances of $c$.  Weak
constraints select then those answer sets $X$ of the weak-constraint-free part of 
a program $\Pi$ for which the associated vector
\[
\mathit{weights}(X) =
(\mathit{weight}_{l_{\max}}(X),\mathit{weight}_{l_{\max-1}}(X),\ldots,\mathit{weight}_0(X))
\]
is lexicographic smallest, where $l_{\max}$ is the highest priority
level occurring and $\mathit{weight}_{l}(X) = \sum_{c\in \mathit{wc}(\Pi)}
\mathit{weight}_{c,l}(X)$, for each $l$, with $\mathit{wc}(\Pi)$ denoting the set of all 
weak constraints occurring in $\Pi$.
Informally, first those
answer sets are pruned for which the weight of violated constraints is
not minimal at the highest priority level; from the remaining answer sets, those are 
pruned where the sum of weights of violated
constraints in the next lower level is not minimal, and so on. For
example, if we add in Example~\ref{example:kb-tv} the weak constraints
$c_1:$ $\Leftarrow n,\naf\, w\, [3:1]$ and $c_2$: $\Leftarrow t,
w\,[1:2]$, then we have $\mathit{weights}(X_1)=(1,0)$ and
$\mathit{weights}(X_2)=(0,3)$; hence, the answer set $X_1$ is
discarded.  

The numeric lexicographic preference can be reduced by usual
techniques to an objective function $H^\Pi(X)$, which assigns each
answer set $X$ an integer such that those answer sets $X$ for which
$H^\Pi(X)$ is minimal are precisely those for which
$\mathit{weights}(X)$ is lexicographically smallest. In the above
example, $X_2$ is selected as the ``optimal'' answer set.
While the availability of both weights and levels is syntactic sugar, they are
very useful for expressing preferences in a more natural and convenient
form. In the example above, putting $c_2$ at level 2 dominates $c_1$
which is at level 1. Weights within the same layer can be used for fine-tuning. 
For  formal details and more discussion, we refer the
reader to~\citeN{leon-etal-2002-dlv}.

\subsection{XML-QL}
\label{XML-QL}

We next introduce basic concepts of
XML-QL~\cite{deut-etal-99}, a query language for data 
stored in the \emph{Extensible Markup Language} (XML).
We assume that
the reader is familiar with XML, which has emerged as a standard for
providing (semi-structured) data on the Web. While syntactically
similar to the \emph{Hypertext Markup Language} (HTML), features have
been added in XML for data-representation purposes such as
user-defined tags and nested elements. Unlike relational or
object-oriented data, XML is \emph{semi-structured}, \iec it can have
irregular (and extensible) structure and attributes (or schemas) are stored with the data.  
The structure of an XML document
can be optionally modeled and validated against a \emph{Document Type
Descriptor} (DTD). In this paper, we take a database-oriented view of
XML documents, considering them as databases and a corresponding DTD
as its database schema. For a comprehensive introduction to
semistructured data and database aspects about them, 
we refer to \citeN{abit-etal-00}. 

XML-QL is a declarative, relationally complete query language for XML data, 
which can not only query XML data, but also construct new XML documents from 
query answers, \iec it can also be used to restructure XML data. Its syntax 
deviates from the well-known ``select-from-where syntax'' of the 
\emph{Structured Query Language} (SQL), but can be 
decomposed into three syntactical units as well: 
\begin{enumerate}
\item 
a {\em where part}  
(keyword {\tt WHERE}), specifying a selection condition by element reference
and comparison predicates; 

\item
a {\em source part} (keyword {\tt IN}), 
declaring a data source for the query (an external file, or an internal variable); and 
\item a {\em construct part} (keyword {\tt CONSTRUCT}), defining a structure 
for the resulting document.
\end{enumerate}
In the latter part, subqueries can be built by nesting.

XML-QL uses element patterns to match data in an XML document. 
Elements are referenced by their names and are traversed according to the 
XML source structure. Thus, \emph{reference paths} can be identified with every 
matching.
Variables are in general not bound to elements but
to element contents (but syntactic sugar exists for element binding). 
Furthermore, elements can be joined by values using the same variable in two 
matchings, \iec theta-joins can be expressed.

Let us briefly illustrate the most basic concepts in the following example; 
for further details, we refer to~\citeN{deut-etal-99} and \citeN{abit-etal-00}.

\begin{example} \label{exa:xmlql-prelim}{\rm 
Throughout the paper, we consider XML-QL queries stored as XML-QL functions, 
which serves two purposes. First, it allows us to efficiently query several 
XML documents by dynamic bindings of data sources, and second, we can 
additionally specify that a data source has to obey a certain DTD. 
The following query is represented as an XML-QL function. Upon its invocation,  
the variable {\tt \$MovieDB} is instantiated with the name of an XML document 
to be queried, which has to be structured according to the DTD detailed in~\ref{app:dtd}.

\medskip
{\small
\begin{verbatim}
    FUNCTION ExampleQuery($MovieDB:"Movie.dtd") {
      WHERE <MovieDB> <Movie> $m1 </Movie> </MovieDB> IN source($MovieDB),
            <Actor> $a </Actor> IN $m1, 
            <MovieDB> <Movie> $m2 </Movie> </MovieDB> IN source($MovieDB),  
            <Actor> $a </Actor> IN $m2,
            $m1 != $m2
      CONSTRUCT <x2Actor> $a </x2Actor> 
    }
\end{verbatim} 
}

\noindent
In the where part of the above query, variables {\tt \$m1} 
and {\tt \$m2} are bound to different matchings under the
reference path
{\it MovieDB/Movie}. Furthermore, 
the two matchings are joined by 
the (common) value of variable {\tt \$a}, referenced under element 
{\it Actor}. The construct part of the query creates a new XML 
document by listing the values of {\tt \$a} marked-up with tags 
{\tt <x2Actor>}. Intuitively, the query returns all actors 
found in a given XML document about movies which act in at least two movies.
}
\end{example}

\section{Overview of the approach}
\label{sec:system}

Before presenting the technical details of our approach, it is helpful to give a
short overview.  While the motivating example in
Section~\ref{sec:intro} is simple, it shows that the source-selection
process involves different kinds of knowledge, including
\begin{itemize}
 \item knowledge about which ``interesting'' information should be
       extracted from a given formal query expression $Q$, 
\item  knowledge about the
information sources and their properties, 
 \item background knowledge about the application domain
and the ontology used for its formalization, and
\item  specific rules which 
guide the source selection, based on preferences or generic principles. 
\end{itemize}

In our approach, this is formalized in terms 
of the notion of a {\em selection
base} 
\[
{\cal S} = \tuple{\prgm{qa},\prgm{sd},\prgm{dom},\prgm{sel},<_u},
\]
where $\prgm{qa},\prgm{sd},\prgm{dom}$ are ELPs, called \emph{query-analysis program}, 
\emph{source description}, and \emph{domain theory}, respectively, and 
$(\prgm{sel},<_u)$ is a prioritized logic program with a special syntax, called 
\emph{source-selection program}.
Given a selection base $\cal S$ as above, the possible solutions of a query $Q$ 
relative to $\cal S$ are determined by the \emph{selection answer sets} of the 
source-selection program $(\prgm{sel},<_u)$, which are defined as preferred answer 
sets of a prioritized logic program ${\mathcal E}({\cal S},Q)=(\prgm{Q},<)$, associated with $\cal S$ 
and $Q$, as shown in Figure~\ref{fig:program}.

The components of a selection base serve the following purposes:

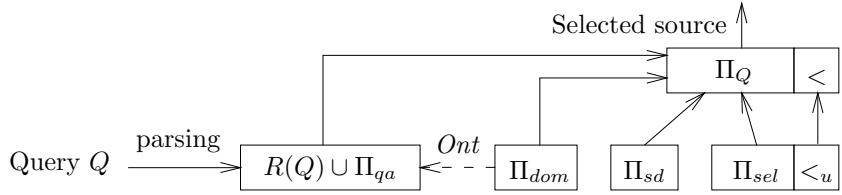
\begin{figure}
{\small\centerline{\setlength{\unitlength}{0.00087489in}
\begingroup\makeatletter\ifx\SetFigFont\undefined%
\gdef\SetFigFont#1#2#3#4#5{%
  \reset@font\fontsize{#1}{#2pt}%
  \fontfamily{#3}\fontseries{#4}\fontshape{#5}%
  \selectfont}%
\fi\endgroup%
{\renewcommand{\dashlinestretch}{30}
\begin{picture}(5090,1164)(0,-10)
\path(4230,282)(4725,282)(4725,12)
	(4230,12)(4230,282)
\path(3195,282)(3195,687)(3960,687)
\path(3840.000,657.000)(3960.000,687.000)(3840.000,717.000)
\path(4440.000,1017.000)(4410.000,1137.000)(4380.000,1017.000)
\path(4410,1137)(4410,867)
\path(3825,282)(4185,597)
\path(4114.446,495.402)(4185.000,597.000)(4074.936,540.557)
\path(2925,282)(3420,282)(3420,12)
	(2925,12)(2925,282)
\path(4868,292)(4868,597)
\path(4898.000,477.000)(4868.000,597.000)(4838.000,477.000)
\path(4500,282)(4410,597)
\path(4471.812,489.859)(4410.000,597.000)(4414.121,473.375)
\path(1395,282)(2475,282)(2475,12)
	(1395,12)(1395,282)
\path(3622,282)(4027,282)(4027,12)
	(3622,12)(3622,282)
\path(4725,282)(4995,282)(4995,12)
	(4725,12)(4725,282)
\path(3960,867)(4725,867)(4725,597)
	(3960,597)(3960,867)
\path(4725,867)(4995,867)(4995,597)
	(4725,597)(4725,867)
\put(2700,237){\makebox(0,0)[b]{{\SetFigFont{10}{13.2}{\rmdefault}{\mddefault}{\updefault}$\mathit{Ont}$}}}
\put(0,102){\makebox(0,0)[lb]{{\SetFigFont{10}{13.2}{\rmdefault}{\mddefault}{\updefault}Query $Q$}}}
\put(765,237){\makebox(0,0)[lb]{{\SetFigFont{10}{13.2}{\rmdefault}{\mddefault}{\updefault}parsing}}}
\put(4320,957){\makebox(0,0)[rb]{{\SetFigFont{10}{13.2}{\rmdefault}{\mddefault}{\updefault}Selected source}}}
\put(3825,57){\makebox(0,0)[b]{{\SetFigFont{10}{13.2}{\rmdefault}{\mddefault}{\updefault}$\Pi_{sd}$}}}
\put(4500,57){\makebox(0,0)[b]{{\SetFigFont{10}{13.2}{\rmdefault}{\mddefault}{\updefault}$\Pi_{sel}$}}}
\put(4860,57){\makebox(0,0)[b]{{\SetFigFont{10}{13.2}{\rmdefault}{\mddefault}{\updefault}$<_u$}}}
\put(4365,642){\makebox(0,0)[b]{{\SetFigFont{10}{13.2}{\rmdefault}{\mddefault}{\updefault}$\Pi_Q$}}}
\put(4860,642){\makebox(0,0)[b]{{\SetFigFont{10}{13.2}{\rmdefault}{\mddefault}{\updefault}$<$}}}
\put(3195,57){\makebox(0,0)[b]{{\SetFigFont{10}{13.2}{\rmdefault}{\mddefault}{\updefault}$\Pi_{dom}$}}}
\put(1935,57){\makebox(0,0)[b]{{\SetFigFont{10}{13.2}{\rmdefault}{\mddefault}{\updefault}$R(Q)\cup \Pi_{qa}$}}}
\dashline{60.000}(2925,147)(2475,147)
\path(2595.000,177.000)(2475.000,147.000)(2595.000,117.000)
\path(1890,282)(1890,822)(3960,822)
\path(3840.000,792.000)(3960.000,822.000)(3840.000,852.000)
\path(720,147)(1395,147)
\path(1275.000,117.000)(1395.000,147.000)(1275.000,177.000)
\end{picture}
}}}
\caption{Using a selection base ${\cal S}= \tuple{\prgm{qa},\prgm{sd},\prgm{dom},\prgm{sel},<_u}$ 
for source 
selection for a query $Q$.\label{fig:program}}
\end{figure}

\begin{description}
\item[%
Query-analysis program $\boldsymbol{\prgm{qa}}$:]
For any query $Q$ as in Example~\ref{exa:movie}, a high-level
description is extracted from a low-level (syntactic) representation,
$R(Q)$, given as a set of elementary facts, by applying 
$\prgm{qa}$ to $R(Q)$ and ontological knowledge, $\mathit{Ont}$,
about concepts (types) and synonyms from the domain
theory $\prgm{dom}$, in terms of facts for predicates $\ob(O)$ and $\synonym(C_1,C_2)$. 
Informally, the rules of $\prgm{qa}$ single out the essential
parts of $Q$, such as occurrence of attributes and values in the
query, comparison and joins, or subreference paths of attributes from
objects on a reference path.
For instance, in Example~\ref{exa:movie}, the
attribute $\it FirstName$ from an object of type $\it Director$ 
is referenced via path $\it
Personalia/FirstName$ 
on the reference path $\it Movie/Director/Personalia/FirstName$ from the
root.

\item[Source description $\boldsymbol{\prgm{sd}}$:] 
This program contains information about the available
sources, using special predicates for query topics, 
cost aspects, and technical aspects.

\item[Domain theory $\boldsymbol{\prgm{dom}}$:]
The 
agent's knowledge about the specific application domain (like, e.g., the movie area) is represented in the domain theory $\prgm{dom}$. 
It includes ontological
knowledge and further background knowledge, permitting (modest)
common-sense reasoning. The ontology is assumed to have concepts
(classes), attributes, and instance and subconcept
information, which are provided via 
$\ob(O)$, $\oattr(C,A)$,
 $\inst(O,C)$, and $\isa(C_1,C_2)$ 
predicates, respectively.  
Furthermore, it is assumed to have information about concept synonyms, provided via
predicate $\synonym(C_1,C_2)$. The ontology may be partly established
using meta-informa\-tion about the data in the information sources
(e.g., an XML DTD), and with ontology rules. Since ontological reasoning is 
orthogonal to our
approach, we do not consider it here and refer to \citeN{eite-etal-02d} for
a further elaboration.

\item[Source-selection program $\boldsymbol{(\prgm{sel},<_u)}$:]
The information
source selection is specified by rules and constraints, which
refer to predicates defined in the above programs. It comprises both
{\em qualitative aspects} and {\em
quantitative aspects} in terms of optimization criteria (concerning,
e.g., cost or response time), which are expressed using weak
constraints~\cite{bucc-etal-99b}. Furthermore, the user can define 
preferences between rules, in terms of a
strict partial order, $<_u$. These preferences are combined with
implicit priorities that emerge from the {\em context} in which source
selection rules should be applied, and possible preference conflicts are resolved.

\end{description}

Given a query $Q$, the overall evaluation relative to ${\cal S}%
$, then, proceeds in three
steps: 

\begin{description}
\item[Step 1 (query description):] The input query $Q$ is parsed
and mapped into the internal query representation, $R(Q)$, which is
extended using $\prgm{qa}$ and $\mathit{Ont}$ to the full query description.

\item[Step 2 (qualitative selection):] 
From $R(Q)$, $\prgm{qa}$, $\prgm{sd}$, and $\prgm{dom}$, the
qualitative part of $\prgm{sel}$ is used to single out
different query options by respecting qualitative aspects only, where
explicit preferences, $<_u$, and implicit priorities must be taken
into account. To this end, a priority relation $<$ is computed on
rules, which is then used in a prioritized logic program
$(\prgm{Q},<)$. Candidate solutions are computed as preferred answer sets of
$(\prgm{Q},<)$.

\item[Step 3 (optimization):] Among the
candidates of Step~2, the one is chosen which is best
under the quantitative aspects of $\prgm{sel}$, and the
selected source is output. 
\end{description}

%%%%%%%%%%%%%%%%%%%%%%%%%%%%%%%%%%%%%%%%%%%%%%%%%%%%%%%%%%%%%%%%%%%%%%
\section{Query description}\label{sec:query-abstract}
%%%%%%%%%%%%%%%%%%%%%%%%%%%%%%%%%%%%%%%%%%%%%%%%%%%%%%%%%%%%%%%%%%%%%%

An integral feature 
of our approach is a meaningful description of a
given formal query expression $Q$ in an internal format. For our
purposes, we need a suitable representation of the constituents of $Q$
in terms of predicates and objects.  Simply mapping $Q$ (which is
represented as a string) to logical facts which encode its syntax tree
(i.e., the external format) does not serve our purposes. Rather, we
need a meta-level description which provides ``interesting''
information about $Q$, such as occurrence of an attribute or a value
in $Q$, related to the scope of appearance.

For example, in the query of Example~\ref{exa:movie}, the value
``Hitchcock'' occurs in a selection on
the attribute $\it LastName$ reached by the reference path $\it
Personalia/LastName$ from $\it Director$. In the internal query
representation, this selection will be represented by the fact
$\selects(o_3,{\it equal},``{\it Hitchcock}")$,
where $o_3$ is an
internal name for the full reference path 
\(
``{\it
Movie/Director/Personalia/LastName}"
\)
(i.e., the reference path from the root), and by a fact 
$\occ(o_3,``{\it
Director}",``{\it Personalia/LastName}",q_1)$,
where $q_1$ is an
internal identifier for the query.  
Also, 
a fact $\occurs(o_3,``{\it
Hitchcock}")$ will be present that less specifically states that $``{\it Hitchcock}"$ 
is associated with 
this reference path. 

The general format of these three predicates, which play a vital part in our architecture,
is as follows:

\begin{itemize}
\item $\occ(O,C,P,Q)$ states that within the full reference path $O$ in
the syntax tree for query $Q$, the path from $C$ to the leaf is $P$;%
\footnote{Note that \citeN{eite-etal-02d} and \citeN{fink-02} name this predicate ${\it
 access}$, and reference paths are called \emph{access paths}.} 

\item ${\it occurs}(O,V)$ states that the value $V$ is associated with
the full reference path $O$ in the overall query; and

\item $\selects(O,R,V)$ is similar to $\it occurs$, but
details the association with a comparison operator $R$.
\end{itemize}

In accord to the syntactical units of XML-QL, in our query-analysis method we 
adopt the general view in which a query expression consists of  a 
{\em where part}, a {\em source part}, and a {\em construct part}.
For the description of $Q$, we employ
facts on designated predicates, which are independent of a fixed query
language. These facts are divided into two groups, which we refer
to as {\em parser facts} and {\em derived facts}, respectively.

%%%%%%%%%%%%%%%%%%%%%%%%%%%%%%%%%%%%%%%%%%%%%%%%%%%%%%%%%%%%%%%%%%%%%%%%%%
\subsection{Parser facts} 
%%%%%%%%%%%%%%%%%%%%%%%%%%%%%%%%%%%%%%%%%%%%%%%%%%%%%%%%%%%%%%%%%%%%%%%%%%
The first group of facts, denoted $R(Q)$, is generated by a query
parser, and is regarded as a ``{low-level}'' part of the query
representation. The query parser scans the query string $Q$ for
extracting ``interesting'' information, and assembles information
about structural information (such as about subqueries, and in which
of them a reference to a certain attribute is made). The main purpose
of $R(Q)$ is to filter and reduce the information which is present in
the syntax tree of $Q$, and to assemble it into suitable facts. For
that, the parser must introduce identifiers (names) for queries,
subqueries, and other query constituents---in particular, references to
{\em items}\/ (i.e., attributes or concepts), which in a query are
selected or compared to values or other items.  Every item reference is
given by a maximal reference path in $Q$, which we call an {\em item
reference path {\rm (}IRP\/{\rm )}}.  The parser names each
occurrence of an IRP with a unique constant (note that the same IRP
may have multiple occurrences in~$Q$).

\begin{example}%\rm 
In Example~\ref{exa:movie}, the query is named $q_1$. 
It has a subquery in the construct {\tt <Movie\-List>} part, 
identified by $q_2$. 
There are three IRPs, namely 
\[
\begin{tabular}{l}
``${\it Movie/Title}$",\\
``${\it
Movie/Director/Personalia/FirstName}$", and\\ 
``${\it
Movie/Director/Personalia/LastName}$". 
\end{tabular}
\]
Their identifiers are 
$o_1$, $o_2$, and
$o_3$, respectively.
\end{example}

%%%%%%%%%%%%%%%%%%%%%%%%%%%%%%%%%%%%%%%%%%%%%%%%%%%%%%%%%%%%%%%%%%%%%%
\subsection{Derived facts}
%%%%%%%%%%%%%%%%%%%%%%%%%%%%%%%%%%%%%%%%%%%%%%%%%%%%%%%%%%%%%%%%%%%%%%

The second group of facts are those which are derived from $R(Q)$ by means of
a further analysis. Compared to $R(Q)$, these facts can be regarded as a
``{high-level}'' description of the query. In particular, for
the attribute or concept at the end of an IRP, the {\em contexts}\/ of
reference are determined, which are the suffixes of the IRP starting
at some concept (as known from the underlying ontology).  Intuitively,
instances of this concept have the referenced item as a (nested)
attribute. Detaching the leading concept from the suffix results in 
the notion of a \emph{context-reference pair}, defined as follows:

\begin{definition}
A pair $(C,P)$, where $C$ is a concept from the ontology and $P$ is a
path, is a {\em context-reference pair} {\rm (}CRP\/{\rm )} of a query $Q$  if $Q$
contains an IRP with suffix ``$C/P$''.
\end{definition}

\begin{example} 
Continuing Example~\ref{exa:movie}, assume that the concepts {\it MovieDB}, {\it
Movie}, {\it Director}, and {\it Person} are in the ontology, and it is
known that ``{\it Personalia}'' is a synonym of ``{\it Person}'' in the ontology.
Then, from the IRP $o_1$ = ``${\it Movie/Title}$",
the CRPs 
$$
(``{\it MovieDB}",``{\it Movie/Title}") \textrm{ and } (``{\it Movie}",``{\it Title}")
$$
are determined,
and from the IRP $o_2$ = ``${\it Movie/Director/Personalia/LastName}$", the CRPs 
$$\begin{array}{l}
(``{\it MovieDB}",``{\it Movie/Director/Personalia/FirstName}"), \\%[.8ex]
(``{\it Movie}",``{\it Director/Personalia/FirstName}"),\\%[.8ex] 
(``{\it Director}",``{\it Personalia/FirstName}"), \textrm{ and }\\%[.8ex]
(``{\it Personalia}",``{\it FirstName}")
\end{array}
$$
are obtained.
\end{example}

The high-level description facts are computed declaratively
by evaluating a query-analysis logic program, $\prgm{qa}$, to
which the facts $R(Q)$ and further facts $\mathit{Ont}$, which provide
ontological knowledge about concepts and synonyms from the domain
theory, are added as ``input''. Furthermore, the program enriches the
low-level predicate $\subpath$ by synonym information and closing
$\subpath$ transitively.  In summary, the query description is given
by the (unique) answer set of the logic program $\mathit{Ont} \cup
\prgm{qa}\cup R(Q)$.

A detailed list of all query-description predicates, as well as the complete
query-analysis program, can be found in~\ref{subsec:qd-preds}.

%%%%%%%%%%%%%%%%%%%%%%%%%%%%%%%%%%%%%%%%%%%%%%%%%%%%%%%%%%%%%%%%%%%%%%
\section{Source description}\label{sec:site-descr}
%%%%%%%%%%%%%%%%%%%%%%%%%%%%%%%%%%%%%%%%%%%%%%%%%%%%%%%%%%%%%%%%%%%%%%

Besides query information and domain knowledge, the source-selection 
process requires a suitable description of the information sources to 
select from. This is provided by means of meta-knowledge represented 
in the source-description part of the knowledge base, given in the 
form of a (simple) logic program, $\prgm{sd}$, which is assumed to 
have a unique answer set. 
Different predicates can be used for this purpose, depending on 
the specific application.
In the following, we introduce, in an exemplary fashion, a basic 
suite of predefined  \emph{source-description predicates}, which 
cover several aspects of an information source:

\begin{enumerate}
\item[(i)]
Thematic aspects:

\begin{itemize}
\item $\acc(S,T,V)$:  source $S$, topic $T$, value $V$;
\item $\cov(S,T,V)$: source $S$, topic $T$, value $V$;
\item $\spec(S,T)$: source $S$, topic $T$;
\item $\relevant(S,T)$: source $S$, topic $T$.

\end{itemize}

The first two predicates express the accuracy and coverage of a source
for a topic, using values from $\{{\it low}, {\it med}, {\it high}\}$. The others are
for stating that a source is specialized or relevant for a particular
topic, respectively.

\item[(ii)]
Cost aspects:

\begin{itemize} 
\item $\loadtime(S,V)$: source $S$, value $V$;
\item $\downtime(S,V)$: source $S$, value $V$;
\item $\charge(S,V)$: source $S$, value $V$.
\end{itemize}

Costs for accessing an information source can be expressed by these
predicates, again using values ${\it low}$, ${\it med}$, 
${\it high}$, and, for $\charge$,
also ${\it no}$. While $\charge$ is used for direct costs, $\loadtime$ and
$\downtime$ are indirect costs (taking network traffic into account).

\item[(iii)]
Technical aspects:

\begin{itemize} 
\item $\type(S,T_1,T_2)$: source $S$, organizational type $T_1$, query type $T_2$;
\item $\sitelang(S,L)$: source $S$, language $L$;
\item $\format(S,F)$: source $S$, format $F$;
\item $\freq(S,V)$: source $S$, value $V$;
\item $\lastupd(S,D)$: source $S$, date $D$;
\item $\reliable(S,V)$: source $S$, value $V$;
\item $\site(S)$: source $S$;
\item $\siteup(S)$: source $S$. 
\end{itemize}

Different kinds of sources are distinguished by their type of organization
(commercial or public) and by the type of data access provided
(queryable, downloadable, or both). Besides source language and data
format (XML, relational, HTML, text, or other), the frequency of data
update (low, medium, or high), the date of the last update, or the
reliability of a source (low, medium, or high) may be criteria for source
selection. Finally, $\site$ and $\siteup$ are used to identify sources
and to express that a source is currently accessible, respectively.
\end{enumerate}

As already pointed out, the above predicates are just a rudiment of a 
vocabulary for source description, and we
are far from claiming that they capture all aspects or that they capture each one 
in sufficient detail or granularity (like, e.g., the three-valued scale
used). However, the user or administrator has the possibility to
introduce further predicates and define them in the source-description
program $\prgm{sd}$.  Note that $\prgm{sd}$ can take
advantage of default rules to handle incomplete information, e.g.,
that a source is accessible by default, or that the language of text
items is English. 

We assume here furthermore that detailed
source descriptions are edited by an administrator of an overall information
system hosting 
the considered selection process.
This does
not preclude that a preliminary or partial description is created
automatically, addressing aspects such as source language, type, data
format, etc., nor that the information system is open for new sources
entering it, advertising their description to a source
registration. However, a number of aspects for selection, such as
coverage, specialization, or relevance, might be difficult to assess
automatically and require experience gained from interaction with a
source like in real-life scenarios (think of different travel
agencies offering flights, for instance). Here, the administrator might bring in
such knowledge initially, and the description might be updated
in accord to new information obtained, e.g., by performance monitoring
and user feedback. For updating an employed description, approaches 
such as those discussed by 
\citeN{alfe-etal-02} or \citeN{eite-etal-01j} may be applied.
In general, however, this is a complex and interesting issue, but is 
beyond the scope of this paper.

In concluding, we remark that the proviso that $\prgm{sd}$ possesses a unique answer
set can be ensured, \egc by requiring (local) stratification of $\prgm{sd}$,
or by the condition that its well-founded model is total. In
principle, the case of multiple answer sets of $\prgm{sd}$ could be admitted as
well, which would give rise to different scenarios that could be handled in
different ways; e.g., adhering to a credulous or skeptical reasoning principle,
according to which the different scenarios are considered {\em en
par} or such that only selections in all scenarios are retained,
or to a preference-based approach which discriminates between the
different scenarios.  However, we do not elaborate further on
this issue.

%%%%%%%%%%%%%%%%%%%%%%%%%%%%%%%%%%%%%%%%%%%%%%%%%%%%%%%%%%%%%%%%%%%%%%%%
\section{Source selection}\label{sec:ss}
%%%%%%%%%%%%%%%%%%%%%%%%%%%%%%%%%%%%%%%%%%%%%%%%%%%%%%%%%%%%%%%%%%%%%%%%

We now introduce the central part of our architecture, viz.\ \emph{source-selection
programs}. 
Basically, a source-selection program is a prioritized
logic program  $(\prgm{sel},<_u)$ having four parts: (i)~a core unit
$\prgm{sel}^c$, containing the actual source-selection rules,
(ii) a set
$\prgm{sel}^{\it aux}$ of auxiliary rules,
(iii) an order relation
$<_u$ defined over members of $\prgm{sel}^c$, and 
(iv) an optimization
part $\prgm{sel}^o$, containing weak constraints.

%%%%%%%%%%%%%%%%%%%%%%%%%%%%%%%%%%%%%%%%%%%%%%%%%%%%%%%%%%%%%%%%%%%%%%%%
\subsection{Syntax}\label{sec:syntax}
%%%%%%%%%%%%%%%%%%%%%%%%%%%%%%%%%%%%%%%%%%%%%%%%%%%%%%%%%%%%%%%%%%%%%%%%

We first make the vocabulary of source-selection programs formally precise.  

\begin{definition}
A \emph{source-selection vocabulary},
$\at_{\it sel}$, 
consists of the following 
pairwise disjoint categories:

\begin{itemize}
\item[{\rm (}i\/{\rm )}] function-free 
vocabularies $\at_{\it qd}$, $\at_{\it sd}$, and $\at_{\it
dom}$, referred to as the \emph{query-description vocabulary}, the 
\emph{source-description vocabulary}, and the \emph{domain-theory 
vocabulary of $\at_{\it sel}$}, respectively, where $\at_{\it qd}$ 
and $\at_{\it sd}$ contain the predicates introduced in 
Section~\ref{sec:query-abstract} and \ref{sec:site-descr};

\item[{\rm (}ii\/{\rm )}] the predicate $\qs(S,Q)$, expressing 
that source $S$ is selected for evaluating query $Q$;

\item[{\rm (}iii\/{\rm )}] 
the predicates $\defobj(O,C,Q)$ and $\defpath(O,P,Q)$;
and

\item[{\rm (}iv\/{\rm )}] a set $\at_{\it aux}$ of auxiliary predicates.
\end{itemize}

\end{definition}

Informally, the predicates $\defobj(O,C,Q)$ and $\defpath(O,P,Q)$ are 
projections of $\occ(O,C,P,Q)$ and serve to specify a
default status for selection rules depending on
context-refer\-ence pairs 
matched in the query $Q$. For example, 
a predicate 
$\defobj(O,``{\it Person}",Q)$ in
the body of rule $r$ expresses that $r$ is eligible in case the concept
${\it Person}$ occurs in the reference path $O$ and there is no other rule $r'$
that refers to some
CRP $(C',P')$ matched in $Q$.
These defaults are semantically realized using a
suitable rule ordering.

The set of all literals over atoms in $\at_\ell$, for
$\ell
\in \{{\it qd,sd,dom,aux,sel}\}$, is denoted by $\Lit_\ell$.

\begin{definition}
Let $\at_{\it sel}$ be a source-selection vocabulary.
A \emph{source-selection program over $\at_{\it sel}$} is a tuple
$(\prgm{sel},<_u)$, where
\begin{itemize}
\item[{\rm (}i\/{\rm )}] $\prgm{sel}$ is a collection of 
rules over $\at_{\it sel}$ consisting of the following parts:

\begin{enumerate}
\item[{\rm (}a\/{\rm)}] 
 the \emph{core unit}, $\prgm{sel}^c$, containing rules of form 
\[
\qs(S,Q) \la  L_1\commadots L_m, \naf L_{m+1}\commadots \naf L_n,
\]

\item[{\rm (}b\/{\rm )}]  a set $\prgm{sel}^{\it aux}$ of \emph{auxiliary rules} of form 
\[
L_0 \la  L_1\commadots L_m, \naf L_{m+1}\commadots \naf L_n,
\]
and

\item[{\rm (}c\/{\rm )}]  an \emph{optimization part}, $\prgm{sel}^o$, containing weak
constraints of form
\[
\Leftarrow \; L_1\commadots L_m,\naf L_{m+1}\commadots \naf L_n \;[w:l],
\]
\end{enumerate}
where $L_0$ is either a literal from $\Lit_{\it aux}$ or is 
of form $\neg\qs(\cdot,\cdot)$, $L_i\in\Lit_{\it sel}$ for $1\leq
i\leq n$, and $w,l\geq 1$ are
integers, and

\item[{\rm (}ii\/{\rm )}] 
$<_u$ is a strict partial order between rules in
$\prgm{sel}^c$. 
\end{itemize}

The elements of $<_u$ are called \emph{user-defined preferences}.
If ${r_1}<_u {r_2}$, 
then $r_2$ is said to have \emph{preference over} $r_1$.
\end{definition}

The rules in the core unit $\prgm{sel}^c$ serve for selecting a
source, based on information from the domain description, 
the source description, the query  description, 
and possibly from auxiliary rules. The latter may be
used, \egc for evaluating complex conditions. 
In terms of $<_u$, preference
of source selection can be expressed.
As well, the weak constraints in $\prgm{sel}^o$ are used to
filter answer sets under {quantitative conditions}.

By assembling all constituents for source selection into a single compound, 
we arrive at the notion of a \emph{selection base}, as informally 
described in Section~\ref{sec:system}.

%%%%%%%%%%%%%%%%%%%%%%%%%%%%%%%%%%%%%%%%%%%%%%%%%%%%%definition
\begin{definition}
Let $\at_{\it sel}$  be a source-selection vocabulary.
A \emph{selection base over $\at_{\it sel}$} is a quintuple ${\cal S}=\tuple{\prgm{qa},\prgm{sd},\prgm{dom},\prgm{sel},<_u}$, consisting of the query-analysis program $\prgm{qa}$ over $\at_{\it qd}$, programs $\prgm{sd}$ and $\prgm{dom}$ over $\at_{\it sd}$ and $\at_{\it dom}$, respectively,
and a source-selection program $(\prgm{sel},<_u)$ over $\at_{\it sel}$.

\end{definition}
 
Given that the components $\prgm{qa}$, $\prgm{sd}$, and
$\prgm{dom}$ are understood, the source-selection program
$(\prgm{sel},<_u)$ in a selection
base ${\cal S}$ is the most interesting part, and ${\cal S}$ might be
referred to just by this program.
Furthermore, we assume in what follows that the source-selection 
vocabulary $\at_{\it sel}$ contains only those constants actually 
appearing in the elements of a selection base over $\at_{\it sel}$.
Thus, we usually leave $\at_{\it sel}$ implicit.
 
\begin{example}\label{example:site-selection}{\rm
Consider a simple source-selection program, $(\prgm{sel},<_u)$, for
our movie domain, consisting of the following constituents:

\begin{itemize}
\item Source-selection rules:

\medskip
\(
\begin{array}{rr@{~}c@{~}l}
r_1: & \qs(s_2,Q) & \la & \defobj(O,``{\it Person}",Q); \\[1ex]
r_2: & \qs(s_1,Q) & \la & \selects(O,{\it equal},``Hitchcock"), \\ 
&&& \occ(O,``{\it Director}",``{\it Personalia/}\\
&&& \hphantom{\occ(}{\it LastName}",Q);\\[1ex]
r_3: & \qs(S,Q) & \la & \defpath(O,``{\it LastName}",Q), \\
&&&  \defobj(O,T,Q),\acc(S,T,{\it high}).
\end{array}
\)

\bigskip

\item Auxiliary rules:

\medskip
\(
\begin{array}{rr@{~}c@{~}l}
r_4: & {\it high\_acc}(T,Q) & \la & \occ(O,T,P,Q), {\it accurate}(S,T,{\it high}); \\[1ex]
r_5: & {\it high\_cov}(T,Q) & \la & \occ(O,T,P,Q), {\it covers}(S,T,{\it high}).
\end{array}
\)

\bigskip

\item Optimization constraints:

\medskip
\(
\begin{array}{r@{~}c@{~}l}
c_1: & \ \ \Leftarrow & \qs(S,Q), {\it high\_acc}(T,Q), \\
&& \naf \acc(S,T,{\it high}) ~[10:1];\\[1ex]
c_2: & \ \ \Leftarrow & \qs(S,Q), {\it high\_cov}(T,Q), \naf \cov(S,T,{\it high}) ~[5:1].
\end{array}
\)

\bigskip

\item User preferences:

\medskip
\(
\begin{array}{l}
{r_1(Q,\_)}<_u {r_3(Q,\_,\_,\_)}.
\end{array}
\)

\end{itemize}

\medskip
Intuitively, 
$r_1$ advises to choose source $s_2$ if
the query involves persons and no more specific rule is eligible. 
Rule $r_2$ states to choose source~$s_1$ if the query
contains an explicit select on the movie director Hitchcock. Rule
$r_3$ demands to choose a source if, on some 
query reference path, ``{\it LastName}'' is accessed under some 
concept $T$ (with arbitrary
intermediate reference path), and the source is highly accurate for
$T$. Rules $r_4$ and $r_5$ define auxiliary predicates
which hold on concepts $T$ appearing in the query 
such that some
source with high accuracy and coverage for $T$
exists. The weak constraints $c_1$ and $c_2$ state penalties for
choosing a source that does not have high accuracy (assigning weight 10) or coverage
(assigning weight 5) for a concept in the query while such a source exists.
Finally, ${r_1(Q,\_)}<_u {r_3(Q,\_,\_,\_)}$ expresses
preference of instances of $r_3$ over $r_1$ on the same query.
}
\end{example}

%%%%%%%%%%%%%%%%%%%%%%%%%%%%%%%%%%%%%%%%%%%%%%%%%%%%%%%%%%%%%%%%%%%%%%%%%%%%%%%
\subsection{Semantics}\label{sec:semantics}
%%%%%%%%%%%%%%%%%%%%%%%%%%%%%%%%%%%%%%%%%%%%%%%%%%%%%%%%%%%%%%%%%%%%%%%%%%%%%%%

The semantics of a source-selection program $(\prgm{sel},<_u)$ in a
selection base ${\cal S} = \tuple{\prgm{qa},$ $\prgm{dom},\prgm{sd},\prgm{sel},<_u}$ 
on a query
$Q$ is given by means of a \emph{selection answer set} of $(\prgm{sel},<_u)$, which is
defined as a preferred answer  set of a prioritized ELP 
${\mathcal E}({\cal S},Q)$ 
associated with ${\cal S}$ and $Q$. The program ${\mathcal E}({\cal S},Q)$
is of the form $(\prgm{Q},<)$, where 
program $\prgm{Q}$ 
contains ground instances of rules and constraints
in $\prgm{sel}$, and further rules ensuring that a single source is
selected per query and rules defining the default-context
predicates. The order relation $<$ is formed from the user preferences
$<_u$ and the implicit priorities derived from context references in
the core unit and from auxiliary rules.
Thereby, preference information must be suitably combined, as well
as arising conflicts resolved,
which we do by means of a cautious conflict-elimination policy. 

We commence the formal details with the following notation: 
For any rule $r=\head{r}\la\body{r}
$, its \emph{defaultization},
$r^\Delta$, is given by
$\head{r}\la\body{r},\naf \neg\head{r}$.\footnote{Defaultization is
also known in the literature as the 
\emph{extended version} of a rule~\cite{kowa-sadr-90,vann-verm-2002}.}
We assume that user-defined preferences between rules carry over 
to their defaultizations.

\begin{definition}\label{def:evalprgm} 
Let  ${\cal S}=
\tuple{\prgm{qa},\prgm{sd},$ $\prgm{dom},\prgm{sel},<_u}$ be 
a selection base and $Q$ a query.
Then, the program $\prgm{Q}$ 
contains all ground instances of
the rules and constraints in $\prgmsup{sel}{aux}\cup\prgmsup{sel}{o}$, 
as well as all ground instances of
the following rules:

\begin{itemize}
\item[{\rm (}i{\rm )}] the defaultization $r^\Delta$ of $r$, 
for each $r\in \prgm{sel}^c$;

\item[{\rm (}ii{\rm )}]

the \emph{structural rule}
\begin{equation}\label{eq:structural}
\neg \qs(S,Q) \la \qs(S',Q),S\neq S';
\end{equation}
and
\item[{\rm (}iii{\rm )}] 
the \emph{default-context rules}
\[
\begin{array}{r@{~}c@{~}l}%\label{eq:default}
\defobj(O,C,Q)  & \la & \occ(O,C,\_,Q),%\label{eq:default:1}
\\[.8ex]
\defpath(O,P,Q)  &\la & \occ(O,\_,P,Q).%\label{eq:default:2}
\end{array}
\]

\end{itemize}

\end{definition}

Intuitively, the defaultization makes the selection rules in $\prgm{sel}$
defeasible with respect to the predicate $\qs$, the structural rule 
enforces that
only one source is selected, and the
default-context rules 
define the two default predicates. 
Since our language has no function symbols, $\prgm{Q}$ is finite, and its size
depends on the constants appearing in $\prgm{qa}$, $R(Q)$, $\prgm{sd}$,
and $\prgm{dom}$.

\begin{definition}
For ${\cal S}=
\tuple{\prgm{qa},\prgm{sd},$ $\prgm{dom},\prgm{sel},<_u}$ and query $Q$,  
we call any answer set of 
$\prgm{qa}\cup R(Q)\cup \prgm{sd}\cup\prgm{dom}$ a 
\emph{selection input of $\cal S$ for $Q$}.
The set of all selection inputs of $\cal S$ for $Q$ is denoted by $\SI({\cal S},Q)$.
For $Y\in\SI({\cal S},Q)$, we define 
\[
Y_{\it def}=Y\cup\{\defobj(o,c,q),\, \defpath(o,p,q)\mid \occ(o,c,p,q)\in Y\}.
\]
\end{definition}

Note that, in general, a selection base may admit multiple selection
inputs for a query $Q$. However, in many cases, there may exist only a
single selection input---in particular, if the source description
$\prgm{sd}$ and the domain knowledge $\prgm{dom}$ have unique
answer sets. In our framework, this is ensured if, \egc these
components are represented by (locally) stratified programs.

\begin{definition}
Given a selection base ${\cal S}=
\tuple{\prgm{qa},\prgm{sd},$ $\prgm{dom},\prgm{sel},<_u}$ and a query $Q$,
a rule $r\in \prgm{Q}$ is \emph{relevant for $Q$} iff there is some
$Y\in\SI({\cal S},Q)$ such that $B^\dagger(r)$ is true in $Y_{\it
def}$, where $B^\dagger(r)$ results from $\body{r}$ by deleting each
element which does not contain a predicate symbol from $\at_{\it
qd}\cup\at_{\it sd}\cup\at_{\it dom}\cup 
\{ \defobj, \defpath\}$.
\end{definition}

In the sequel, we denote for any binary relation $R$ its 
transitive closure by $R^*$.

We continue with the construction of the preference relation $<$, used
for interpreting a source-selection program $(\prgm{sel},<_u)$ relative
to a selection base ${\cal S}$ and a query $Q$ in terms of an
associated prioritized logic program $(\prgm{Q},<)$.

Informally, the specification of $<$ depends on the following
auxiliary relations:

\begin{itemize}

\item the preference relation
$\preceq_c$, taking care of implicit context priorities;

\item the intermediate relation $\unlhd$, representing a direct
combination of user-defined preferences with context preferences; and

\item the preference relation $<'$, removing possible conflicts within
the joined relation $\unlhd$ and ensuring transitivity of the
resultant order $<$.  \end{itemize}

More specifically, the relation $\preceq_c$ is the first step
towards $<$, transforming structural context
information into explicit preferences, in virtue of the following
specificity conditions:

\begin{itemize} 

\item default contexts for concepts are assumed to be more specific than
default contexts for attributes; 

\item context references are more specific than default contexts; and

\item with respect to the same IRP, rules with a larger CRP $(C,P)$ are 
considered more specific than rules with a shorter CRP 
$(C^{\,\prime},P^{\,\prime})$  (\iec where $P^{\,\prime}$ is a subpath of $P$).

\end{itemize}

The second step in the construction of $<$ is the relation $\unlhd$,
which is just the union of the user preferences $<_u$ and the context
priorities $\preceq_c$. In general, this will not be a strict partial
order. To enforce irreflexivity, we remove all tuples
$\name{r_1}\unlhd \name{r_2}$ lying on a cycle, 
resulting in $<'$.  Finally, taking the transitive closure
of $<'$ yields $<$.  The formal definition of relation $<$ is as
follows.

\begin{definition}\label{def:order}
Let  ${\cal S}$ be a selection base, $Q$ a query, and
$\prgm{Q}$ as in Definition~\ref{def:evalprgm}. 
For $r_1,r_2 \in \prgm{Q}$, define

\begin{enumerate}
\item[{\rm (}i{\rm )}]
${r_1}\preceq_c{r_2}$  iff  $r_1$ and $r_2$ are relevant for $Q$, $r_1\neq r_2$, and one
of {\rm(}$O_1${\rm)}--{\rm(}$O_3${\rm)} holds:

\begin{itemize}
\item[{\rm (}$O_1${\rm )}] $\defpath(o_1,p_1,q)\in\body{r_1}$, and  
either $\occ(o_2,t_2,p_2,q)\in\body{r_2}$ 
 or $\defobj(o_2,t_2,q)\in\body{r_2}$,

\item[{\rm (}$O_2${\rm )}] $\defobj(o_1,t_1,q)\in\body{r_1}$ and 
$\occ(o_2,t_2,p_2,q)\in\body{r_2}$,

\item[{\rm (}$O_3${\rm )}] $\occ(o,t_1,p_1,q)\in\body{r_1}$, $\occ(o,t_2,p_2,q)\in\body{r_2}$, 
and $t_1/p_1$ is a subpath of  $t_2/p_2$,

\end{itemize}

\item[{\rm (}ii{\rm )}]
${r_1}\unlhd{r_2}$ iff  ${r_1}<_u{r_2}$ and $r_1$ and $r_2$ are relevant for 
$Q$, or ${r_1}\preceq_c{r_2}$, and

\item[{\rm (}iii{\rm )}]
${r_1}<'{r_2}$ iff  ${r_1}\unlhd{r_2}$ but not ${r_2}\unlhd^*{r_1}$.

\end{enumerate}

Then, the relation $<$ is given as the transitive closure of $<'$.

\end{definition}

\begin{example}\label{example:site-selection-eval}
Reconsider $(\prgm{sel},<_u)$ from
Example~\ref{example:site-selection}. Suppose the domain ontology 
contains the concepts ``{\it MovieDB}'', ``{\it Actor}'', ``{\it
Movie}'', ``{\it Director}'', and ``{\it Person}'', and that ``{\it
Personalia}'' and ``{\it Person}'' are synonymous. Assume further 
that the query 
of Example~\ref{exa:movie} (represented by $q_1$) has a unique selection 
input $Y$, containing the source-description facts
\begin{center}
$\acc(s_1,``{\it Director}",{\it high})$, $\cov(s_2,``{\it Person}",{\it high})$, and ${\it reliable}(s_3,{\it low})$,
\end{center}
together with the following facts resulting 
from the query description and the default-context rules:

\[
\begin{array}{l}
\occ(o_2,``{\it Person}",``{\it FirstName}",q_1),\\
\occ(o_2,``{\it Director}",``{\it Personalia/FirstName}",q_1),\\
\occ(o_3,``{\it Person}",``{\it LastName}",q_1), \\
\occ(o_3,``{\it Director}",``{\it Personalia/LastName}",q_1), \\
\selects(o_3,{\it equal},``{\it Hitchcock}"), \\
\defobj(o_2,``{\it Person}",q_1), \\
\defobj(o_3,``{\it Person}",q_1), \\
\defobj(o_3,``{\it Director}",q_1), \\
\defpath(o_3,``{\it LastName}",q_1).
\end{array}
\]

These elements are exactly those contributing to relevant instances of 
$\prgm{Q}$. The relevant instances of $r_1$, $r_2$, and $r_3$ 
are given by the ground rules 
$r_1(q_1,o_2)$, $r_1(q_1,o_3)$, $r_2(q_1,o_3)$, and 
$r_3(q_1,s_1,o_3,``{\it D}")$.\footnote{For brevity, we write here and 
in the remainder of this example $``{\it D}"$ 
for $``{\it Director}"$.}
Intuitively, we expect $r_2(q_1,o_3)$ to have highest priority among 
these rule instances, since the bodies of the instances of $r_1$ and $r_3$ 
contain default predicates while $r_2$ references a specific context.
Actually, the order relation $<$ includes, for the relevant instances of
$r_1$, $r_2$, 
and $r_3$, the 
pairs
${r_1(q_1,o_2)} < {r_2(q_1,o_3)}$, ${r_1(q_1,o_3)} < {r_2(q_1,o_3)}$, and ${r_3(q_1,s_1,o_3,``{\it D}")} < {r_2(q_1,o_3)}$.

Note that both $r_4$ and $r_5$ have two relevant
instances. However, they do not influence the above rule ordering.
Informally, they are either unrelated to or ``ranked between''  
$r_2(q_1,o_3)$ and the relevant instances
of $r_1$ and $r_3$\/ {\rm (}since the $\occ$ predicates of $r_4$ and
$r_5$ refer to the same context as the context referenced in the body
of $r_2$, or to a subpath of such a context\/{\rm )}. Hence, the
relevant instance of $r_2$ has highest priority.

As for $r_1$ and $r_3$, the
auxiliary relation $\unlhd$ contains two further structural priorities, namely
$r_3(q_1,s_1,o_3,``{\it D}") \preceq_c {r_1(q_1,o_2)}$ and $r_3(q_1,s_1,o_3,``{\it D}") \preceq_c{r_1(q_1,o_3)}$.
They are in conflict with the user preferences
$
{r_1(q_1,o_2)} <_u {r_3(q_1,s_1,o_3,``{\it D}")}$ and
${r_1(q_1,o_3)} <_u {r_3(q_1,s_1,o_3,``{\it D}")}$,
respectively.
This is resolved in the resultant relation $<$ by removing these 
preferences.

\end{example}

Note that, in Definition~\ref{def:order},  the final order $<$ enforces a
cautious conflict resolution strategy, in the sense that it remains
``agnostic'' with respect to priority information causing
conflicts. Alternative definitions of $<'$, such as removal of a
minimal cutset eliminating all cycles in $\unlhd$, may be considered as well;
however, this may lead to a nondeterministic choice since, in general,
multiple such cutsets exist.
Different choices lead to different orders $<$, which may lead to
different results of the source-selection program. Thus, unless a
well-defined {\em specific} minimal cutset is singled out, by virtue
of preference conflicts, the result of the source-selection process might
not be deterministic. Furthermore, an extended logic program 
component computing a final order
based on minimal cutsets is more involved than a component computing the
relations in Definition~\ref{def:order}.

Combining Definitions~\ref{def:evalprgm} and \ref{def:order},
we obtain the translation $\mathcal{E}(\cdot,\cdot)$ as follows:

\begin{definition}
Let ${\cal S}$ 
be a selection base and $Q$ a query. 
Then, the \emph{evaluation $\mathcal{E}({\cal S},Q)$ of ${\cal S}$ with
respect to $Q$} is given by the prioritized logic program $(\prgm{Q},<)$,
where $\prgm{Q}$ and $<$ are as in Definitions~\ref{def:evalprgm} 
and~\ref{def:order}, respectively.

\end{definition}

Selection answer sets of source-selection programs are then obtained as follows: 

\begin{definition}
Let ${\cal S}=\tuple{\prgm{qa},\prgm{sd},$ $\prgm{dom},\prgm{sel},<_u}$ be
a selection base, $Q$ a query, and ${\mathcal E}({\cal S},Q)=(\prgm{Q},<)$ 
the evaluation of ${\cal S}$ with respect to
$Q$.  Then, $X\subseteq \Lit_{\it sel}$ is a \emph{selection answer set of
$(\prgm{sel},<_u)$ for $Q$ with respect to $\cal S$} iff $X$ is a preferred
answer set
of the prioritized logic program  $(\prgm{Q}\cup Y,<)$, 
for some $Y\in \SI({\cal S},Q)$.

A source $s$ is \emph{selected for $Q$%
} iff $\qs(s,q)$ belongs to some selection answer set of  $(\prgm{sel},<_u)$ 
for $Q$ {\rm (}with respect to $\cal S${\rm )},  where the 
constant $q$ represents~$Q$.
\end{definition}

\begin{example}
{\rm In our running example, $(\prgm{sel},<_u)$ has
a unique selection answer set $X$ with respect to $\cal S$  
for query $q_1$ 
from Example~\ref{example:site-selection}. It contains $\qs(s_1,q_1)$,
which is derived from the core rule $r_2(q_1,o_3)$, having the
highest priority among the applicable rules leading to a single
preferred answer set for the weak-constraint free part of $\prgm{sel}$.
If we replace, \egc $r_1$ by the rule
\[
\qs(s_2,Q) \la \occ(O,``{\it Person}",P,Q)
\]
and adapt the corresponding user preference to
${r_1(Q,\_,\_)}<_u {r_3(Q,\_,\_,\_)},$
\noindent  then the weak-constraint free part of $\prgm{sel}$ has two preferred
answer sets: one, $X_1$, is identical to $X$ {\rm (}where applying
$r_2(q_1,o_3)$ is preferred to applying $r_1(q_1,o_3)$, given that
${r_1(q_1,o_3)} < {r_2(q_1,o_3)}$\/{\rm )}; in the other
answer set, $X_2$, the rule $r_1(q_1,o_2)$ is applied
and $\qs(s_2,q_1)$ is derived. Informally, the replacement removes the
preference of $r_2(q_1,o_3)$ over $r_1(q_1,o_2)$,
since 
the corresponding 
$\occ$ predicates refer to different contexts {\rm
(}$``{\it .../FirstName}"$ and $``{\it .../LastName}"$,
respectively\/{\rm )}. 
Thus, 
$r_1(q_1,o_2)$ has maximal
preference like $r_2(q_1,o_3)$.

Given that $X_1$ has weight~5, caused by violation of
$c_2(s_1,q_1,``{\it Person}")$, but $X_2$ has weight 10, caused by
violation of $c_1(s_2,q_1,``{\it Director}")$, $X_1$ is the selection answer set
of $(\prgm{sel},<_u)$ for $Q$.
}
\end{example}

%%%%%%%%%%%%%%%%%%%%%%%%%%%%%%%%%%%%%%%%%%%%%%%%%%%%%%%%%%%%%%%%%%%%%%%%%%%%%%%
\subsection{Properties}\label{sec:properties}
%%%%%%%%%%%%%%%%%%%%%%%%%%%%%%%%%%%%%%%%%%%%%%%%%%%%%%%%%%%%%%%%%%%%%%%%%%%%%%%

In this section, we discuss some basic properties 
of our framework.

The first property links our evaluation method of source-selection
programs to the usual semantics of prioritized logic programs.  
For this purpose, 
we introduce the following concept: Given logic programs
$\Pi_1$ and $\Pi_2$, we say that $\Pi_1$ is \emph{independent} of
$\Pi_2$ iff each predicate symbol occurring in some rule head of
$\Pi_2$ does not occur in $\Pi_1$.  Intuitively, if $\Pi_1$ is
independent of $\Pi_2$, then $\Pi_1$ may serve as an ``input'' for
$\Pi_2$.  This idea is made precise by the following proposition,
which is an immediate consequence of 
results due to \citeN{Eiter:1997:DD} and \citeN{lifs-turn-94}: 
\begin{proposition}\label{thm:splitting}
Let $\Pi_1$ and $\Pi_2$ be two extended logic programs,  
possibly containing weak constraints, and let $X$ be a set of ground
literals. If\/ $\Pi_1$ is independent of\/ $\Pi_2$, then $X$ is an answer 
set of\/ $\Pi_1\cup\Pi_2$ iff there is some answer set $Y$ of\/ 
$\Pi_1$ such that $X$ is an answer set of\/ $\Pi_2\cup Y$. 
\end{proposition}

Now, taking the specific structure of our source-selection architecture into 
account,  we obtain the following characterization.

\begin{theorem}\label{thm:adequacy}
Suppose ${\cal S} = \tuple{\prgm{qa},\prgm{sd},\prgm{dom},\prgm{sel},<_u}$
is a selection base and $Q$ 
a query. Let
$\mathcal{E}(\mathcal{S},Q)=(\Pi_Q,<)$ and $\Pi_{\cal S}(Q) = \prgm{qa}
\cup R(Q)\cup \prgm{dom} \cup \prgm{sd} \cup \Pi_{Q}$.
Then, $X$ is a selection answer set of
$(\prgm{sel},<_u)$ for $Q$ with respect to $\cal S$ iff $X$ is a preferred
answer set of $(\Pi_{\cal S}(Q),<)$.
\end{theorem}

\begin{proof}
Let $\Pi_0$ denote the program $\prgm{qa}\cup R(Q)\cup\prgm{sd}\cup\prgm{dom}$.
Recall that $X$ is a selection answer set of $(\prgm{sel},<_u)$ for $Q$ with 
respect to $\cal S$ iff $X$ is a preferred answer set of $(\Pi_Q\cup Y,<)$, 
for some answer set $Y$ of $\Pi_0$.
Since the predicate symbols occurring
in the heads of rules in a source-selection program do not occur in
rules from the query description, the source description, or the domain
theory, we obviously have that $\Pi_0$ is independent of $\Pi_Q$.
Moreover, it holds that $X$ is a preferred answer set of $(\Pi_Q\cup Y,<)$ 
only if $X$ is an answer set of $\Pi_Q\cup Y$.
Hence, applying Proposition~\ref{thm:splitting}, we have that $X$ 
is an answer set of $\Pi_Q\cup Y$, for some answer set $Y$ of 
$\Pi_0$, iff $X$ is an answer set of $\Pi_Q\cup\Pi_0$.
From this, the assertion of the theorem is an immediate consequence.
\end{proof}

We remark that from a logic programming point of view,
Theorem~\ref{thm:adequacy} might seem to be a more natural definition
of selection answer sets. However, our approach is motivated by
providing a high-level means for specifying source-selection
problems, which is accomplished by decomposition. Note, in particular,
that
% , for a selection input, 
a user will only need to specify the relation $<_u$ as opposed to $<$.
  Hence, the property of
Theorem~\ref{thm:adequacy} shall rather be understood as a possibility
to ``compile'' a selection base and selection inputs with respect to a
query into a single logic program.

Strengthening Theorem~\ref{thm:adequacy}, the construction of
$\mathcal{E}(\mathcal{S},Q)$ can itself be realized in terms of a 
{\em single} logic program of the form $\Pi_{\cal S}(Q)\cup \prgm{obj}(Q)$ 
over an extended vocabulary, 
by describing preference relations \emph{directly at the object
level}, such that each answer set encodes the priority relation $<$
and is a preferred answer set of $(\Pi_{\cal S}(Q)\cup \prgm{obj}(Q),<)$ if
and only if its restriction to $\Lit_{\it sel}$ is a selection answer
set of $(\prgm{sel},<_u)$ for $Q$ with respect to 
$\cal S$.
More details about this property are given in~\ref{app:pref-object}.  

Concerning the computational complexity of source selection, we note that,
given a query $Q$ and the grounding of the program $\Pi_{\cal S}(Q)$
for a selection base $\cal S$ as in Theorem~\ref{thm:adequacy},
deciding whether $(\prgm{sel},<_u)$ has some selection answer set for
$Q$ is $\NP$-complete (since the grounding of $\prgm{obj}(Q)$ can 
be constructed in polynomial time from the grounding of 
$\Pi_{\cal S}(Q)$), and computing
any such selection answer set is complete for {\rm FP}$^{\rm NP}$, which is
the class of all problems solvable in polynomial time with an \NP\
oracle. However, for a fixed selection base and small query size
(which is a common assumption for databases), the problems are
solvable in {\em polynomial time} (cf.\ again~\ref{app:pref-object}
for more details about the complexity of source-selection programs).

One of the desiderata of our approach is that each answer set selects 
at most one 
source, for any query $Q$.  The following result states that
this property is indeed fulfilled.

\begin{theorem}
Let $X$ be a selection answer set of $(\prgm{sel},<_u)$ for 
query $Q$ with respect to
$\mathcal{S}$. 
Then, for any constant $q$, it holds that 
\[
|\{ s \mid \qs(s,q)
\in X\}| \leq 1.
\]
\end{theorem}

\begin{proof}
The presence of the structural rule (\ref{eq:structural})
in the evaluation program $\Pi_Q$ enforces that, 
whenever $X$ contains two ground atoms 
$\qs(s,q)$ and $\qs(s',q)$,  $X$ must be inconsistent, and thus $X$ 
violates the consistency criterion of answer sets.
\end{proof}

Lastly, the following result concerns the order of application of
source-selection rules, stating that source selection is blocked in
terms of priorities as desired.

\begin{theorem}
Let $X$ be a selection answer set of\/ $(\prgm{sel},<_u)$ for 
query $Q$ with respect to
${\cal S}$, 
and let\/ $r^\Delta\in\Pi_Q$ be the defaultization of some rule $r$
belonging to the grounding of\/ $\prgmsup{sel}{c}$ for $Q$ with
respect to $\cal S$.  
Suppose that $\body{r}$ is true in $X$ but
$\head{r}\notin X$.  Then, there is some $r'\in\Pi_Q$ such that
\begin{enumerate}
\item[{\rm (}i\/{\rm )}]\label{prop:1}%[{\rm (}i\/{\rm )}]
either $r'$ belongs to the grounding of\/ $\prgmsup{sel}{aux}$ for
$Q$ with respect to $\cal S$ 
and $\head{r'} = \neg \head{r}$, 
or $r'$ is the defaultization 
of a rule from the grounding of\/ $\prgmsup{sel}{c}$ for
$Q$ with respect to $\cal S$,
\item[{\rm (}ii\/{\rm )}]\label{prop:2}%[{\rm (}ii\/{\rm )}]
$\body{r'}$ and $\head{r'}$ are true in $X$,
and 
\item[{\rm (}iii\/{\rm )}]\label{prop:3}
either $r^\Delta$ and $r'$ are incompatible with respect to $<$, 
or else $r^\Delta < r'$ holds.
\end{enumerate}
\end{theorem}

\begin{proof}
Given that $X$ is a selection answer set of $(\prgm{sel},<_u)$ for query $Q$ 
with respect to ${\cal S}$, we have that $X$ is a preferred answer set 
of the prioritized logic program  $(\Pi_Q\cup Y,<)$, where $Y$ is 
some selection input of $\cal S$ for query $Q$, and, \emph{a fortiori}, 
that $X$ is an answer set of $\Pi_Q\cup Y$.
From the latter and the hypothesis that $\head{r}\notin X$, it follows that 
$r^\Delta=\head{r}\la\body{r},\naf\neg\head{r}$ 
is not a member of $\GR{\Pi_Q\cup Y}{X}$.
Hence, in view of the assumption that $\body{r}$ is true in $X$, we get that
$\neg \head{r}\in X$ must hold.

Since $X$ is a preferred answer set of $(\Pi_Q\cup Y,<)$, there is
some enumeration \( \langle r_i\rangle_{i\in I} \) of\/ $\GR{\Pi_Q\cup
Y}{X}$ such that Conditions $(P_1)$--$(P_3)$ hold (cf.\
Section~\ref{sec:elps}). We take $r' = r_{\ell}$, where $\ell$ is as
follows. Given that $\neg\head{r}\in X$, there is a smallest index
$i_0\in I$ such that $r_{i_0}\in\GR{\Pi_Q\cup Y}{X}$ and
$\head{r_{i_0}}=\neg\head{r}$.  If $r_{i_0}$ belongs to the grounding
of\/ $\prgmsup{sel}{aux}$ for $Q$ with respect to $\cal S$, then $\ell =
i_0$. Otherwise, by the syntactic form of a source-selection program,
$r_{i_0}$ must be an instance of the structural rule~(\ref{eq:structural}).
By $(P_1)$--$(P_3)$, the
defaultization $\bar{r}$ of a rule from the grounding of\/
$\prgmsup{sel}{c}$ for $Q$ with respect to $\cal S$ must exist such that $\bar{r}=
r_{{j}_0} \in
\GR{\Pi_Q\cup Y}{X}$ and $j_0<i_0$. In this case, $\ell =
j_0$.

We show that $r'$ satisfies Conditions~(i)--(iii).
Clearly, Condition~(i) is satisfied.
Furthermore, Condition~(ii) is an immediate consequence of 
the fact that $r' \in\GR{\Pi_Q\cup Y}{X}$. 
It remains to show that Condition~(iii) holds.

Towards a contradiction, assume that $r' <r^\Delta$. 
Since $r' \in\GR{\Pi_Q\cup Y}{X}$ and $r^\Delta\notin \GR{\Pi_Q\cup Y}{X}$,
from Condition~$(P_3)$ we get that 
$\bodyn{r^\Delta}\cap\{\head{r_k}\mid k<\ell \}\neq\emptyset$, 
as $\bodyp{r}\subseteq X$ and $\bodyp{r^\Delta}=\bodyp{r}$.
Now, obviously $\{\head{r_k}\mid k<\ell \}\subseteq X$.
Moreover, since $\bodyn{r}\cap X=\emptyset$ and 
$\bodyn{r^\Delta}=\bodyn{r}\cup\{\neg\head{r}\}$, 
we obtain that $\neg\head{r}\in \{\head{r_k}\mid k<\ell\}$.
Hence, there must be some $k_0<\ell$ such that 
$r_{k_0}\in\GR{\Pi_Q\cup Y}{X}$ and $\head{r_{k_0}}=\neg\head{r}$. 
But this contradicts the condition that $i_0$ ($\geq \ell$) is the 
smallest index $i$ such that $r_{i}\in\GR{\Pi_Q\cup Y}{X}$ and 
$\head{r_{i}}=\neg\head{r}$.
Hence, we either have that $r'$ and $r^\Delta$ are 
incompatible with respect to $<\,$, or $r^\Delta<r'$ must hold.
\end{proof}

%%%%%%%%%%%%%%%%%%%%%%%%%%%%%%%%%%%%%%%%%%%%%%%%%%%%%%%%%%%%%%%%%%%%%%%%%%%%%
\subsection{Extended source selection}\label{sec:extensions}
%%%%%%%%%%%%%%%%%%%%%%%%%%%%%%%%%%%%%%%%%%%%%%%%%%%%%%%%%%%%%%%%%%%%%%%%%%%%%

The semantics of source-selection programs we defined so far aims at selecting 
at most one source.  We can easily modify this definition, however, to accommodate 
also the selection of
multiple sources at a time.
To this end, we only have to modify the
structural rule (\ref{eq:structural}) in
Definition~\ref{def:evalprgm} appropriately.

For example, using language elements provided by the \DLV 
system~\cite{leon-etal-2002-dlv,fabe-etal-jelia04}, the 
simultaneous selection of up to a given number $k$ of
sources can be 
accomplished by replacing
(\ref{eq:structural}) with the following rules: 
\[
\begin{array}{r@{~}c@{~}l}
\mathit{false} &\la&  \naf\, \mathit{false},\, \query(Q),\, \mathit{max\_sources}(K), \\
 && \#\mathit{count}\{S': \qs(S',Q)\} >
K,\\[1ex]
\neg \qs(S,Q) &\la&  \source(S),\, \query(Q),\, \mathit{max\_sources}(K), \\ &&
1 <= \#\mathit{count}\{ S': \qs(S',Q)\} <=
K,  \\
&&  \naf\, \qs(S,Q), 
\end{array}
\]
where $\mathit{max\_sources}(K)$ holds for $K=k$. Here,
$\#\mathit{count}\{S': \qs(S',Q)\}$ is an \emph{aggregate expression} 
 which singles out the number of all
sources $S'$ for which 
an instance of 
$\qs(S',Q)$ is in the answer set, and ``$<$'' and ``$<=$'' are 
comparison built-ins.
This modification can also be
expressed with (ordinary) ELPs as introduced in Section~\ref{sec:elps}, 
but is more involved then.

The setting of selecting a ``best'' source with a single selection 
result can be easily generalized to a setting with multiple, ranked
selection results---in particular, to the computation of all outcomes
with a cost valuation within a given distance $d$ to a given value,
as well as to the computation of the $k$ best outcomes, for a given
integer $k$, akin to \emph{range queries} and \emph{$k$-nearest neighbor
queries}, respectively, in information retrieval. Such ranked computations can be
orthogonally combined with the type of selection outcome (\iec single
source vs.\ up to a number of sources).  Furthermore, they can be easily
accomplished using the features of the underlying {\DLV} system.

%%%%%%%%%%%%%%%%%%%%%%%%%%%%%%%%%%%%%%%%%%%%%%%%%%%%%%%%%%%%%%%%%%%%%%%%%%%%
\section{Implementation and application}\label{sec:appl}
%%%%%%%%%%%%%%%%%%%%%%%%%%%%%%%%%%%%%%%%%%%%%%%%%%%%%%%%%%%%%%%%%%%%%%%%%%%%

\subsection{Implementation}

We have implemented our source-selection approach 
on top of the {\DLV} system \cite{leon-etal-2002-dlv} and its
front-end {\PLP}~\cite{delg-etal-01} for prioritized logic
programs.\footnote{Details about \DLV\ and \PLP\ can also be 
found at \url{http://www.dlvsystem.com} and 
\url{http://www.cs.uni-potsdam.de/~torsten/plp}, respectively.} 
The evaluation of source-selection programs proceeds in
three steps: 
(i)~the set of all selection inputs for a query $Q$ is
computed from $R(Q)$, $\Pi_{qa}$, $\Pi_{sd}$, and $\Pi_{dom}$, using
{\DLV} (cf.~Theorem~\ref{thm:adequacy:2} in~\ref{app:pref-object}); 
(ii)~a call to {\DLV}
calculates the priority relation $<$ from the set of selection
inputs and $\Pi_{sel}$; and 
(iii)~the answer sets of $(\Pi_{Q},<)$
are determined by employing {\PLP} and {\DLV}. 
Note that this
three-step approach might appear to be overly complex, given that
computing a selection answer set is feasible in polynomial time with
an $\NP$ oracle (see Section~\ref{sec:properties} and
Theorem~\ref{cor:complexity} in \ref{app:pref-object}), 
and one might wonder why \DLV (which can
handle $\SigmaP{2}$-complete problems) is called several
times. 
The reason for proceeding in this fashion is that it actually 
greatly improves the performance since,
due to built-in optimization techniques of {\DLV}, groundings can be
kept smaller.

The entire process is implemented as an ECLiPSe Prolog program, which
served as a rapid prototyping language, and is independent of the
actual query language. For XML-QL queries, however, a query parser,
written in C{\small++}, for generating the low-level representation
$R(Q)$ of a query $Q$ has been developed. A query parser for SQL
queries is also available~\cite{schi-02}   
and further languages can be deployed in
the same way.

\begin{figure}[t!]
\centering
\includegraphics[scale=.85%width=13.78cm
]{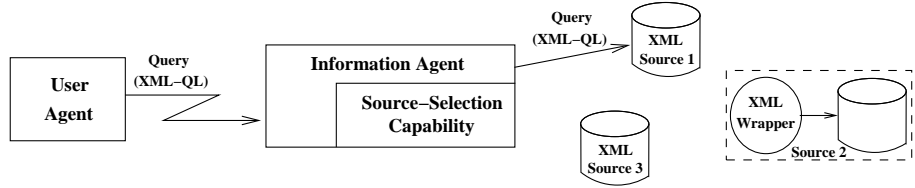} 
\caption{Architecture of a simple agent-based source-selection system
\label{fig:siteselection}}
\end{figure}

We have also ``agentized'' the source-selection system using the
{\impact} agent platform~\cite{subr-etal-00}, enabling the realization of
source-selection agents 
which may also issue the
execution of XML-QL queries on XML data sources. A generic agent-based
source-selection setup, as implemented in {\impact}, is shown in
Figure~\ref{fig:siteselection}.  Data are stored in XML databases, and
queries are posed in an XML query-language such as XML-QL. Some of the
databases may be wrapped from non-XML data sources.
A query is handed over to an information agent, which has to pick one
of several databases that comply with the same (universal) schema to
answer the query. 

The architecture in Figure~\ref{fig:siteselection} is 
only one of several possible agent-based architectures; others may be as follows:

\begin{itemize}
\item there may be
multiple information agents in a system, avoiding a centralization
bottleneck;
 
 \item the source-selection capability may be realized not in terms of a
special source-selection agent, but being part of a more powerful
mediator agent; or
 
 \item the sources may be accessed through specialized wrapper
agents, which control access and might refuse requests. 
\end{itemize}

%%%%%%%%%%%%%%%%%%%%%%%%%%%%%%%%%%%%%%%%%%%%%%%%%%%%%%%%%%%%%%%%%%%%%%%%%%%%
\subsection{An application for movie databases}
%%%%%%%%%%%%%%%%%%%%%%%%%%%%%%%%%%%%%%%%%%%%%%%%%%%%%%%%%%%%%%%%%%%%%%%%%%%%

As an application domain, we considered the area of movie
databases, and we have built an experimental environment for source
selection in this domain, using the prototype implementation described above. 

\subsubsection{Movie sources}

We used the \emph{Internet Movie Database {\rm (}IMDb\/{\rm )}} as 
the main source for raw data, as well as the
\emph{EachMovie Database} provided by Compaq Computer 
Corporation,\footnote{These two databases are available at 
\url{http://www.imdb.org} and 
\url{http://www.research.compaq.com/SRC/eachmovie}, respectively.}
to generate a suite of
XML movie databases. 
 To this end, (parts of) the large
databases were wrapped offline to XML, using a DTD (provided 
in~\ref{app:dtd}) which we modeled from a set of relevant  movie
concepts captured by the 
\emph{Open Directory Project}.\footnote{See \url{http://dmoz.org}.}
The XML databases we constructed are the following:

\begin{description}
\item[{\sc RandomMovies} {\rm (}{\sc RM}{\rm ):}] 
This source contains data about numerous movies, randomly wrapped from the 
IMDb. 
Besides title and language information (always having value 
``English''), each item comprises,  where available, entries 
containing genre classification, the release date, the running time, 
review ratings, the names of the two main actors, directors, and 
screenwriters, as well as details about the soundtrack (listing, 
in some cases, the name of the composer of the soundtrack).

\item[{\sc RandomPersons} {\rm (}{\sc RP}{\rm ):}]  Like {\sc RM}, 
{\sc RandomPersons} is derived from the IMDb, containing randomly 
wrapped data about numerous actors, directors, 
screenwriters, and some composers.
Besides names, person data comprise the date and country of birth, and a 
biography, and may, as for {\sc RM}, again be incomplete. 

\item[{\sc EachMovie} {\rm (}{\sc EM}{\rm ):}] Wrapped from Compaq's 
EachMovie Database, this source 
stores English movies plus ratings. For most entries, it provides genre
information, and for half of them a release date (after 1995). It 
has no information about actors, directors, soundtracks, etc., however.

\item[{\sc Hitchcock} {\rm (}{\sc HC}{\rm ):}] Wrapped from the IMDb, 
this source stores all
movies directed by Alfred Hitchcock, in the format of RandomMovies but
with all involved actors listed. For each person, it also contains
information (if available) about the date and country of birth, and a biography.

\item[{\sc KellyGrant} {\rm (}{\sc KG}{\rm ):}] Similar to {\sc HC}, 
this source stores the titles of all movies
in which either Grace Kelly or Cary Grant were actors, 
as well as the names of all persons involved. 
        
\item[{\sc Horror60} {\rm (}{\sc H60}{\rm ):}] Being the last of 
our databases, {\sc H60} is a collection
of horror movies from the 
1960s, as found in the IMDb. Movie and person data are as before, 
but almost no soundtrack or composer information is stored.
\end{description}

The information about these databases is stored in the 
source-description program $\Pi_{sd}$, using the predicates 
introduced in
Section~\ref{sec:site-descr}.
For illustration, we list some elements of this program, 
modeling one of the sources, 
and refer to \citeN{eite-etal-02d} and \citeN{fink-02} for 
a detailed account of the complete program $\Pi_{sd}$.

\begin{example} 
For providing information about  database {\sc KellyGrant}, 
the program $\Pi_{sd}$ contains the following facts:

\begin{tabbing}
X\= $\cov(s\_{\it KellyGrant,c,high}),$ \ \  \= \kill \>
$\source(s\_{\it KellyGrant});$ \+\\
$\siteup(s\_{\it KellyGrant});$\\
$\format(s\_{\it KellyGrant,xml});$\\
$\freq(s\_{\it KellyGrant,low});$ \\
$\spec(s\_{\it KellyGrant},{\it ``Kelly"});$ \\
$\spec(s\_{\it KellyGrant},{\it ``Grant"});$ \\
$\cov(s\_{\it KellyGrant},{\it ``Movie"},{\it low});$ \\
$\cov(s\_{\it KellyGrant,c,high}),$  
% \> 
for $c \in \{$\=${\it ``ReleaseDate"}\!,{\it 
``Person"}\!, {\it ``BirthDate"}\!$,\\
\> ${\it ``Actor"} \}$;\\
$\cov(s\_{\it KellyGrant,fifties,high});$ \\
$\cov(s\_{\it KellyGrant,sixties,high});$ \\
$\neg \relevant(s\_{\it KellyGrant},p),$ for $p \in \{ {\it seventies, eighties, 
nineties, twothousands} \}$. \-
\end{tabbing}
% }

Informally, $\prgm{sd}$ expresses that {\sc KellyGrant} is an 
XML source which is (currently) up 
and rarely updated. It is specialized in topics 
${\it ``Kelly"}$ and ${\it ``Grant"}$, and has high coverage 
about persons, especially 
actors, and their birth dates, but provides low coverage 
about movies in general. 
However, it highly covers the release dates of the stored 
movies, most of 
which are from the fifties and sixties.
Further information about {\sc KellyGrant} is derived 
from default rules like 
the ones given below, stating that English is the 
default language for all sources:

\[
\begin{array}{r@{~}c@{~}l} 
\sitelang(S,{\it ``English"}) & \la & \site(S), \\
&&                            \naf \neg \sitelang(S,{\it ``English"});\\[.8ex]
\neg \sitelang(S,{\it ``English"}) & \la &\sitelang(S,L),
                             L\neq {\it ``English"}.
\end{array}
\]

\end{example}

\subsubsection{Domain knowledge}
\label{sec:domain-knowledge}

The ontology part of the domain knowledge, $\Pi_{dom}$, includes the  
facts 
\[
\begin{array}{r@{}l}
\ob(O), \mbox{ for } O \in \{&{\it ``MovieDB"},\,{\it ``Movie"},\,
{\it ``Director"},\,{\it ``Actor"},\,{\it ``Screenwriter"}, %\,
\\
& {\it ``Composer"},\,{\it ``Person"},\,{\it ``Soundtrack"},\,{\it ``Review"}\},
\end{array}
\]
as they 
may be extracted from the XML DTD, and the
fact 
\[
\synonym({\it ``Personalia"},{\it ``Person"}).
\]
The attributes
of the concept ${\it ``Movie"}$ are given by the facts 
\[
\begin{array}{r@{}l}
\oattr({\it ``Movie",att}), 
\mbox{ where }
 {\it att} \in
\{& {\it title},{\it alternativeTitles},
{\it genre},{\it releaseDate},\\
& {\it  runningTime}, {\it language}, {\it review}\}.
\end{array}
\]
 For example, 
a concrete instance of 
${\it ``Movie"}$ is given by 
$\inst(m12,{\it ``Movie"})$. 
For further details, cf.\ \citeN{eite-etal-02d} or \citeN{fink-02}.

The background part of $\Pi_{dom}$ serves to formalize ``common-sense''
knowledge of the application domain, which is an important source of
information for the selection process.  This part is usually
quite extensive. On the one hand, it contains rules capturing typical
relationships between ontological concepts, and, on the other hand, it
comprises ``well-known'' instances of these concepts. For space reasons,
we only show a few rules of $\Pi_{dom}$ here. 
We note in passing
that
this part also implements a simple form of reasoning about time, 
viz.\ reasoning about \emph{decades}, 
by associating every year since 1920 its corresponding decade.

\begin{example} 
Some (typical) rules from the background knowledge are:  

\[
\begin{array}{rr@{~}c@{~}l} 
s_1: & \inst(P,{\it ``Director"}) &\la & {\it directed}(P,M); \\[.8ex]
s_2: & {\it involved}(P,M) &\la & {\it directed}(P,M); \\[.8ex]
s_3: & {\it life\_period}(P,B,E) & \la & \inst(P,{\it ``Person"}), \naf {\it dead}(P),\\
&&&  \attr(P,birthDate,B_1), {\it current\_year}(E),\\
&&&                             {\it calender\_year}(B_1,B); \\[.8ex]
s_4: & {\it possible\_genre}(M,G) & \la & {\it involved}(P,M),  
                             {\it default\_genre}(P,G), \\ 
&&&                       \naf {\it defined\_genre}(M).
\end{array}
\]

Intuitively, rules $s_1$ and $s_2$ infer, from a role ${\it acted}$
between a person and a movie, that the
corresponding person is an actor and that he or she is
involved in the movie. Rule $s_3$ assigns a life period to a
person from his or her birth date, while rule $s_4$ infers a possible genre for
a movie, if an involved person and his or her default genre are known.

Furthermore, the following facts are representations of specific movie-historic 
incidents (actually, they model information about Grace Kelly and 
the movie ``Arsenic and Old Lace''):
\[
\begin{array}{l}
\inst({\it perKelly},{\it ``Actor"});\\
\attr({\it perKelly,name,nameKelly}); \\
\attr({\it perKelly,birthDate},1929); \\
\attr({\it perKelly,dateOfDeath},1982); \\
{\it prod\_period}({\it perKelly},1945,1960); \\[.8ex]
\inst({\it nameKelly,name}); \\
\attr({\it nameKelly,firstName},{\it ``Grace"});
\\
\attr({\it nameKelly,firstName},{\it ``Patricia"}); \\
\attr({\it nameKelly,lastName},{\it ``Kelly"}); \\[.8ex]
\inst(m12,{\it ``Movie"}); \\
\attr(m12,{\it title},{\it ``Arsenic\ and\
Old\ Lace"}); \\
\attr(m12,{\it releaseDate},1944); \\[.8ex]
{\it acted}({\it perGrant},m12).
\end{array}
\]
\end{example}

%%%%%%%%%%%%%%%%%%%%%%%%%%%%%%%%%%%%%%%%%%%%%%%%%%%%%%%%%%%%%%%%%%%%%
\subsubsection{Source-selection program}\label{sec:movie-ss}
%%%%%%%%%%%%%%%%%%%%%%%%%%%%%%%%%%%%%%%%%%%%%%%%%%%%%%%%%%%%%%%%%%%%%

The experimental movie source-selection program fills several pages and
is too complex to be listed and discussed here in detail.
Therefore, similar as before, we only give an informal description, 
highlighting the most important aspects, and refer to
\citeN{eite-etal-02d} and \citeN{fink-02} for more details.

Among the source-selection rules, default rules have lowest priority
and are used only in the core part.  They make default suggestions for
query sources in case no other core source-selection rule is
eligible. Some examples are the following default rules:
\begin{tabbing} 
XX\=$r_5$: ~~ \= $\querysite(s\_{\it RandomPersons},Q)~$\= \' $\la~$\= \kill 
\>$r_1$: \>\> $\querysite(S,Q)$  \' $\la$~ $\defpath(O,P,Q),$   \+\\
\>\>\> $\occurs(O,V),\spec(S,P)$; \\[1ex]
$r_{2}$: \>\> $\querysite({\it s\_RandomMovies},Q)$  \' $\la$~ 
                             $\defobj(O,{\it ``Movie"},Q)$; \\[1ex]
$r_{3}$: \>\> $\querysite({\it s\_RandomPersons},Q)$  \' $\la$~ 
                             $\defobj(O,{\it ``Person"},Q)$.\- 
\end{tabbing}

The first rule is generic, whilst the others are specific.
Informally, $r_1$ advises to query source $S$ if it is specialized for
$P$, where $P$ is some path of a reference in the query that is compared
to some value. For example, suppose $P$ is instantiated with ${\it
LastName}$. If some source is specialized for last names, then it is
chosen unless a source-selection rule with higher priority
is applicable. Similarly, the specific rules $r_2$ and $r_3$ suggest to select {\sc RandomMovies}
or {\sc RandomPersons} if the query entails a reference
under object ${\it ``Movie"}$ or ${\it ``Person"}$, respectively.

Non-default core source-selection rules also appear in either generic or specific form:

\begin{tabbing} 
XX\=$r_{9}$: ~~ \= $\querysite(s\_Hitchcock,Q)~$\= \' $\la~$\= \kill 
\> $r_{4}$: \>\>$\querysite(S,Q)$ \' $\la$~  
                             $\site(S),$ 
                             $\query(Q),$ \+\\
                         \>\>\>    ${\it high\_coverage}(S,Q)$; \\[1ex]
$r_{5}$: \>\>$\querysite(S,Q)$ \' $\la$~ 
                             $\site(S),$
                             $\query(Q),$ 
                             ${\it special}(S,Q)$;\\[1ex]
$r_{6}$: \>\>$\querysite({\it s\_Hitchcock},Q)$ \' $\la$~ $\occ(O_1,{\it ``Person"},{\it ``LastName"},Q),$ \\
\>\>\>                        $\selects(O_1,equal,{\it ``Hitchcock"}),$ \\
                        \>\>\>  $\occ(O_2,{\it ``Person"},{\it ``FirstName"},Q),$ \\
\>\>\>                        $\selects(O_2,{\it equal},{\it ``Alfred"})$. \-
\end{tabbing}
The generic rules $r_{4}$ and 
$r_{5}$ suggest to query any source that highly covers the 
query or is special for it, respectively.
The specific rule $r_{6}$ advises to query the source {\sc Hitchcock} if a
query selects a person named  Alfred Hitchcock.  Note that ${\it
high\_coverage}$ and ${\it special}$ are {auxiliary} predicates,
defined by auxiliary rules (see below). 

Since no $\occ$ predicate and no default predicates occur in 
$r_{4}$ and $r_{5}$, there is no (direct) structural precedence
between them and rule $r_{6}$, as well as between 
$r_1$, $r_{2}$, and $r_{3}$.
The following user preferences explicitly establish preferences 
among them: 
\[
\begin{array}{r@{~}lr@{~}lr@{~}l} 
 r_1(\_,Q,\_,\_,\_) <_u & r_{4}(\_,Q);\qquad \quad   & r_{4}(\_,Q)  <_u & r_{5}(\_,Q); \\[.8ex]
r_1(\_,Q,\_,\_,\_) <_u & r_{5}(\_,Q); & r_{4}(\_,Q) <_u & r_{6}(Q,\_,\_);\\[.8ex]
r_{2}(Q,\_) <_u & r_{5}(\_,Q); & r_{5}(\_,Q) <_u & r_{6}(Q,\_,\_). \\[.8ex]
r_{3}(Q,\_) <_u & r_{5}(\_,Q);
\end{array}
\]

Auxiliary rules are used to define auxiliary predicates as well as to filter 
irrelevant sources:

\begin{tabbing} 
XX\=$r_9$: ~~ \= ${\it special_topic}(S,Q,T)~$ \= $\la~$ \= \kill 
\> $a_1$: \>\>${\it special}(S,Q)$\'$\la$ ${\it special\_topic}(S,Q,T)$; \+\\[.8ex]
   $a_2$: \>\> ${\it special\_topic}(S,Q,T)$\' $\la$ ${\it inferred\_topic}(Q,T),$ 
                             $\spec(S,T)$;\\[.8ex]
$a_3$: \>\> ${\it inferred\_topic}(S,Q,T)$\' $\la$ \= 
                            ${\it matchingMovie}(Q,M),{\it involved}(P,M),$  \\
                        \> \> \> $\attr(P,{\it name},N),\attr(N,{\it lastName},T)$;\\[.8ex]
$a_4$: \>\> $\neg \querysite(S,Q)$\' $\la$   
                            ${\it irrelevant}(S,Q)$; \\[.8ex]
$a_5 $: \>\> ${\it irrelevant}(S,Q)$\' $\la$ \= 
                             $\occ($\=$O,{\it ``MovieDB"},$\\
			     \>\>\>\>${\it ``Movie/ReleaseDate/Date"},Q),$ \\
\>\>\>                             $\selects(O,{\it equal},V),$
                        ${\it calender\_year}(V,Y),$\\
\>\>\> 
                            ${\it decade}(Y,D),\neg {\it relevant}(S,D)$. \-
\end{tabbing}
Informally, rule $a_1$ states that a source is special for a
query if a topic associated with the query exists for which it is
special.  Rule $a_2$ expresses that one way to associate a topic to a
query is to infer a topic, like, \egc realized in terms of rule $a_3$. Hence, if the query
accesses a movie that is known and $T$ is the last name of a person
involved in it, then a source is concluded to be special for that
query if it is specialized for $T$.  Rule $a_4$ states that a source
must not be queried if it is irrelevant for a query; in view of
rule $a_5$, this is the case if the source is not relevant for the
decade in which movie has been released.

Finally, the quantitative part of the source-selection program has weak 
constraints like the following:
\begin{tabbing} 
XX\= $w_1$:~~~~\= $\Leftarrow$~ \= \kill 
\> $w_1$:~~~~\> $\Leftarrow$~ $\querysite(S,Q), \defobj(O,T,Q),$  \+\\
\>\>                         $\constructs(O,C,P), \cov(S_1,T,high),$ 
                             $\naf \cov(S,T,high)$ [3:1]; \\[1ex]
$w_2$:~~~~\> $\Leftarrow$~ $\querysite(S,Q)$, ${\it high\_covered\_topic}(S_1,Q,T),$ \\
\>\>                          ${\it decade\_name}(T),$ $\naf {\it
                            high\_covered\_topic}(S,Q,T)$ [1:1]. \-
\end{tabbing}
Intuitively, $w_1$ assigns a penalty of $3$ per 
concept $T$ that is asked (resp., constructed) by the query to any answer set 
which selects a source that does not highly cover $T$ while a source 
highly covering $T$ exists.
Similarly, $w_2$ assigns a penalty of $1$ per decade 
that is associated to the query to any answer set which selects 
a source which does not highly cover this decade while some other 
highly covering source exists.

%%%%%%%%%%%%%%%%%%%%%%%%%%%%%%%%%%%%%%%%%%%%%%%%%%%%%%%%%%%%%%%%%%%%%%%%%%
\subsection{Experiments}\label{sec:exp}
%%%%%%%%%%%%%%%%%%%%%%%%%%%%%%%%%%%%%%%%%%%%%%%%%%%%%%%%%%%%%%%%%%%%%%%%%%

We tested the above movie-application scenario by means of a number 
of natural user queries. 
More specifically, our tests involved 18 queries, some of which are 
the following (for the complete list of queries, cf.\
 \citeN{eite-etal-02d} or \citeN{fink-02}):
\begin{itemize}
\item[$q_1$:] \emph{Which movies were directed by Alfred Hitchcock?}

\item[$q_2$:] \emph{In which movies, directed by Josef von Sternberg, did Marlene Dietrich act?}

\item[$q_3$:] \emph{In which year has the movie ``Arsenic and Old Lace'' been released?}

\item[$q_4$:] \emph{In which movies, directed by Alfred Hitchcock, did Marlene Dietrich act?}

\item[$q_5$:] \emph{In which film noirs did Marilyn Monroe act?}

\item[$q_6$:] \emph{In which movies did Laurel and Hardy act in 1940?}

\item[$q_7$:] \emph{Which movies where Frank Sinatra appeared in have 
a soundtrack composed by Elmer Bernstein?}

\item[$q_8$:] \emph{When was James Dean born?}
\end{itemize}
The formulation of these queries in 
XML-QL is
straightforward (as a matter of fact, $q_1$ is expressed by the 
XML-QL query of Example~\ref{exa:movie}; for the formulation of 
all queries in XML-QL, cf.\ \citeN{eite-etal-02d} or \citeN{fink-02}).

Source selection for the considered queries was performed employing the movie
databases described above as well as variants thereof. Each process took
from a couple to up to tens of seconds, which is due to the size of the
programs involved. However, performance was not a central issue here.
Since
our implementation and the used tools are unoptimized, there is a
large potential for performance improvements. Also, the
underlying solvers
% reasoning engines 
might gain efficiency in future releases.

\subsubsection{Results}

The results of the source selection process for $q_1$--$q_8$, using the above source
descriptions, are shown in Table~\ref{tab:results:original}. Note
that, by the semantics of source-selection programs, per selection answer set and
query, a single source is chosen. Thus, query decomposition is not
considered here, although our method for computing a query description allows for it
in principle. The entries show the sources which are
selected by the different answer sets, where the labels ``Candidates'' 
and ``Best'' refer to selection with optimization part dropped (i.e., 
qualitative selection only) and enabled, respectively.

%%%%%%%%%%%%%%%%%%%%%%%%%%%%%%%%%%%%%%%%%%%%%%%%%%%%table
\begin{table} 
\caption{\label{tab:results:original}
Experimental Results for the Movie Application} 
\begin{center}
\begin{tabular}{l@{\extracolsep\fill}*{8}{c@{~~}}} 
\hline \hline
\emph{Query} &$q_1$&$q_2$&$q_3$&$q_4$&$q_5$&$q_6$&$q_7$&$q_8$
\\
\hline 
\multicolumn{9}{c}{~}
\\[-1ex]
% \multicolumn{9}{c}{ (a) Original Selection Base}\\[1ex]
% \hline
\emph{Candi-} 
&{\sc HC}&{\sc RM}, {\sc H60},&{\sc KG}&{\sc HC}&{\sc RM}, {\sc HC},&{\sc RM}, {\sc HC},&{\sc RM}, {\sc H60},&{\sc RP}\\
\emph{dates}&&{\sc HC}, {\sc KG}&&&{\sc KG}&{\sc KG}&{\sc HC}, {\sc KG}&\\ \hline
\emph{Best}
&{\sc HC}&{\sc RM}&{\sc KG}&{\sc HC}&{\sc RM}&{\sc RM}&{\sc RM}&{\sc RP} \\ 
% \\[-1ex]
\hline\hline
\multicolumn{9}{c}{~}
\end{tabular}
\end{center}
\end{table}

The results can be informally explained as follows. 
For $q_1$, a specific core source-selection rule, $r_{6}$, which 
has highest 
preference, fires and {\sc HC} is chosen, as expected.

For $q_2$, there is some background knowledge about Marlene Dietrich,
but no source can be found as being special for this query, 
while generic default source-selection 
rules trigger for all sources. Nonetheless, {\sc RP} 
and {\sc EM} are recognized as being irrelevant for $q_2$ and 
eventually discarded: 
$q_2$ asks for (resp., ranges over) concepts these sources are 
not relevant for (viz.\ {\it ``Movie"} and 
{\it ``Person"}, respectively). The best source among the 
candidates {\sc RM}, {\sc H60}, {\sc HC}, and {\sc KG} is 
{\sc RM}, since it is the only one highly covering the concept asked for. 

For $q_3$, since ``Arsenic and Old Lace'' is in the background 
knowledge (cf.\ above), and since Cary Grant acted in it, we would expect {\sc KG} 
to be queried. 
Indeed, this is what actually 
happens. 
It is not a specific core source-selection rule that triggers 
the selection (Cary Grant does not explicitly appear in the query), 
rather Grant is 
inferred as a query topic from the background knowledge and, 
thus, the generic 
core selection rule suggesting to query {\sc KG} has highest priority. 

Query $q_4$ is a refinement of $q_1$; the same specific core source-selection 
rule, $r_{6}$, as for $q_1$ triggers.

Similar as for $q_2$, {\sc RM} is chosen for $q_5$, $q_6$, and 
$q_7$, but for the former two,
{\sc H60} is recognized as being irrelevant on different grounds: $q_5$
asks for film noirs, and so {\sc H60}, which contains horror movies, is
eliminated by reasoning over genre information, while $q_6$ involves movies
from 1940, and thus {\sc H60}, which contains only movies produced in the 1960s, is
excluded by reasoning over decades.

Finally, {\sc RP} is chosen for $q_8$, as expected: a specific default 
source-selection rule triggers for {\sc RP}, which has precedence over generic default 
rules that would trigger for other sources. 

\subsubsection{Results with modified selection bases}

In a slightly different scenario, {\sc RM} is designed to have high 
coverage about composers and western movies, too, and a new 
random movie source,
{\sc RandomMoviesNew} ({\sc RMN}), similar to {\sc RM}, but with less coverage about
genres, release dates, composers, and western movies, while having
high coverage about directors, dramas, and comedies, is introduced. 
Respective changes to the source descriptions and the addition of a specific default 
source-selection rule for {\sc RMN} (similar to rule $r_{2}$ for {\sc RM} 
in Section~\ref{sec:movie-ss}) and corresponding user 
preferences to the source-selection
program yield an ``extended'' selection base, for which the results for
$q_1$--$q_8$ are shown in Table~\ref{tab:results:extended}. 

%%%%%%%%%%%%%%%%%%%%%%%%%%%%%%%%%%%%%%%%%%%%%%%%%%table
\begin{table}[t] 
\caption{\label{tab:results:extended}
Experimental Results for an Extended Selection Base} 
\begin{center}
%\begin{tabular}{|l|@{\extracolsep\fill}*{8}{c@{~~}|}} 
\begin{tabular}{l@{\extracolsep\fill}*{8}{c@{~~}}} 
\hline \hline
\emph{Query} &$q_1$&$q_2$&$q_3$&$q_4$&$q_5$&$q_6$&$q_7$&$q_8$
\\
\hline 
\multicolumn{9}{c}{~}
\\[-1ex]
\emph{Candi-} 
&{\sc HC}&{\sc RM}, {\sc RMN},&{\sc KG}&{\sc HC}&{\sc RM}, {\sc RMN},&{\sc RM}, {\sc  RMN},&{\sc RM}, {\sc RMN},&{\sc RP} \\
\emph{dates}&&{\sc H60}, {\sc HC}, {\sc KG}&&&{\sc HC}, {\sc KG}&{\sc HC}, {\sc KG}&{\sc H60}, {\sc HC}, {\sc KG}&\\ \hline
\emph{Best}
&{\sc HC}&{\sc RMN}&{\sc KG}&{\sc HC}&{\sc RM}, {\sc RMN}&{\sc RM}, {\sc RMN}&{\sc RM}&{\sc RP}\\ 
\hline\hline
\multicolumn{9}{c}{~}
\end{tabular}
\end{center}
\end{table}

The change does not influence the results 
for $q_1$, $q_3$, $q_4$, and $q_8$.
This is intuitive, since
the suitability of the chosen sources 
is unaffected. For the other queries, the new source {\sc RMN} is a further
candidate, as the generic default source-selection rule is also
applicable to it. By a similar reason as before, {\sc RM} and {\sc RMN} are
better than the other candidates. For $q_2$, {\sc RMN} is ranked above {\sc RM}:
it highly covers dramas, which is an inferred query topic, since
drama is a default genre for Marlene Dietrich in the background
knowledge. {\sc RM} is ranked above {\sc RMN} for $q_7$ since {\sc RM} highly covers
composers, a concept occurring in the query (which asks for the composer
Elmer Bernstein). {\sc RM} and {\sc RMN} are ranked equal for $q_5$ and $q_6$:
they highly cover actors, but the background 
knowledge has no information about Laurel and Hardy; and that
Marilyn Monroe's default genre is comedy has no consequence for $q_5$,
as it explicitly asks for film noirs.

%%%%%%%%%%%%%%%%%%%%%%%%%%%%%%%%%%%%%%%%%%%%%%%%%%table
\begin{table}[t]
\caption{\label{tab:results:reduced}
Experimental Results for a Reduced Selection Base} 
\begin{center}
\begin{tabular}{l@{\extracolsep\fill}*{8}{c@{~~}}} 
\hline \hline
\emph{Query} &$q_1$&$q_2$&$q_3$&$q_4$&$q_5$&$q_6$&$q_7$&$q_8$
\\
\hline 
\multicolumn{9}{c}{~}
\\[-1ex]
\emph{Candi-} 
&{\sc HC}&{\sc H60}, {\sc HC},&{\sc KG}&{\sc HC}&{\sc HC}, {\sc KG},&{\sc HC}, {\sc KG}& {\sc H60}, {\sc HC},&{\sc RP}\\
\emph{dates}&&{\sc KG}&&&&&{\sc KG}&\\ \hline
\emph{Best}
&{\sc HC}&{\sc KG}&{\sc KG}&{\sc HC}&{\sc KG}&{\sc HC}, {\sc KG}&{\sc KG}&{\sc RP}\\ \hline\hline
\multicolumn{9}{c}{~}
\end{tabular}
\end{center}
\end{table}

As a further modification, we considered a ``reduced'' selection base where {\sc RM} is
down and, thus, cannot be queried. The results are given in
Table~\ref{tab:results:reduced}. The candidate sources remain the same,
except that {\sc RM} is missing; thus, the change has no impact on $q_1$,
$q_3$, $q_4$, and $q_8$, as one would expect. 
For $q_6$, the optimization part imposes no preference between {\sc HC}
and {\sc KG}; interestingly, it selects {\sc KG} as being best for $q_2$, $q_5$, and
$q_7$. This is because the background knowledge entails information
about the productive period of Dietrich, Monroe, and
Bernstein, 
which occur in the queries.  Thus, the
1950s and 1960s are inferred as being relevant topics for these queries,
and {\sc KG}, covering both decades highly, outranks {\sc HC} and {\sc H60}, which
highly cover only one of the decades each.

%%%%%%%%%%%%%%%%%%%%%%%%%%%%%%%%%%%%%%%%%%%%%%%%%%%%%%%%%%%%%%
\section{Related work}\label{sec:rel-conc}
%%%%%%%%%%%%%%%%%%%%%%%%%%%%%%%%%%%%%%%%%%%%%%%%%%%%%%%%%%%%%%

The selection of data sources is a component in many 
information-integration systems (cf., \egc
\citeN{aren-etal-93}, \citeN{baya-etal-97}, \citeN{garc-etal-97}, 
\citeN{cann-etal-97}, \citeN{gene-etal-97}, and \citeN{levy-etal-96b};
see also \citeN{levy-weld-00} and references therein).  However, most
center around mappings between a global scheme and local schemes, on
query rewriting, and on query planning to optimally reconstruct
dispersed information. Our work, instead, is concerned with {\em
qualitative} selection from different alternatives, based on rich
meta-knowledge and a formal semantics respecting preference and
context information involving heuristic defaults, which is not an
issue there. Furthermore, no form of query description similar as in
our method is considered in these approaches. 

In the following
subsection, we review some of the above mentioned information-integration systems
in more detail. 
Afterwards, we discuss approaches bearing a closer relation to our work.

\subsection{Information integration systems}

SIMS~\cite{aren-etal-93,aren-knob-92,aren-etal-96}, 
short for \emph{Services and Information
Management for Decision Systems},
is a data integration
system which exploits a semantic model of a problem domain to
integrate information dispersed over various heterogeneous information sources.
The latter
are typically databases, or, more generally, knowledge bases. The
domain model is formulated in the Loom knowledge-representation
language~\cite{macg-bate-87}, and comprises a declarative
description of the objects and activities possible in the specific
domain. SIMS aims at providing the user a transparent access to the
data, without being aware of the underlying heterogenous data
sources. It accepts user queries in the form of a description of a
class of objects about which information is required. Any such query
over the domain model is mapped to a query over the information sources,
 by translating the concepts of the domain to corresponding
concepts in the data models of the information sources; 
if a direct translation does not
exist, a query rewriting is performed, and, if needed, multiple
databases are accessed in a query plan. SIMS strives for singling out
optimal query plans, for which aspects such as costs of accessing the
different sources and combining the results returned are taken
into account. This is apparently different from the contributions of
our work, which is concerned with selecting a single information
source among a set of candidate sources. Furthermore, aspects of
incomplete information and nonmonotonic constructs to overcome it
were not addressed in SIMS, nor a method similar to query description.

The Carnot project at MCC~\cite{cann-etal-97,coll-etal-91,huhn-sing-92} was an early
effort to
provide a logically unifying view of enterprise-wide, distributed, and
possibly heterogeneous data. The Carnot system has a layered
architecture, whose top layer consists of semantic services providing
a suite of tools for enterprise modeling, model integration, data
cleaning, and knowledge discovery. The \emph{Model Integration and Semantics
Tool} (MIST) is used for creating mappings between local schemas and a
common ontology expressed in Cyc~\cite{cyc} or in a specific knowledge
representation language, which is done once at the time of
integration. Besides relational databases, also knowledge-based
systems (with an extensional part containing facts and an intensional 
part containing
rules) may be integrated and, moreover, play a mediator role between
applications and different databases. As an important feature, local
database schemas remain untouched, and queries to them are translated
to the global schema and back to (other) local schemas for data
retrieval. Similar to SIMS, Carnot aims at providing a uniform and
consistent view of heterogeneous data. A selection of information
sources for query answering, based on similar criteria and methods as
in our approach, is not evident.  

InfoSleuth~\cite{baya-etal-97,barg-etal-99,nodi-etal-03}, which has
its roots in Carnot, is an agent-based system for information
discovery and retrieval in a dynamic, open environment, broadening the
focus of database research to the challenge of the World-Wide Web. It
extends the capabilities of Carnot to an environment in which the
identities of the information sources need not be known at the time of
generating the mapping. In this approach, agents are the constituents
of the systems, whose knowledge and their relationships to each other
are described in an InfoSleuth ontology.  Decisions about user-query
decompositions are based on a domain ontology, which is selected
by the user and describes knowledge about the relationships of the
data stored in the sources that subscribe to the ontology. As for
selection of information sources, special broker agents provide, upon
request, information about which resource agents (i.e., information
sources behind them) should be accessed for specific information
sought. The broker performs a \emph{semantic matchmaking} of the user
request with the service descriptions of the provider agents  (which may be
viewed as an advanced yellow-pages service), aiming at ruling out, by
means of constraints (e.g., over the range of values, existing
attributes, etc.), all sources which will return a nil result. To this end,
it must reason over explicitly advertised information about agent
capabilities to determine which agent can provide the requested
services. The broker translates KIF statements into queries in the
LDL++ deductive database language, which are submitted to an LDL++
engine for evaluation. In this way, rule-based matching is
facilitated. Our approach differs significantly from InfoSleuth, and is
in fact to some extent complementary to it. Indeed, the
descriptions of constraints and other semantic criteria in InfoSleuth 
for selecting
an information source are at a very low level. Even if the LDL++
language, which can emulate non-stratified negation via choice rules,
is used for rule-based matchmaking, there is no special support
for dealing with contexts, user preferences, or 
optimization constructs as in our approach. Furthermore, it is not
evident that InfoSleuth agents are programmed using a declarative
language which provides similar functionalities for discriminating 
among different sources compliant with the constraints. Instead, our
    formalism
might be mapped to LDL++ by a suitable transformation and thus provide
a plug-in module for realizing semantically richer and refined
brokering in InfoSleuth with a well-defined, formal semantics and
provable properties. 
       
The Information Manifold~\cite{kirk-etal-95,levy-etal-95,levy-etal-96b}
is a system for
browsing and querying multiple networked information sources. Its
architecture is based on a rich domain model which enables the description
of properties of the information sources, such as their addresses,
the protocols used to access them, their structure, etc., using a
combination of the CLASSIC description logic~\cite{borg-etal-89}, Horn
rules, and integrity constraints. An external information source is
viewed as containing extensions of a collection of relations, on which
integrity constraints may be imposed, and which are semantically
mapped by rules to the relations in the global knowledge base. 
Information sources may be associated with topics, allowing to classify
the former along a hierarchy of topics in the domain model. 
This mechanism can be used
for deciding retrieval of a source for related queries.  Like in
SIMS, the user may pose queries in a high-level language on the global
schema, which are mapped to queries over the local sources. The
Information Manifold focuses on optimizing the execution of a user
query, accessing as few information sources as necessary, where
relevance is judged on criteria involving the (static) semantic
mapping, and on combining the results. However, no qualitative
selection similar to the one in our approach is made, and, in
particular, no user preferences or nonmonotonic rules (including
default contexts) can be expressed by constructs in the language.

Infomaster~\cite{gene-etal-97} provides integrated
access to multiple distributed heterogeneous information 
sources on the Internet,
which gives the illusion of a centralized, homogeneous information
system in a virtual schema. The system handles both structural 
and content translations to
resolve differences between multiple data sources and the multiple
applications for the collected data, where mappings between the
information sources and the global schema are described by rules and
constraints.  The user may pose queries on the virtual schema, which
are first translated to queries over base relations at the information
sources and then further rewritten  to queries over site relations,
which are views on the base relations, by applying logical abduction. 
The core of Infomaster is a facilitator that dynamically determines an
efficient way to answer the user's query employing as few sources as
necessary and harmonizes the heterogeneities among these
sources. However, like in the other information systems above, neither
rich meta-data about the quality of information sources is considered, nor
preferences or context information is used to heuristically
discriminate between optional choices.

\subsection{Other work}

More related to our approach than the methods in the previous 
subsection is the work
by~\citeN{scot-stei-95}, which outlines an interactive tool for
information specialists in query design. It relieves them from
searching through data-source specifications and can suggest sources
to determine trade-offs. However, no formal semantics or richer domain
theories, capable of handling incomplete and default information, is
presented.

Remotely related to our work are the investigations by \citeN{fuhr-97}, 
presenting a
decision-theoretic model for selecting data sources based on retrieval
cost and typical information-retrieval parameters. 

\citeN{goto-etal-01}
consider a problem setting related to ours, where source
descriptions include semantic knowledge about the source. 
In contrast to our work, however, 
a query is viewed merely as a set of terms, and a source description 
is a thesaurus automatically
constructed from the documents of the source. 
A further thesaurus, WordNet~\cite{wordnet}, 
is used for the source evaluation algorithm, 
which is based on the calculation of weighted similarity measures.
The main differences to our approach are that the 
selection method is not declarative and just numeric, semantic
knowledge is limited to a thesaurus, and no further background
knowledge, reasoning, or semantic query analysis is involved.

Semantic analysis of queries has been incorporated to document 
retrieval by 
\citeN{wendlandt-driscoll-91}. Starting from conventional 
information-retrieval methods
that accept natural-language queries against text collections and
calculate similarity measures for query keywords, semantic modeling
was introduced by trying to detect entity attributes and thematic
roles from the query to the effect of a modified similarity
computation. While richer ontological knowledge than thesauri is used,
source descriptions have no semantic knowledge. Again, the
approach is not declarative but numeric in nature, and neither rich domain
theories nor automated reasoning is involved.

FAQ FINDER~\cite{burke-etal-95} is a natural-language
question-answering system that uses files of frequently-asked
questions (FAQ) as its knowledge base. It uses standard 
information-retrieval methods to
narrow the search to one FAQ file and to calculate a term-vector
metric for the user's question and question/answer
pairs. Moreover, it uses a comparison of question types  in 
a taxonomy derived from the query, and a semantic similarity score
in question matching. The latter is calculated by passing through the
hypernym links, \iec is-a links, through WordNet.

Recent proposals for Web-based information retrieval built on 
ontology-based agents which search for, maintain, and mediate 
relevant information for a user or other 
agents are discussed by \citeN{luke-etal-97}, \citeN{sim-wong-01}, 
and \citeN{chen-soo-01}. 
More specifically,
\citeN{sim-wong-01}
describe a society of software agents where query-processing agents 
assist users in selecting Web pages.
They search for URLs using
search engines and ontological WordNet relations for 
query specialization or generalization to keep the number of located,
relevant URLs within given limits. An
architecture for ontology-based 
information-gathering agents appears also in the work of \citeN{chen-soo-01}, 
but here special domain search engines and Web documents are used as
well. Ontologies are represented in a usual object-oriented
language, and queries are partial instances of ontological concepts.

%%%%%%%%%%%%%%%%%%%%%%%%%%%%%%%%%%%%%%%%%%%%%%%%%%%%%%%%%%%%%%%%%%%%%%
\section{Conclusion}\label{sec:concl}
%%%%%%%%%%%%%%%%%%%%%%%%%%%%%%%%%%%%%%%%%%%%%%%%%%%%%%%%%%%%%%%%%%%%%%

In this paper, we have presented a knowledge-based approach for
information-source selection, using meta-knowledge about the
quality of the sources for determining a ``best'' information source
to answer a given query, which is posed in a formal query language (as
considered here, XML-QL).
We have described a rule-based language for expressing
source-selection policies in a fully declarative way, which supports
reasoning tasks that involve different components such as background
and ontological knowledge, source descriptions, and query
constituents. Furthermore, the language provides a number of features
which have proven valuable in the context of knowledge representation,
viz.\ the capability of dealing with incomplete information,
default rules, and preference information.

We have developed a novel method for automated query
analysis at a generic level in which interesting
information is distilled from a given query expression in a formal query
language, 
as well as an approach to preference handling in source
selection, which combines implicit rule priorities, given by the
context of rule applications, and explicit user preferences. 
As pointed out previously, context-based rule application is 
a different concept as inheritance-based reasoning---to the 
best of our knowledge, no similar
approach for handling default-context rules has been considered before.  
We presented a formal model-theoretic semantics of our approach, which is
based on the answer-set semantics of extended logic
programs. Furthermore, we analyzed semantical and computational
properties of our approach, where we showed that source-selection
programs possess desirable properties which intuitively should be
satisfied. We emphasize that for other, related approaches no similar
results are evident, since lacking a formal semantics makes them less
accessible to reason about their behavior.

The results that we have obtained in the implementation of the
experimental movie application are encouraging, and suggest several
directions for further work. 
One issue concerns the supply of rich
background and common-sense knowledge. The coupling with available
ontology and common-sense engines via suitable interfaces
is suggestive for this purpose.
Extensions of logic programs under the answer-set semantics allowing 
such a coupling have been realized, e.g., by \citeANP{eite-etal-04} 
\shortcite{eite-etal-04,eite-etal-05,eite-etal-2005}. 
Also, other recent efforts aim at mapping description logics 
underlying different
ontology languages to logic programs~\cite{Grosof2003,moti-etal-03,Swift2004}.

Another direction for further work involves the application of 
our results in the context of
information integration and query systems. They might be valuable for
enriching semantic brokering in open agent-based systems, but also
for more traditional closed systems in which information sources
must be manually registered. In particular, the advanced 
information-integration methods, employing extended logic 
programming tools, developed within the INFOMIX project is 
a natural candidate for incorporating a heuristic 
source-selection component.\footnote{See 
\url{http://sv.mat.unical.it/infomix/} for details about INFOMIX.}

Our results are also relevant for adaptive source 
selection which is customized, e.g., by user
profiles.
This subject is important for realizing personalized
information systems in a dynamic environment, which, 
to a large extent, involve user preferences and reasoning 
with incomplete
information and defaults, as well as dynamic updates of source
descriptions.

\subsubsection*{Acknowledgments}
We would like to thank the referees for their helpful and
constructive comments which helped improving the presentation of this paper. 
This work was partially supported by the Austrian Science Fund (FWF)
under grants P13871-INF and Z29-INF, as well as by
the European Commission 
under
projects FET-2001-37004 WASP, IST-2001-33570 INFOMIX, 
and
the
IST-2001-33123 CologNeT Network of Excellence.

%%%%%%%%%%%%%%%%%%%%%%%%%%%%%%%%%%%%%%%%%%%%%%%%%%%%%%%%%%%%%%%%%%%%%%%%
\appendix 
%%%%%%%%%%%%%%%%%%%%%%%%%%%%%%%%%%%%%%%%%%%%%%%%%%%%%%%%%%%%%%%%%%%%%%%%

%%%%%%%%%%%%%%%%%%%%%%%%%%%%%%%%%%%%%%%%%%%%%%%%%%%%%%%%%%%%%%%%%%%%%%%%
\section{The XML DTD for the movie databases}\label{app:dtd}
%%%%%%%%%%%%%%%%%%%%%%%%%%%%%%%%%%%%%%%%%%%%%%%%%%%%%%%%%%%%%%%%%%%%%%%%

{\small
\begin{alltt} 
\parskip=0.4ex
  <!ELEMENT MovieDB (Movie|Actor|Director|Screenwriter|
                     Composer|Person|Award|Filmfestival)*>
  <!ELEMENT Movie (Title,AlternativeTitle*,ReleaseDate?,
                   RunningTime?,Culture?,LeadingRole*,Role*,Actor*,
                   Director*,Screenwriter*,Soundtrack*,Review*,Award*)>
  <!ATTLIST Movie 
            Genre (Action|Animation|Classic|Comedy|CowboyWestern|
                   CultMovie|Documentary|Experimental|FilmNoir|
                   Horror|Romance|SciFiFantasy|Series|Silent|Travel|Other)
           #IMPLIED Language CDATA "English">
  <!ELEMENT Person (FirstName*,LastName,BirthDate?,Country?,Biography?)> 
  <!ATTLIST Person ID ID #REQUIRED Gender (male|female) #IMPLIED>
  <!ELEMENT Award (AwardTitle,Date,AwardType?,AwardCategory?)>
  <!ELEMENT Character (#PCDATA)>
  <!ELEMENT Filmfestival (#PCDATA)>
  <!ELEMENT Actor (Award*)>
  <!ATTLIST Actor Personalia IDREF #REQUIRED> 
  <!ELEMENT Director (Award*)>
  <!ATTLIST Director Personalia IDREF #REQUIRED> 
  <!ELEMENT Screenwriter (Award*)>
  <!ATTLIST Screenwriter Personalia IDREF #REQUIRED> 
  <!ELEMENT Composer (Award*)>
  <!ATTLIST Composer Personalia IDREF #REQUIRED> 
  <!ELEMENT Soundtrack (Title,Composer*,Award*)>
  <!ELEMENT Biography (#PCDATA)>
  <!ELEMENT AlternativeTitle (#PCDATA)>
  <!ELEMENT Title (#PCDATA)>
  <!ELEMENT FirstName (#PCDATA)>
  <!ELEMENT LastName (#PCDATA)>
  <!ELEMENT BirthDate (Date)>
  <!ELEMENT Date (#PCDATA)>
  <!ELEMENT Country (#PCDATA)>
  <!ELEMENT AwardCategory (#PCDATA)>
  <!ELEMENT AwardType (#PCDATA)>
  <!ELEMENT AwardTitle (#PCDATA)>
  <!ELEMENT ReleaseDate (Date)>
  <!ELEMENT RunningTime (#PCDATA)>
  <!ELEMENT LeadingRole (Character,Award*)>
  <!ATTLIST LeadingRole Actor IDREF #REQUIRED>
  <!ELEMENT Role (Character,Award*)>
  <!ATTLIST Role Actor IDREF #REQUIRED>
  <!ELEMENT Review (ReviewText,Rating?)>
  <!ELEMENT ReviewText (#PCDATA)>
  <!ELEMENT Rating (#PCDATA)>
  <!ELEMENT Culture (#PCDATA)>
\end{alltt}
}

%%%%%%%%%%%%%%%%%%%%%%%%%%%%%%%%%%%%%%%%%%%%%%%%%%%%%%%%%%%%%%%%%%%%%%%%
\section{Query description}\label{subsec:qd-preds}
%%%%%%%%%%%%%%%%%%%%%%%%%%%%%%%%%%%%%%%%%%%%%%%%%%%%%%%%%%%%%%%%%%%%%%%%

In what follows, we provide 
details about the query description
predicates and the query-analysis program. 

\subsection{Low-level predicates}\label{subsec:low-level-preds}

The query and its syntactic subqueries are named by constants (e.g., $q_1,
q_2$,\ldots). The facts $R(Q)$ are formed using the following predicates:
\begin{itemize}

\item $\subquery(Q',Q)$: $Q'$ is a structural subquery of query~$Q$
(possibly itself a subquery);
\item $\lowquerycand(Q)$: identifies the overall query;
\item $\source(S,Q)$: query $Q$ accesses source $S$;
\item $\dbname(S)$: source $S$ is a database;
\item $\whereRef(O,T,P,Q)$: an IRP $O$ references an item under element 
                            $T$ and remaining path $P$ in the where part of 
                            query $Q$;
\item $\subpath(O,T_1,P_1,T_2,P_2)$: the path $T_1/P_1$ is a direct subpath 
                                     of $T_2/P_2$ in the IRP~$O$;
\item $\whereRefCmp(O_1,R,O_2)$: the items of IRPs $O_1$ 
                                 and $O_2$ are compared using operator $R$;
\item $\whereCmp(O,R,V)$: the item of IRP $O$ is compared 
                          to value $V$ using opera\-tor~$R$;
\item $\consRef(O,T,P)$: the item of IRP $O$ is 
                         constructed under element $T$ and remaining path 
                         $P$ in the (answer) construction part of query $Q$.
\end{itemize}

$R(Q)$ must respect that query languages may allow for
nested queries. However, 
in a query expression, an outermost query as the ``root'' of nesting
should be identifiable, as well as structural (syntactic) subqueries
of it. They are described using $\lowquerycand$ and $\subquery$,
respectively.

Along an IRP, item references relative to a position are captured by
the $\whereRef$ predicate, and suffix inclusions for this IRP are
stored as $\subpath$ facts. The predicates $\whereRefCmp$ and
$\whereCmp$ mirror the comparison of two items and the comparison of
an item with a value, respectively. Items that occur in the
construction part of a query are also identified by an IRP and stored
using $\consRef$.

\begin{example}\label{example:query-representation} 
The low-level representation $R(Q)$ of the query in Example~\ref{exa:movie}
contains
\[
\begin{array}{l}
\mbox{$\subquery(q_2,q_1)$, $\lowquerycand(q_1)$, 
$\source(``{\it MovieDB}",q_2)$, and}\\[.2ex] 
\dbname(``{\it MovieDB}"),
\end{array}
\]
and, \egc for the third IRP, $o_3$, which references ``{\it LastName}'', the facts:
\begin{tabbing}
 X\= \kill \>
  $\whereRef(o_3,``{\it LastName}",``\ ",q_2)$; \+\\[.2ex]
 $\whereRef(o_3,``{\it Personalia}",``{\it LastName}",q_2)$;  \\[.2ex]
  $\whereRef(o_3,``{\it Director}",``{\it Personalia/LastName}",q_2)$;\\[.2ex]
  $\whereRef(o_3,``{\it Movie}",``{\it Director/Personalia/LastName}",q_2)$;\\[.2ex]
  $\whereRef(o_3,``{\it MovieDB}",``{\it 
 Movie/Director/Personalia/LastName}",q_2)$;\\%[1ex]
  $\subpath(o_3,``{\it LastName}",``\ ",``{\it Personalia}",``{\it LastName}")$;\\[.2ex]
  $\subpath(o_3,``{\it Personalia}",``{\it LastName}",``{\it Director}",``{\it 
 Personalia/LastName}")$;\\[.2ex]
  $\subpath(o_3,\,$\=$``{\it Director}",``{\it Personalia/LastName}",``{\it 
 Movie}"$,\\%[0.2ex]
 \>$``{\it Director/Personalia/LastName}")$;\\[.2ex]
  $\subpath(o_3,$\>$``Movie",``{\it Director/Personalia/LastName}",``{\it 
 MovieDB}",$\\%[0.2ex]
 \>$``{\it Movie/Director/Personalia/LastName}")$; \\[.2ex]
  $\whereCmp(o_3,{\it equal},``{\it Hitchcock}")$.\- 
 \end{tabbing}
\end{example}

The complete low-level representation $R(Q)$ of the query is given in~\ref{app:qrep}
(cf.\ also \citeN{eite-etal-02d} or \citeN{fink-02}).

%%%%%%%%%%%%%%%%%%%%%%%%%%%%%%%%%%%%%%%%%%%%%%%%%%%%%%%%%%%%%%%%%%%%%%%%
\subsection{High-level predicates}\label{subsec:high-level-preds}
%%%%%%%%%%%%%%%%%%%%%%%%%%%%%%%%%%%%%%%%%%%%%%%%%%%%%%%%%%%%%%%%%%%%%%%%

The following high-level description predicates are defined:

\begin{itemize}

\item ${\it query}(Q)$: identifies an ``independent'' (sub-)query 
$Q$ (\iec $Q$ is executable on some source),  which is,
moreover, not a purely syntactic subquery (\iec which is not 
embraced by a sourceless query $Q'$ merely restructuring the result 
of $Q$; for details, cf.\ the explanation of rules $qa_{8}$--$qa_{12}$ 
of $\prgm{qa}$ below);

\item $\occ(O,C,P,Q)$: states that $(C,P)$ is a CRP for $Q$ via 
IRP $O$ in the where-part of~$Q$;

\item ${\it occurs}(O,V)$: the value $V$ is associated with an IRP 
$O$ in the overall query;

\item $\selects(O,R,V)$: like $\it occurs$, but
details the association with a comparison operator~$R$; 

\item $\constructs(O,I,P)$: states that the item of IRP $O$,
by use of a variable, also appears in the construct-part of the global query,
as an item $I$ under path $P$ (which may be different from the path in
the where-part);

\item ${\it joins}(O_1,O_2,R)$: records (theta-)joins of 
(or within) queries between
IRPs $O_1$ and $O_2$ under comparison operator $R$.
\end{itemize}

\begin{example}\label{example:query-description}{
For the query in Example~\ref{exa:movie}, we have ${\it query}(q_1)$
but not ${\it query}(q_2)$, since the embracing query $q_1$ has no
source and merely structures the result of $q_2$. The following $\occ$
facts result from $o_1$ and $o_3$:
\[
\begin{array}{l}
\occ(o_1,``{\it MovieDB}",``{\it Movie/Title}",q_1); \\
\occ(o_1,``{\it Movie}",``{\it Title}",q_1);\\[0.2ex]
\occ(o_3,``{\it MovieDB}",``{\it Movie/Director/Personalia/LastName}",q_1);\\[0.2ex]
\occ(o_3,``{\it Movie}",``{\it Director/Personalia/LastName}",q_1); \\[0.2ex]
\occ(o_3,``{\it Director}",``{\it Personalia/LastName}",q_1); \\[0.2ex]
\occ(o_3,``{\it Person}",``{\it LastName}",q_1). 
\end{array}
\]

Here, $``{\it MovieDB}"$, $``{\it Movie}"$, $``{\it Director}"$, and $``{\it Person}"$ are
concepts given by the ontology, and $``{\it Personalia}"$ 
is known to be a
synonym of $``{\it Person}"$ (cf.\ Appendix~\ref{subsec:query-description}
for further discussion).

The fact ${\it occurs}(o_3,``{\it Hitchcock}")$ states that value $\it
Hitchcock$ is associated with $o_3$.  This is detailed by
$\selects(o_3,{\it equal},``{\it Hitchcock}")$, where $\it equal$
represents equality.  For the $\constructs$ predicate, the fact
$\constructs(o_1,``{\it Movie}",`` \ ")$ is included. There are no
${\it joins}$ facts since the query has no join.  The complete
high-level description is given in~\ref{app:qrep} (cf.\ also
\citeN{eite-etal-02d} or \citeN{fink-02})}.\end{example}

\subsection{Query-analysis program}\label{subsec:query-description}

The query-analysis program $\prgm{qa}$ is composed of the following 
groups of rules. The first rules enlarge the low-level
predicate $\subpath$ as follows:%
\footnote{In a clean separation of $R(Q)$ and the high-level
description, a fresh predicate would be in order here. However, it is
convenient and economic to re-use the predicate $\subpath$, as it is only enlarged.}

\newcounter{bctr}
\newcommand{\qr}{\addtocounter{bctr}{1}$qa_{\thebctr}:$}

\[
\begin{array}{rr@{~}c@{~}l} 
\mbox{\qr} & \subpath(O,T_1,P_1,T_3,P_3) &\la &\subpath(O,T_1,P_1,T_2,P_2),\\
&&&  \subpath(O,T_2,P_2,T_3,P_3);\\[.8ex]
\mbox{\qr} & \subpath(O,T,P_1,T_2,P_2) & \la & \subpath(O,L,P_1,T_2,P_2), \synonym(L,T);\\[.8ex]
\mbox{\qr} & \subpath(O,T_1,P_1,T,P_2) & \la & \subpath(O,T_1,P_1,L,P_2), \synonym(L,T).
\end{array}
\]

Rule $qa_1$ expresses transitivity for elements
occurring in paths, and $qa_2$ and $qa_3$ deal with synonyms, 
which is imported ontological knowledge;
$\synonym$ applies to all pairs of  synonymous element names (\egc names of 
IDREF attributes\footnote{If in a DTD an attribute is declared of type IDREF, 
this means that its value is the identifier of another element.}).

The following two rules define useful projections of low-level
predicates: 
\[
\begin{array}{rr@{~}c@{~}l} 
\mbox{\qr} & \hassource(Q) & \la &\source(\_,Q);\\[.8ex]
\mbox{\qr} & \issubquery(Q) & \la & \subquery(Q,\_).
\end{array}
\]

Using them, an auxiliary predicate $\querycand$ is defined 
for candidates which may satisfy the $\query$ predicate; these are
the overall query and subqueries having a database or a document as its source:
\[
\begin{array}{rr@{~}c@{~}l} 
\mbox{\qr} & \querycand(Q) & \la & \lowquerycand(Q);\\[.8ex]
\mbox{\qr} & \querycand(Q) & \la & \issubquery(Q), \source(Z,Q), \dbname(Z).
\end{array}
\]

Concerning the high-level predicates,
independent,
separate queries are specified by respecting the nesting structure:
\[
\begin{array}{rr@{~}c@{~}l}
\mbox{\qr} &  \query(Q) & \la & \topquery(Q,Q);\\[.8ex]
\mbox{\qr} & \topquery(Q,Q) & \la & \querycand(Q), \naf \issubquery(Q);\\[.8ex]
\mbox{\qr} & \topquery(Q,Q) & \la & \querycand(Q), \subquery(Q,S), \source(Z,S);\\[.8ex]
\mbox{\qr} & \topquery(S,Q) & \la & \subquery(S,Z), \querycand(S), \topquery(Z,Q),\\
&&& \naf \hassource(Z);\\[.8ex]
\mbox{\qr} & \topquery(S,Q) &\la & \subquery(S,Z), \naf \querycand(S),\\
&&& \topquery(Z,Q).
\end{array}
\]

Rule $qa_{8}$ expresses the property that a query is considered to 
be independent if it is the topmost
independent query of itself. This is the case if the query is
a candidate for a separate query and it is either the outermost query
(dealt with by Rule~$qa_{9}$) or a direct structural subquery of a query to a source
(expressed by Rule~$qa_{10}$).  Moreover, $qa_{10}$ intuitively states that a 
candidate  query
nested within another query is viewed as a separate query only if the
nesting was not for 
purely syntactic reasons, 
\iec it has its own source. In case of a purely syntactic subquery, or 
if
a nested query is not a candidate for a separate query, its topmost
independent query is the one of the embracing query, as taken care of 
$qa_{11}$ and $qa_{12}$, respectively.

The next rules define the remaining high-level description
predicates. The auxiliary predicate ${\it has\_constructs}$ guarantees
that at least one $\constructs$ fact is generated for each context
reference constructed in the query answer.
\[
\begin{array}{rr@{~}c@{~}l}
\mbox{\qr} & \occ(O,T,P,Q) & \la &\whereRef(O,T,P,S), \ob(T),\\
&&& \topquery(S,Q);\\[.8ex]
\mbox{\qr} & \occ(O,T,P,Q) & \la & \whereRef(O,L,P,S), \synonym(L,T),
         \\
&&& \ob(T), \topquery(S,Q); \\[.8ex]
\mbox{\qr} & \constructs(O,T,P) & \la & \consRef(O,T,P), \ob(T);\\[.8ex]
\mbox{\qr} & \constructs(O,T,P) & \la & \consRef(O,L,P), \synonym(L,T), \ob(T);\\[.8ex]
\mbox{\qr} & {\it has\_constructs}(O) & \la & \consRef(O,T,P), \ob(T);\\[.8ex]
\mbox{\qr} & {\it has\_constructs}(O) & \la & \consRef(O,L,P), \synonym(L,T), \ob(T);\\[.8ex]
\mbox{\qr} & \constructs(O,``\ ",``\ ") & \la & \consRef(O,\_,\_), \naf {\it has\_constructs}(O);\\[.8ex]
\mbox{\qr} & \occurs(O,V) & \la & \whereCmp(O,C,V);\\[.8ex]
\mbox{\qr} & \selects(O,C,V) & \la & \whereCmp(O,C,V);\\[.8ex]
\mbox{\qr} & \joins(O_1,O_2,C) & \la &\whereRefCmp(O_1,C,O_2).
\end{array}
\]

Note that some rules reference the ontology predicate $\ob$.
A fact $\ob(e)$ should exist in (or being entailed by) the domain ontology
for all elements $e$ that are considered to be concepts.

When queries are joined over CRPs, then some of 
the occurrence, selection, and construction information of one 
CRP is also valid for the other. Hence, we can build a form of
a closure over joined CRPs, which is expressed by the following rules:
\[
\begin{array}{rr@{~}c@{~}l}
\mbox{\qr} & \constructs(O_1,T,P) & \la &\joins(O_1,O_2,{\it equal}), \constructs(O_2,T,P);\\[.8ex]
\mbox{\qr} & \constructs(O_2,T,P) & \la & \joins(O_1,O_2,{\it equal}), \constructs(O_1,T,P);\\[.8ex]
\mbox{\qr} & \occurs(O_1,V) & \la & \joins(O_1,O_2,C), \occurs(O_2,V);\\[.8ex]
\mbox{\qr} & \occurs(O_2,V) & \la & \joins(O_1,O_2,C), \occurs(O_1,V);\\[.8ex]
\mbox{\qr} & \selects(O_1,C,V) & \la & \joins(O_1,O_2,{\it equal}), \selects(O_2,C,V);\\[.8ex]
\mbox{\qr} & \selects(O_2,C,V) & \la & \joins(O_1,O_2,{\it equal}), \selects(O_1,C,V);\\[.8ex]  
\mbox{\qr} & \selects(O_1,{\it notequal},V) & \la & \joins(O_1,O_2,{\it notequal}), \selects(O_2,{\it equal},V); \\[.8ex]
\mbox{\qr} & \selects(O_2,{\it notequal},V) & \la & \joins(O_1,O_2,{\it notequal}), \selects(O_1,{\it equal},V).
\end{array}
\]

We remark that, as easily seen, the rules of $\prgm{qa}$ form a
locally stratified logic
program, and thus $\mathit{Ont} \cup \prgm{qa}\cup R(Q)$ has a unique
answer set.

\begin{example}\label{example:query-abstraction}{\rm
Let us consider how the high-level fact {$\occ(o_3,``{\it
Person}",``{\it LastName}",q_1)$} is derived in $\prgm{qa}$, given 
$R(Q)$ of the query in Example~\ref{exa:movie}. 

Since {$\lowquerycand(q_1)$} is in $R(Q)$, we obtain, by $qa_6$,
{$\querycand(q_1)$}. Since the fact $\issubquery(q_1)$ is not
derivable, $qa_{10}$ yields $\topquery(q_1,q_1)$ (\iec stating that
$q_1$ is independent). Next, we can derive $\issubquery(q_2)$ by means of
$qa_5$, and thus $\querycand(q_2)$ in view of $qa_7$, given that $R(Q)$ includes 
the facts 
$\subquery(q_2,q_1)$, $\source(``{\it MovieDB}",q_2)$, and
$\dbname(``{\it MovieDB}")$. Since $q_1$ has no source (\iec
$\hassource(q_1)$ is not derivable), we can derive
$\topquery(q_2,q_1)$ from $qa_{9}$. The fact $\occ(o_3,\!``{\it
Person}",\!``{\it LastName}",q_1)$ is now derived by means of $qa_{14}$, making use of
$\whereRef(o_3,\!``{\it Personalia}",\!``{\it LastName}",q_2)$ from
$R(Q)$, together with the facts
$\synonym(``{\it Personalia}",\!``{\it
Person}")$ and $\ob(``{\it Person}")$ from the ontology, and the derived fact
$\topquery(q_2,q_1)$. Note that $\occ(o_3,\!``{\it
Personalia}",``{\it LastName}",q_1)$ is not derivable, as 
$\ob(``{\it Personalia}")\notin\mathit{Ont}$.  
}
\end{example}

%%%%%%%%%%%%%%%%%%%%%%%%%%%%%%%%%%%%%%%%%%%%%%%%%%%%%%%%%%%%%%%%%%%%%%%%%%%%%%
\section{Query-representation for Example~\ref{exa:movie}}\label{app:qrep}
%%%%%%%%%%%%%%%%%%%%%%%%%%%%%%%%%%%%%%%%%%%%%%%%%%%%%%%%%%%%%%%%%%%%%%%%%%%%%%

The low-level representation $R(Q)$ of the query in
Example~\ref{exa:movie} comprises the following facts:
{%\footnotesize
\begin{tabbing}
% \phantom{X}$\{$\= 
\phantom{X} \= 
 $\dbname(``{\it MovieDB}")$; \+\\[.2ex]
 $\lowquerycand(q_1)$;\\[.2ex]
 $\subquery(q_2,q_1)$;\\[0.2ex]
 $\source(``{\it MovieDB}",q_2)$;\\[0.2ex]
 $\whereRef(o_1,\!``{\it Title}",\!``\ ",q_2)$;\\[.2ex]
 $\whereRef(o_1,\!``{\it Movie}",\!``{\it Title}",q_2)$;\\[0.2ex]
 $\whereRef(o_1,\!``{\it MovieDB}",\!``{\it Movie/Title}",q_2)$;\\[0.2ex]
 $\subpath(o_1,\!``{\it Title}",\!``\ ",\!``{\it Movie}",\!``{\it Title}")$;\\[0.2ex] 
 $\subpath(o_1,\!``{\it Movie}",\!``{\it Title}",\!``{\it MovieDB}",\!``{\it Movie/Title}")$;\\[0.2ex]
 $\whereRef(o_2,\!``{\it FirstName}",\!``\ ",q_2)$;\\[0.2ex]
 $\whereRef(o_2,\!``{\it Personalia}",\!``{\it FirstName}",q_2)$;\\[0.2ex]
 $\whereRef(o_2,\!``{\it Director}",\!``{\it Personalia/FirstName}",q_2)$;\\[0.2ex]
 $\whereRef(o_2,\!``{\it Movie}",\!``{\it Director/Personalia/FirstName}",q_2)$;\\[0.2ex]
 $\whereRef(o_2,\!``{\it MovieDB}",\!``{\it Movie/Director/Personalia/FirstName}",q_2)$;\\[0.2ex]
 $\subpath(o_2,\!``{\it FirstName}",\!``\ ",\!``{\it Personalia}",\!``{\it FirstName}")$;\\[0.2ex]
 $\subpath(o_2,\!``{\it Personalia}",\!``{\it FirstName}",\!``{\it Director}",\!``{\it Personalia/FirstName}")$;\\[0.2ex]
 $\subpath(o_2,$\=$``{\it Director}",\!``{\it Personalia/FirstName}",\!``{\it Movie}",$ \\[.2ex]
 \>$``{\it Director/Personalia/FirstName}")$;\\[0.2ex]
 $\subpath(o_2,\!``{\it Movie}",\!``{\it Director/Personalia/FirstName}",``{\it MovieDB}",$\\[0.2ex]
   \>$``{\it Movie/Director/Personalia/FirstName}")$;\\[0.2ex]
 $\whereRef(o_3,\!``{\it LastName}",\!``\ ",q_2)$;\\[0.2ex]
 $\whereRef(o_3,\!``{\it Personalia}",\!``{\it LastName}",q_2)$;\\[0.2ex]
 $\whereRef(o_3,\!``{\it Director}",\!``{\it Personalia/LastName}",q_2)$;\\[0.2ex]
 $\whereRef(o_3,\!``{\it Movie}",\!``{\it Director/Personalia/LastName}",q_2)$;\\[0.2ex]
 $\whereRef(o_3,\!``{\it MovieDB}",\!``{\it Movie/Director/Personalia/LastName}",q_2)$;\\[0.2ex]
 $\subpath(o_3,\!``{\it LastName}",\!``\ ",\!``{\it Personalia}",\!``{\it LastName}")$;\\[0.2ex]
 $\subpath(o_3,\!``{\it Personalia}",\!``{\it LastName}",\!``{\it Director}",\!``{\it Personalia/LastName}")$;\\[0.2ex]
 $\subpath(o_3,\!``{\it Director}",\!``{\it Personalia/LastName}",\!``{\it Movie}",$\\[.2ex]
 \>$``{\it Director/Personalia/LastName}")$;\\[0.2ex]
 $\subpath(o_3,"Movie",\!``{\it Director/Personalia/LastName}"$;\\[0.2ex]
    \>$``{\it MovieDB}",\!``{\it Movie/Director/Personalia/LastName}")$;\\[0.2ex]
 $\whereCmp(o_2,{\it equal},\!``{\it Alfred}")$;\\[.2ex]
 $\whereCmp(o_3,{\it equal},\!``{\it Hitchcock}")$;\\[0.2ex]
 $\consRef(o_1,\!``{\it Movie}",\!``\ ")$;\\[.2ex]
 $\consRef(o_1,\!``{\it MovieList}",\!``{\it Movie}")$.\-
\end{tabbing}
}

The high-level description, except for auxiliary predicates and the 
completion of the subpath predicates, is given by the following facts:
\[
\begin{array}{l}
\query(q_1); \\[0.2ex]
\occ(o_1,\!``{\it MovieDB}",\!``{\it Movie/Title}",q_1);\\[0.2ex]
\occ(o_1,\!``{\it Movie}",\!``{\it Title}",q_1); \\[0.2ex]
\occ(o_2,\!``{\it MovieDB}",\!``{\it
Movie/Director/Personalia/FirstName}",q_1); \\[0.2ex]
\occ(o_2,\!``{\it Movie}",\!``{\it
Director/Personalia/FirstName}",q_1);
\\[0.2ex]
\occ(o_2,\!``{\it Director}",\!``{\it
Personalia/FirstName}",q_1);
\\[0.2ex]
\occ(o_2,\!``{\it Person}",\!``{\it FirstName}",q_1); \\[0.2ex]
\occ(o_3,\!``{\it MovieDB}",\!``{\it
Movie/Director/Personalia/LastName}",q_1); \\[0.2ex]
\occ(o_3,\!``{\it Movie}",\!``{\it
Director/Personalia/LastName}",q_1);\\[0.2ex]
\occ(o_3,\!``{\it Director}",\!``{\it Personalia/LastName}",q_1);
\\[0.2ex]
\occ(o_3,\!``{\it Person}",\!``{\it LastName}",q_1); \\[0.2ex]
\occurs(o_2,\!``{\it Alfred}"), \occurs(o_3,\!``{\it
Hitchcock}");\\[0.2ex]
\selects(o_2,{\it equal},\!``{\it Alfred}");\\[.2ex]
\selects(o_3,{\it
equal},\!``{\it Hitchcock}");\\[0.2ex]
\constructs(o_1,\!``{\it Movie}",\!``\ ").\-
\end{array}
\]

%%%%%%%%%%%%%%%%%%%%%%%%%%%%%%%%%%%%%%%%%%%%%%%%%%%%%%%%%%%%%%%%%%%%%%
\section{Further properties of source-selection programs}\label{app:pref-object}
%%%%%%%%%%%%%%%%%%%%%%%%%%%%%%%%%%%%%%%%%%%%%%%%%%%%%%%%%%%%%%%%%%%%%%

In order to realize the construction of $\mathcal{E}(\mathcal{S},Q)$
in terms of a single logic program, we introduce a set
$\nameset$ of constants serving as names for rules, and a new binary
predicate $\mathit{pref}(\cdot,\cdot)$, defined over 
$\nameset$, expressing preference between rules.
The extended vocabulary 
$\at_{\it sel}\cup\{\mathit{pref}(\cdot,\cdot)\}\cup \nameset$ 
is denoted by $\bar{\at}_{\it sel}$. 
We furthermore assume an injective function $\namef{\cdot}$ which
assigns to each rule 
$r\in\Pi_Q$ a name
$\namef{r}\in\nameset$.
To ease notation, we also write $\name{r}$ instead of
$\namef{r}$.
Finally, $\Lit_{\it pref}$ denotes the set of all literals having 
predicate symbol $\mathit{pref}$. 
Note that $\Lit_{\it pref}\cap\Lit_{\it sel}=\emptyset$.

\begin{theorem}\label{thm:adequacy:2}
Let ${\cal S} = \tuple{\prgm{qa},\prgm{dom},\prgm{sd},\prgm{sel},<_u}$
be a selection base, $Q$ a query, and 
${\mathcal{E}}({\mathcal{S}},Q)=(\Pi_Q,<)$.
Furthermore, let\/ $\Pi_{\cal S}(Q)
= \prgm{qa} \cup R(Q)\cup \prgm{dom} \cup \prgm{sd} \cup \Pi_{Q}$.
Then, there exists a logic program $\prgm{obj}(Q)$
over a vocabulary 
$\hat{\at}\supseteq \bar{\at}_{\it sel}$ 
such that every answer set $X$ of\/ $\Pi_{\cal S}(Q)\cup \prgm{obj}(Q)$
satisfies the following conditions:

\begin{enumerate}
\item\label{thm:adequacy:2:1} $X\cap\Lit_{\it pref}$ represents $<$,
i.e., $\mathit{pref}(n_r,n_{r'})\in X$ iff $r < r'$; and

\item\label{thm:adequacy:2:2} $X\cap\Lit_{\it sel}$ is 
a selection answer set of $(\prgm{sel},<_u)$ for $Q$ with respect to 
$\cal S$ iff $X$ is a preferred answer set of the prioritized 
program $(\Pi_{\cal S}(Q)\cup \prgm{obj}(Q),<)$.
\end{enumerate}
\end{theorem}

\begin{proof} 
We give a description of $\prgm{obj}(Q)$ but omit a detailed argument 
that it satisfies the desired properties. Informally, $\prgm{obj}(Q)$ 
consists of two parts: 
the first,  $\prgm{rel}$, is derived from
$\Pi_{\cal S}(Q)$
and takes care of computing the relevant rules for $Q$, by utilizing weak
constraints; the second part, $\prgm{pref}$, is a (locally) stratified 
logic program determining the relations of Definition~\ref{def:order}.
We start with the construction of $\prgm{rel}$.

For each predicate $p\in\at_{\it sel}$ 
and each $r\in\Pi_Q$, we introduce a new predicate 
$\extp{p}_r$ of the same arity as $p$.
In addition, we introduce a new atom ${\it rel}_r$, 
informally expressing that rule $r$ is relevant.
If $\alpha$ is either a literal, a set of literals, a rule, or a program, then
by $\ext{\alpha}_{{r}}$ we denote the result of uniformly replacing 
each atom $p(x_1,\commadots x_n)$ occurring in $\alpha$ by 
$\extp{p}_{{r}}(x_1\commadots x_n)$.

For each $r\in\Pi_Q$, we define a program $\Pi_r$ containing the following items:
\begin{enumerate}[\hspace{4.2ex}]
\item%[{\rm (}i\/{\rm )}]
each rule in $\ext{\Pi_{\it{qa}} \cup R(Q)\cup \Pi_{\it{sd}} \cup {\Pi}_{\it{dom}}}_r$;

\item%[{\rm (}ii\/{\rm )}]
the rule
\(
\it{rel}_{{r}}\la \ext{B^\dagger(r)}_r;
\)
and

\item%[{\rm (}iii\/{\rm )}]
the extended default-context rules
\[
\begin{array}{r@{~}c@{~}l}
\extp{\defobj}_{{r}}(O,C,Q)  &\la & \extp{\occ}_{{r}}(O,C,\_,Q),\\[1ex]
\extp{\defpath}_{{r}}(O,P,Q)  &\la & \extp{\occ}_{{r}}(O,\_,P,Q).
\end{array}
\]
\end{enumerate}

As easily checked, $r$ is relevant for $Q$ iff  
$\Pi_r$ has some answer set containing ${\it rel}_{r}$.
Now, $\prgm{rel}$ is defined as the collection of each 
of the programs $\Pi_r$, together with weak constraints of form 
\begin{equation}\label{eq:wc}
\Leftarrow \; \naf {\it{rel}}_{r} \;[1:m+1],
\end{equation}
for every $r\in\Pi_Q$, where $m$ is the maximal priority level 
of the weak constraints occurring in $\prgm{sel}^o$.
Since, for any $r_1,r_2\in\Pi_Q$ with $r_1\neq r_2$,
the programs $\Pi_{r_1}$ and $\Pi_{r_2}$ are
defined
over disjoint vocabularies, 
and given the inclusion of the weak constraints (\ref{eq:wc}) 
in $\prgm{rel}$, we obtain that $\prgm{rel}$ satisfies the following property:
\begin{description}
%\begin{itemize}
\item[{\rm (}$*${\rm )}]  for every answer set $X$ of $\prgm{rel}$ and 
every $r\in\Pi_Q$, $r$ is relevant for $Q$ iff $\it{rel}_{r}\in X$.
% \end{itemize}
\end{description}
These answer sets are used as inputs for the program $\prgm{pref}$, which is defined next.

Let $\it{pr}(n,m)$ and $\it{pr}'(n,m)$ be new binary predicates, where $n,m$ are names. 
Then, $\prgm{pref}$ consists of the following rules:

\begin{enumerate}[\hspace{4.2ex}]

\item ${\it pr}(\name{r_1},\name{r_2})\la{\it rel}_{r_1},{\it rel}_{r_2}$, 
for every  $r_1,r_2\in\Pi_Q$ such that either $r_1<_u r_2$ or $r_1,r_2$ 
satisfy Conditions~($O_1$) or ($O_2$) of Definition~\ref{def:order};

\item ${\it pr}(\name{r_1},\name{r_2})\la{\it rel}_{r_1},{\it 
rel}_{r_2},\subpath(o,t_1,p_1,t_2,p_2)$, for every $r_1,r_2\in\Pi_Q$ 
such that $\occ(o,t_1,p_1,q)\in\body{r_1}$ and $\occ(o,t_2,p_2,q)\in\body{r_2}$; and

\item the rules
\[
\begin{array}{r@{~}c@{~}l}
{\it pr}'(N_1, N_2) &  \la & {\it pr}(N_1, N_2), \\[1ex]
{\it pr}'(N_1, N_3) & \la & {\it pr}'(N_1, N_2), {\it pr}(N_2, N_3), \\[1ex]
{\it pref}(N_1, N_2)  & \la & {\it pr}(N_1, N_2), \naf {\it pr}'(N_2, N_1),\\[1ex]
{\it pref}(N_1, N_3) & \la & {\it pref}(N_1, N_2), {\it pref}(N_2, N_3).
\end{array}
\]

\end{enumerate}

Obviously, $\prgm{pref}$ is a (locally) stratified program.  Moreover, in view
of Condition~($*$), and since $\prgm{rel}$ is independent of
$\prgm{pref}$ and $\Pi_{\cal S}(Q)$ is independent of $\prgm{obj}(Q)$, for
every answer set $X$ of $\Pi_{\cal S}(Q)\cup \prgm{obj}(Q)=\Pi_{\cal S}(Q)\cup
\prgm{rel}\cup \prgm{pref}$, we have that (i)~%
${\it pr}(\name{r_1},\name{r_2})\in X$ iff $r_1\unlhd r_2$,
(ii)~${\it pr}'(\name{r_1},\name{r_2})\in X$ iff $r_1\unlhd^* r_2$, and
(iii)~${\it pref}(\name{r_1},\name{r_2})\in X$ iff $r_1< r_2$.
This proves Condition~\ref{thm:adequacy:2:1} of the theorem.

As for Condition~\ref{thm:adequacy:2:2}, consider some answer set $X$ of
$\Pi_{\cal S}(Q)\cup\prgm{obj}(Q)$.
Since $\Pi_{\cal S}(Q)$ is independent of $\prgm{obj}(Q)$, $X$ is of form 
$Y\cup Y'$, where $Y$ is an answer set of $\Pi_{\cal S}(Q)$ and $Y'$ is 
a set of ground literals disjoint from $\Lit_{\it sel}$.
Hence, $X\cap\Lit_{\it sel}=Y$.
According to Theorem~\ref{thm:adequacy}, $Y$ is a selection answer set 
of $(\prgm{sel},<_u)$ for $Q$ with respect to $\cal S$ iff $Y$ 
is a preferred answer set of $(\Pi_{\cal S}(Q),<)$.
But it is easily seen that the latter holds just in case 
$Y\cup Y'$ is a preferred answer set of $(\Pi_{\cal
S}\cup\prgm{obj}(Q),<)$. This proves the result.
\end{proof}

We note the following comments.
First, $\Pi_{rel}$ can be simplified by taking
independence of subprograms of $\Pi_{\cal S}(Q)$ and possible uniqueness
of answer sets for them into account. For example, if $\Pi_{sd}$ has a
unique answer set, then we may use in each program $\Pi_r$ simply
$\ext{\alpha}_{{r}} = \alpha$, for each literal over $\at_{\it sd}$. In
particular, if the program $\prgm{qa} \cup R(Q)\cup \prgm{dom} \cup
\prgm{sd}$ has a unique answer set (e.g., if this program is locally
stratified), then we may simply take as $\Pi_{rel}$ the program
$\prgm{qa} \cup R(Q)\cup \prgm{dom} \cup \prgm{sd}$ together with all
rules $\it{rel}_{{r}}\la B^\dagger(r)$, for $r\in
\Pi_Q$. 

Second, the program $\Pi_{\cal S}(Q) \cup \Pi_{obj}(Q)$ in
Theorem~\ref{thm:adequacy:2} represents, via preferred answer sets for
a dynamic rule preference given by the atoms over $\mathit{pref}$, the
selection answer sets of $(\Pi_{sel},<_u)$ for $Q$. It can be easily
adapted to a fixed program $\Pi'_{\cal S}$ such that, for any query
$Q$, the dynamic preferred answer sets of $\Pi'_{\cal S} \cup R(Q)$
represent the selection answer sets of $(\Pi_{sel},<_u)$ for $Q$ (cf.\
\citeN{delg-etal-02} for more details on dynamic preferences).

As for the complexity of source-selection programs, we can derive the
following result as a consequence of Theorem~\ref{thm:adequacy}. 

\begin{theorem}
\label{cor:complexity}
Given a query $Q$ and the
grounding 
% $\ground{\Pi_{\cal S}(Q)}$ 
of $\Pi_{\cal S}(Q)=
\prgm{qa} \cup R(Q)\cup \prgm{dom} \cup \prgm{sd} \cup \Pi_{Q}$, 
for a selection base ${\cal S} = \tuple{\prgm{qa},\prgm{dom},\prgm{sd},\prgm{sel},<_u}$, 
deciding
whether $(\prgm{sel},<_u)$ has some selection answer set for $Q$ with
respect to $\cal S$ is \NP-complete. Furthermore, computing any such
selection answer set is complete for {\rm FP}$^{\rm NP}$.
\end{theorem}

\begin{proof}
Obviously, the groundings of the programs $\prgm{rel}$ and $\prgm{pref}$ in the
proof of Theorem~\ref{thm:adequacy:2} are constructible in polynomial
time from $Q$ and the grounding of $\Pi_{\cal S}(Q)$, and so is the ground program
$\Pi'$, consisting of the groundings of $\Pi_{\cal S}(Q)$, $\prgm{rel}$, and $\prgm{pref}$. 
Furthermore, the preferred answer sets of $(\Pi',<)$ correspond to 
the selection answer sets of $(\prgm{sel},<_u)$. Since deciding
whether a prioritized logic program (with no weak constraints) has a preferred
answer set is \NP-complete~\cite{delg-etal-02}, it follows that deciding
whether $(\prgm{sel},<_u)$ has a selection answer set for $Q$ with respect to
${\cal S}$ is in \NP.
Note that the presence of weak constraints has no influence on the 
worst-case complexity of deciding the existence of (preferred) answer sets.
Moreover,  \NP-hardness is immediate
since the auxiliary rules can form any standard logic program.

From any answer set $X$ of $\Pi'$, an answer set of $(\prgm{sel},<_u)$
is easily computed. Computing such an $X$ is feasible in polynomial
time with an \NP\ oracle, sketched as follows. First, compute the 
minimum vector of weak-constraint
violations, $v^*$, i.e., the sum of weights of violated constraints at each
level, using the oracle, performing binary search at each level, asking
whether a violation limit can be obeyed. Then,
build atom by atom an answer
set $X$ whose violation cost matches $v^*$ using the \NP\
oracle. Overall, this is possible in polynomial time with an \NP\
oracle, hence the problem is in {\rm FP}$^{\rm NP}$. 

The hardness for {\rm FP}$^{\rm NP}$ follows from a reduction given by 
\citeN{bucc-etal-99b}, which shows how the lexicographic maximum truth
assignment to a SAT instance, whose computation is well-known to be
complete for  {\rm FP}$^{\rm NP}$~\cite{kren-88}, can be encoded 
in terms of the answer set of an
ELP with weak constraints.
\end{proof}

Note that under \emph{data complexity}, 
i.e., where the selection base
$\cal S$ is fixed while the query $Q$ (given by the facts
$R(Q)$) may vary, the problems in Theorem~\ref{cor:complexity} are
in $\NP$ resp.\ {\rm FP}$^{\rm NP}$, since the grounding 
of $\Pi_{\cal S}(Q)$ is polynomial in the size of $\cal S$ and $Q$
in this case. If, moreover, the size of $Q$ is small and bounded by a
constant, then the problems are solvable in {\em polynomial time},
since then the number of rules in the grounding of $\Pi_{\cal S}(Q)$
is bounded by some constant as well.

\bibliographystyle{acmtrans}

\makebbltrue

\ifmakebbl

\bibliography{kbiss}

\begin{thebibliography}{}

\bibitem[\protect\citeauthoryear{Abiteboul, Buneman, and Suciu}{Abiteboul
  et~al\mbox{.}}{2000}]{abit-etal-00}
{\sc Abiteboul, S.}, {\sc Buneman, P.}, {\sc and} {\sc Suciu, D.} 2000.
\newblock {\em {Data on the Web: From Relations to Semistructured Data and
  XML}}.
\newblock {Morgan Kaufmann}, {Los Altos}.

\bibitem[\protect\citeauthoryear{Alferes, Pereira, Przymusinska, and
  Przymusinski}{Alferes et~al\mbox{.}}{2002}]{alfe-etal-02}
{\sc Alferes, J.}, {\sc Pereira, L.}, {\sc Przymusinska, H.}, {\sc and} {\sc
  Przymusinski, T.} 2002.
\newblock {LUPS - A Language for Updating Logic Programs}.
\newblock {\em Artificial Intelligence\/}~{\em 138,\/}~1--2, 87--116.

\bibitem[\protect\citeauthoryear{Apt, Blair, and Walker}{Apt
  et~al\mbox{.}}{1988}]{apt-etal-88}
{\sc Apt, K.}, {\sc Blair, H.}, {\sc and} {\sc Walker, A.} 1988.
\newblock {Towards a Theory of Declarative Knowledge}.
\newblock See \citeN{mink-88}, 89--148.

\bibitem[\protect\citeauthoryear{Arens, Chee, Hsu, and Knoblock}{Arens
  et~al\mbox{.}}{1993}]{aren-etal-93}
{\sc Arens, Y.}, {\sc Chee, C.}, {\sc Hsu, C.}, {\sc and} {\sc Knoblock, C.}
  1993.
\newblock {Retrieving and Integrating Data from Multiple Information Sources}.
\newblock {\em International\ Journal\ of Cooperative Information
  Systems\/}~{\em 2,\/}~2, 127--158.

\bibitem[\protect\citeauthoryear{Arens and Knoblock}{Arens and
  Knoblock}{1992}]{aren-knob-92}
{\sc Arens, Y.} {\sc and} {\sc Knoblock, C.} 1992.
\newblock {Planning and Reformulating Queries for Semanti\-cally-Modeled
  Multidatabase Systems}.
\newblock In {\em {Proceedings of the First International Conference on
  Information and Knowledge Managements}}. 92--101.

\bibitem[\protect\citeauthoryear{Arens, Knoblock, and Shen}{Arens
  et~al\mbox{.}}{1996}]{aren-etal-96}
{\sc Arens, Y.}, {\sc Knoblock, C.}, {\sc and} {\sc Shen, W.} 1996.
\newblock {Query Reformulation for Dynamic Information Integration}.
\newblock {\em Journal\ of Intelligent Information Systems\/}~{\em 6,\/}~2--3,
  99--130.

\bibitem[\protect\citeauthoryear{Baral}{Baral}{2003}]{bara-03}
{\sc Baral, C.} 2003.
\newblock {\em {Knowledge Representation, Reasoning and Declarative Problem
  Solving with Answer Sets}}.
\newblock Cambridge University Press.

\bibitem[\protect\citeauthoryear{Bayardo, Bohrer, Brice, Cichocki, Fowler,
  Helal, Kashyap, Ksiezyk, Martin, Nodine, Rashid, Rusinkiewicz, Shea,
  Unnikrishnan, Unruh, and Woelk}{Bayardo et~al\mbox{.}}{1997}]{baya-etal-97}
{\sc Bayardo, R.}, {\sc Bohrer, B.}, {\sc Brice, R.}, {\sc Cichocki, A.}, {\sc
  Fowler, J.}, {\sc Helal, A.}, {\sc Kashyap, V.}, {\sc Ksiezyk, T.}, {\sc
  Martin, G.}, {\sc Nodine, M.}, {\sc Rashid, M.}, {\sc Rusinkiewicz, M.}, {\sc
  Shea, R.}, {\sc Unnikrishnan, C.}, {\sc Unruh, A.}, {\sc and} {\sc Woelk, D.}
  1997.
\newblock {InfoSleuth: Semantic Integration of Information in Open and Dynamic
  Environments (Experience Paper)}.
\newblock In {\em {Proceedings of the ACM SIGMOD International Conference on
  Management of Data {\rm (}SIGMOD '97\/{\rm )}}}. 195--206.

\bibitem[\protect\citeauthoryear{Borgida, Brachman, McGuinness, and
  Resnick}{Borgida et~al\mbox{.}}{1989}]{borg-etal-89}
{\sc Borgida, A.}, {\sc Brachman, R.~J.}, {\sc McGuinness, D.~L.}, {\sc and}
  {\sc Resnick, L.~A.} 1989.
\newblock {CLASSIC: A Structural Data Model for Objects}.
\newblock In {\em {Proceedings of the ACM SIGMOD International Conference on
  Management of Data {\rm (}SIGMOD '89\/{\rm )}}}, {J.~Clifford}, {B.~G.
  Lindsay}, {and} {D.~Maier}, Eds. {ACM Press}, 58--67.

\bibitem[\protect\citeauthoryear{Brewka and Eiter}{Brewka and
  Eiter}{1999}]{brew-eite-99}
{\sc Brewka, G.} {\sc and} {\sc Eiter, T.} 1999.
\newblock {Preferred Answer Sets for Extended Logic Programs}.
\newblock {\em Artificial Intelligence\/}~{\em 109,\/}~1--2, 297--356.

\bibitem[\protect\citeauthoryear{Buccafurri, Leone, and Rullo}{Buccafurri
  et~al\mbox{.}}{1996}]{bucc-etal-96}
{\sc Buccafurri, F.}, {\sc Leone, N.}, {\sc and} {\sc Rullo, P.} 1996.
\newblock {Stable Models and their Computation for Logic Programming with
  Inheritance and True Negation}.
\newblock {\em Journal\ of Logic Programming\/}~{\em 27,\/}~1, 5--43.

\bibitem[\protect\citeauthoryear{Buccafurri, Leone, and Rullo}{Buccafurri
  et~al\mbox{.}}{2000}]{bucc-etal-99b}
{\sc Buccafurri, F.}, {\sc Leone, N.}, {\sc and} {\sc Rullo, P.} 2000.
\newblock {Enhancing Disjunctive Datalog by Constraints}.
\newblock {\em IEEE Transactions\ on Knowledge and Data Engineering\/}~{\em
  12,\/}~5, 845--860.

\bibitem[\protect\citeauthoryear{Burke, Hammond, and Kozlovsky}{Burke
  et~al\mbox{.}}{1995}]{burke-etal-95}
{\sc Burke, R.}, {\sc Hammond, K.}, {\sc and} {\sc Kozlovsky, J.} 1995.
\newblock {Knowledge-Based Information Retrieval from Semi-Structured Text}.
\newblock In {\em Working Notes of the AAAI '95 Fall Symposium, Series on AI
  Applications in Knowledge Navigation and Retrieval, Cambridge, MA}. 19--24.

\bibitem[\protect\citeauthoryear{Chen and Soo}{Chen and
  Soo}{2001}]{chen-soo-01}
{\sc Chen, Y.-J.} {\sc and} {\sc Soo, V.-W.} 2001.
\newblock {Ontology-Based Information Gathering Agents}.
\newblock In {\em Proceedings of the First Asia-Pacific Conference on Web
  Intelligence {\rm (}WI 2001\/{\rm )}}, {{N.~Zhong et al.}}, Ed. LNCS,
  subseries LNAI, vol. 2198. Springer, 423--427.

\bibitem[\protect\citeauthoryear{Collet, Huhns, and Shen}{Collet
  et~al\mbox{.}}{1991}]{coll-etal-91}
{\sc Collet, C.}, {\sc Huhns, M.}, {\sc and} {\sc Shen, W.-M.} 1991.
\newblock {Resource Integration using a Large Knowledge Base in Carnot}.
\newblock {\em IEEE Computer\/}~{\em 24,\/}~12, 55--62.

\bibitem[\protect\citeauthoryear{Decker, Sycara, and Williamson}{Decker
  et~al\mbox{.}}{1997}]{deck-etal-97}
{\sc Decker, K.}, {\sc Sycara, K.}, {\sc and} {\sc Williamson, M.} 1997.
\newblock {Middle-Agents for the Internet}.
\newblock In {\em {Proceedings of the Fifteenth International Joint Conference
  on Artificial Intelligence {\rm (}IJCAI '97\/{\rm )}}}. Vol.~1. {Morgan
  Kaufmann}, 578--583.

\bibitem[\protect\citeauthoryear{Delgrande and Schaub}{Delgrande and
  Schaub}{1994}]{delg-scha-94}
{\sc Delgrande, J.} {\sc and} {\sc Schaub, T.} 1994.
\newblock {A General Approach to Specificity in Default Reasoning}.
\newblock In {\em Proceedings of the Fourth International Conference on
  Principles of Knowledge Representation and Reasoning {\rm (}KR '94{\rm )}}.
  146--157.

\bibitem[\protect\citeauthoryear{Delgrande, Schaub, and Tompits}{Delgrande
  et~al\mbox{.}}{2001}]{delg-etal-01}
{\sc Delgrande, J.}, {\sc Schaub, T.}, {\sc and} {\sc Tompits, H.} 2001.
\newblock {plp: A Generic Compiler for Ordered Logic Programs}.
\newblock In {\em {Proceedings of the Sixth International Conference on Logic
  Programming and Nonmonotonic Reasoning {\rm (}LPNMR 2001\/{\rm )}}},
  {T.~Eiter}, {W.~Faber}, {and} {M.~Truszczy{\'n}ski}, Eds. LNCS, subseries
  LNAI, vol. 2173. {Springer}, 411--415.

\bibitem[\protect\citeauthoryear{Delgrande, Schaub, and Tompits}{Delgrande
  et~al\mbox{.}}{2003}]{delg-etal-02}
{\sc Delgrande, J.~P.}, {\sc Schaub, T.}, {\sc and} {\sc Tompits, H.} 2003.
\newblock {A Framework for Compiling Preferences in Logic Programs}.
\newblock {\em {Theory and Practice of Logic Programming}\/}~{\em 3,\/}~2,
  129--187.

\bibitem[\protect\citeauthoryear{Deutsch, Fernandez, Florescu, Levy, and
  Suciu}{Deutsch et~al\mbox{.}}{1999}]{deut-etal-99}
{\sc Deutsch, A.}, {\sc Fernandez, M.}, {\sc Florescu, D.}, {\sc Levy, A.},
  {\sc and} {\sc Suciu, D.} 1999.
\newblock {A}~{Q}uery {L}anguage for {XML}.
\newblock {\em Computer Networks\/}~{\em 31,\/}~11--16, 1155--1169.

\bibitem[\protect\citeauthoryear{Dimopoulos and Kakas}{Dimopoulos and
  Kakas}{2001}]{dimo-kaka}
{\sc Dimopoulos, Y.} {\sc and} {\sc Kakas, A.} 2001.
\newblock {Information Integration and Computational Logic}.
\newblock {\em {C}omputational {L}ogic, {S}pecial {I}ssue on the {F}uture
  {T}echnological {R}oadmap of {C}ompulog-{N}et\/}, 105--135.

\bibitem[\protect\citeauthoryear{Eiter, Fink, Sabbatini, and Tompits}{Eiter
  et~al\mbox{.}}{2002a}]{eite-etal-01j}
{\sc Eiter, T.}, {\sc Fink, M.}, {\sc Sabbatini, G.}, {\sc and} {\sc Tompits,
  H.} 2002a.
\newblock {On Properties of Update Sequences Based on Causal Rejection}.
\newblock {\em {Theory and Practice of Logic Programming}\/}~{\em 2,\/}~6,
  721--777.

\bibitem[\protect\citeauthoryear{Eiter, Fink, Sabbatini, and Tompits}{Eiter
  et~al\mbox{.}}{2002b}]{eite-etal-00j}
{\sc Eiter, T.}, {\sc Fink, M.}, {\sc Sabbatini, G.}, {\sc and} {\sc Tompits,
  H.} 2002b.
\newblock {Using Methods of Declarative Logic Programming for Intelligent
  Information Agents}.
\newblock {\em {Theory and Practice of Logic Programming}\/}~{\em 2,\/}~6,
  645--719.

\bibitem[\protect\citeauthoryear{Eiter, Fink, and Tompits}{Eiter
  et~al\mbox{.}}{2003}]{eite-etal-02d}
{\sc Eiter, T.}, {\sc Fink, M.}, {\sc and} {\sc Tompits, H.} 2003.
\newblock {A Knowledge-Based Approach for Selecting Information Sources}.
\newblock Tech. Rep. INFSYS RR-1843-03-14, 2003, Institut f{\"u}r
  Informationssysteme, Technische Universit{\"a}t Wien.

\bibitem[\protect\citeauthoryear{Eiter, Gottlob, and Mannila}{Eiter
  et~al\mbox{.}}{1997}]{Eiter:1997:DD}
{\sc Eiter, T.}, {\sc Gottlob, G.}, {\sc and} {\sc Mannila, H.} 1997.
\newblock Disjunctive {Datalog}.
\newblock {\em ACM Transactions on Database Systems\/}~{\em 22,\/}~3, 364--418.

\bibitem[\protect\citeauthoryear{Eiter, Ianni, Schindlauer, and Tompits}{Eiter
  et~al\mbox{.}}{2005a}]{eite-etal-2005}
{\sc Eiter, T.}, {\sc Ianni, G.}, {\sc Schindlauer, R.}, {\sc and} {\sc
  Tompits, H.} 2005a.
\newblock {A Uniform Integration of Higher-Order Reasoning and External
  Evaluations in Answer-Set Programming}.
\newblock In {\em Proceedings of the Nineteenth International Joint Conference
  on Artificial Intelligence {\rm (}IJCAI 2005\/{\rm )}}. Morgan Kaufmann.

\bibitem[\protect\citeauthoryear{Eiter, Ianni, Schindlauer, and Tompits}{Eiter
  et~al\mbox{.}}{2005b}]{eite-etal-05}
{\sc Eiter, T.}, {\sc Ianni, G.}, {\sc Schindlauer, R.}, {\sc and} {\sc
  Tompits, H.} 2005b.
\newblock {Nonmonotonic Description Logic Programs: Implementation and
  Experiments}.
\newblock In {\em Proceedings of the Twelfth International Conference on Logic
  for Programming, Artificial Intelligence and Reasoning {\rm (}LPAR 2004\/{\rm
  )}}, {F.~Baader} {and} {A.~Voronkov}, Eds. LNCS, vol. 3452. Springer,
  511--517.

\bibitem[\protect\citeauthoryear{Eiter, Lukasiewicz, Schindlauer, and
  Tompits}{Eiter et~al\mbox{.}}{2004}]{eite-etal-04}
{\sc Eiter, T.}, {\sc Lukasiewicz, T.}, {\sc Schindlauer, R.}, {\sc and} {\sc
  Tompits, H.} 2004.
\newblock {Combining Answer-Set Programming with Description Logics for the
  Semantic Web}.
\newblock In {\em Proceedings of the Ninth International Conference on
  Principles of Knowledge Representation and Reasoning {\rm (}KR 2004{\rm )}},
  {D.~Dubois}, {C.~Welty}, {and} {M.-A. Williams}, Eds. Morgan Kaufmann,
  141--151.

\bibitem[\protect\citeauthoryear{Faber, Leone, and Pfeifer}{Faber
  et~al\mbox{.}}{2004}]{fabe-etal-jelia04}
{\sc Faber, W.}, {\sc Leone, N.}, {\sc and} {\sc Pfeifer, G.} 2004.
\newblock {Recursive Aggregates in Disjunctive Logic Programs: {S}emantics and
  Complexity}.
\newblock In {\em Proceedings of the Ninth European Conference on Logics in
  Artificial Intelligence {\rm (}JELIA 2004\/{\rm )}}, {J.~J. Alferes} {and}
  {J.~A. Leite}, Eds. LNCS, subseries LNAI, vol. 3229. Springer, 200--212.

\bibitem[\protect\citeauthoryear{Fellbaum}{Fellbaum}{1998}]{wordnet}
{\sc Fellbaum, C.} 1998.
\newblock {\em {WordNet: An Electronic Lexical Database}}.
\newblock MIT Press.

\bibitem[\protect\citeauthoryear{Fink}{Fink}{2002}]{fink-02}
{\sc Fink, M.} 2002.
\newblock {Declarative Logic-Programming Components for Information Agents}.
\newblock Ph.D. thesis, {Institut f{\"u}r Informationssysteme, Technische
  Universit\"{a}t Wien}, Austria.

\bibitem[\protect\citeauthoryear{Fowler, Perry, Nodine, and Bargmeyer}{Fowler
  et~al\mbox{.}}{1999}]{barg-etal-99}
{\sc Fowler, J.}, {\sc Perry, B.}, {\sc Nodine, M.~H.}, {\sc and} {\sc
  Bargmeyer, B.} 1999.
\newblock {Agent-Based Semantic Interoperability in {InfoSleuth}}.
\newblock {\em {SIGMOD Record}\/}~{\em 28,\/}~1, 60--67.

\bibitem[\protect\citeauthoryear{Fuhr}{Fuhr}{1999}]{fuhr-97}
{\sc Fuhr, N.} 1999.
\newblock {A Decision-Theoretic Approach to Database Selection in Networked
  IR}.
\newblock {\em ACM Transactions on Information Systems\/}~{\em 17,\/}~3,
  229--249.

\bibitem[\protect\citeauthoryear{Garcia-Molina, Papakonstantinou, Quass,
  Rajaraman, Sagiv, Ullman, Vassalos, and Widom}{Garcia-Molina
  et~al\mbox{.}}{1997}]{garc-etal-97}
{\sc Garcia-Molina, H.}, {\sc Papakonstantinou, Y.}, {\sc Quass, D.}, {\sc
  Rajaraman, A.}, {\sc Sagiv, Y.}, {\sc Ullman, J.}, {\sc Vassalos, V.}, {\sc
  and} {\sc Widom, J.} 1997.
\newblock {The {TSIMMIS} Approach to Mediation: Data Models and Languages}.
\newblock {\em Journal\ of Intelligent Information Systems\/}~{\em 8,\/}~2,
  117--132.

\bibitem[\protect\citeauthoryear{Geerts and Vermeir}{Geerts and
  Vermeir}{1993}]{geer-verm-93}
{\sc Geerts, P.} {\sc and} {\sc Vermeir, D.} 1993.
\newblock {A Nonmonotonic Reasoning Formalism using Implicit Specificity
  Information}.
\newblock In {\em Proceedings of the Second International Workshop on Logic
  Programming and Nonmonotonic Reasoning {\rm (}LPNMR '93{\rm )}}, {L.-M.
  Pereira} {and} {A.~Nerode}, Eds. LNCS, subseries LNAI. Springer, 380--396.

\bibitem[\protect\citeauthoryear{Geerts and Vermeir}{Geerts and
  Vermeir}{1995}]{geer-verm-95}
{\sc Geerts, P.} {\sc and} {\sc Vermeir, D.} 1995.
\newblock {Specificity by Default}.
\newblock In {\em Proceedings of the European Conference on Symbolic and
  Quantitative Approaches to Reasoning and Uncertainty {\rm (}ECSQARU '95\/{\rm
  )}}. LNCS, subseries LNAI, vol. 946. Springer, 207--216.

\bibitem[\protect\citeauthoryear{Gelfond and Lifschitz}{Gelfond and
  Lifschitz}{1991}]{gelf-lifs-91}
{\sc Gelfond, M.} {\sc and} {\sc Lifschitz, V.} 1991.
\newblock {Classical Negation in Logic Programs and Disjunctive Databases}.
\newblock {\em New Generation Computing\/}~{\em 9,\/}~3--4, 365--386.

\bibitem[\protect\citeauthoryear{Genesereth, Keller, and Duschka}{Genesereth
  et~al\mbox{.}}{1997}]{gene-etal-97}
{\sc Genesereth, M.}, {\sc Keller, A.}, {\sc and} {\sc Duschka, O.} 1997.
\newblock {Infomaster: An Information Integration System}.
\newblock In {\em {Proceedings of the ACM SIGMOD International Conference on
  Management of Data {\rm (}SIGMOD '97\/{\rm )}}}, {J.~Peckham}, Ed. ACM Press,
  539--542.

\bibitem[\protect\citeauthoryear{Goto, Ozono, and Shintani}{Goto
  et~al\mbox{.}}{2001}]{goto-etal-01}
{\sc Goto, S.}, {\sc Ozono, T.}, {\sc and} {\sc Shintani, T.} 2001.
\newblock {A Method for Information Source Selection using Thesaurus for
  Distributed Information Retrieval}.
\newblock In {\em Proceedings of the Pacific Asian Conference on Intelligent
  Systems 2001 {\rm (}PAIS 2001\/{\rm )}}. 272--277.

\bibitem[\protect\citeauthoryear{Grosof, Horrocks, Volz, and Decker}{Grosof
  et~al\mbox{.}}{2003}]{Grosof2003}
{\sc Grosof, B.~N.}, {\sc Horrocks, I.}, {\sc Volz, R.}, {\sc and} {\sc Decker,
  S.} 2003.
\newblock {Description Logic Programs: {C}ombining Logic Programs with
  Description Logics}.
\newblock In {\em Proceedings of the Twelfth International World Wide Web
  Conference {\rm (}WWW 2003\/{\rm )}}. ACM Press, 48--57.

\bibitem[\protect\citeauthoryear{Huffman and Steier}{Huffman and
  Steier}{1995}]{scot-stei-95}
{\sc Huffman, S.~B.} {\sc and} {\sc Steier, D.} 1995.
\newblock {A Navigation Assistant for Data Source Selection and Integration}.
\newblock In {\em {Working Notes of the AAAI '95 Fall Symposium Series on AI
  Applications in Knowledge Navigation and Retrieval, Cambridge, MA}}. AAAI
  Press, 72--77.

\bibitem[\protect\citeauthoryear{Huhns and Singh}{Huhns and
  Singh}{1992}]{huhn-sing-92}
{\sc Huhns, M.} {\sc and} {\sc Singh, M.} 1992.
\newblock {The Semantic Integration of Information Models}.
\newblock In {\em {Proceedings of the AAAI Workshop on Cooperation among
  Heterogeneous Intelligent Agents}}.

\bibitem[\protect\citeauthoryear{Inoue and Sakama}{Inoue and
  Sakama}{2000}]{inou-saka-00b}
{\sc Inoue, K.} {\sc and} {\sc Sakama, C.} 2000.
\newblock {Prioritized Logic Programming and Its Applications to Commonsense
  Reasoning}.
\newblock {\em Artificial Intelligence\/}~{\em 123,\/}~1--2, 185--222.

\bibitem[\protect\citeauthoryear{Kirk, Levy, Sagiv, and Srivastava}{Kirk
  et~al\mbox{.}}{1995}]{kirk-etal-95}
{\sc Kirk, T.}, {\sc Levy, A.}, {\sc Sagiv, Y.}, {\sc and} {\sc Srivastava, D.}
  1995.
\newblock {The Information Manifold}.
\newblock In {\em Proceedings of the AAAI 2001 Spring Symposium on Information
  Gathering in Distributed Heterogeneous Environments}. AAAI Press, 85--91.

\bibitem[\protect\citeauthoryear{Kowalski and Sadri}{Kowalski and
  Sadri}{1990}]{kowa-sadr-90}
{\sc Kowalski, R.~A.} {\sc and} {\sc Sadri, F.} 1990.
\newblock {Logic Programs with Exceptions}.
\newblock In {\em {Proceedings of the Seventh International Conference on Logic
  Programming {\rm (}ICLP '90\/{\rm )}}}. MIT Press, 598--616.

\bibitem[\protect\citeauthoryear{Krentel}{Krentel}{1988}]{kren-88}
{\sc Krentel, M.} 1988.
\newblock {The Complexity of Optimization Problems}.
\newblock {\em Journal\ of Computer and System Sciences\/}~{\em 36}, 490--509.

\bibitem[\protect\citeauthoryear{Laenens and Vermeir}{Laenens and
  Vermeir}{1990}]{laen-verm-90a}
{\sc Laenens, E.} {\sc and} {\sc Vermeir, D.} 1990.
\newblock {A Logical Basis for Object-Oriented Programming}.
\newblock In {\em Proceedings of the Second European Workshop on Logics in
  Artificial Intelligence {\rm (}JELIA '90\/{\rm )}}. LNCS, subseries LNAI.
  Springer, 317--332.

\bibitem[\protect\citeauthoryear{Lenat and Guha}{Lenat and Guha}{1990}]{cyc}
{\sc Lenat, D.~B.} {\sc and} {\sc Guha, R.~V.} 1990.
\newblock {\em {Building Large Knowledge-Based Systems: Representation and
  Inference in the Cyc Project}}.
\newblock Addison-Wesley.

\bibitem[\protect\citeauthoryear{Leone, Pfeifer, Faber, Eiter, Gottlob, Perri,
  and Scarcello}{Leone et~al\mbox{.}}{2006}]{leon-etal-2002-dlv}
{\sc Leone, N.}, {\sc Pfeifer, G.}, {\sc Faber, W.}, {\sc Eiter, T.}, {\sc
  Gottlob, G.}, {\sc Perri, S.}, {\sc and} {\sc Scarcello, F.} 2006.
\newblock {The {DLV} System for Knowledge Representation and Reasoning}.
\newblock {\em {ACM Transactions on Computational Logic}\/}.
\newblock To appear.

\bibitem[\protect\citeauthoryear{Levy, Rajaraman, and Ordille}{Levy
  et~al\mbox{.}}{1996}]{levy-etal-96b}
{\sc Levy, A.}, {\sc Rajaraman, A.}, {\sc and} {\sc Ordille, J.} 1996.
\newblock {Querying Heterogeneous Information Sources using Source
  Descriptions}.
\newblock In {\em {Proceedings of the Twentysecond International Conference on
  Very Large Data Bases {\rm (}VLDB '96\/{\rm )}}}, {T.~Vijayaraman},
  {A.~Buchmann}, {C.~Mohan}, {and} {N.~Sarda}, Eds. {Morgan Kaufmann},
  251--262.

\bibitem[\protect\citeauthoryear{Levy, Srivastava, and Kirk}{Levy
  et~al\mbox{.}}{1995}]{levy-etal-95}
{\sc Levy, A.}, {\sc Srivastava, D.}, {\sc and} {\sc Kirk, T.} 1995.
\newblock {Data Model and Query Evaluation in Global Information Systems}.
\newblock {\em Journal\ of Intelligent Information Systems\/}~{\em 5,\/}~2,
  121--143.

\bibitem[\protect\citeauthoryear{Levy and Weld}{Levy and
  Weld}{2000}]{levy-weld-00}
{\sc Levy, A.} {\sc and} {\sc Weld, D.} 2000.
\newblock {Intelligent Internet Systems}.
\newblock {\em Artificial Intelligence\/}~{\em 118,\/}~1--2, 1--14.

\bibitem[\protect\citeauthoryear{Lifschitz and Turner}{Lifschitz and
  Turner}{1994}]{lifs-turn-94}
{\sc Lifschitz, V.} {\sc and} {\sc Turner, H.} 1994.
\newblock {Splitting a Logic Program}.
\newblock In {\em Proceedings of the Eleventh International Conference on Logic
  Programming {\rm (}ICLP '94\/{\rm )}}. MIT Press, 23--38.

\bibitem[\protect\citeauthoryear{Luke, Spector, Rager, and Hendler}{Luke
  et~al\mbox{.}}{1997}]{luke-etal-97}
{\sc Luke, S.}, {\sc Spector, L.}, {\sc Rager, D.}, {\sc and} {\sc Hendler, J.}
  1997.
\newblock {Ontology-Based Web Agents}.
\newblock In {\em Proceedings of the First International Conference on
  Autonomous Agents {\rm (}Agents '97\/{\rm )}}, {{W.L.~Johnson}}, Ed. 59--66.

\bibitem[\protect\citeauthoryear{MacGregor and Bates}{MacGregor and
  Bates}{1987}]{macg-bate-87}
{\sc MacGregor, R.} {\sc and} {\sc Bates, R.} 1987.
\newblock {The LOOM Knowledge Representation Language}.
\newblock Tech. Rep. RS-87-188, Information Sciences Institute, University of
  Southern California.
\newblock Project Web page \url{http://www.isi.edu/isd/LOOM/}.

\bibitem[\protect\citeauthoryear{Minker}{Minker}{1988}]{mink-88}
{\sc Minker, J.}, Ed. 1988.
\newblock {\em {Foundations of Deductive Databases and Logic Programming}}.
\newblock Morgan Kaufman, Washington DC.

\bibitem[\protect\citeauthoryear{Motik, Volz, and Maedche}{Motik
  et~al\mbox{.}}{2003}]{moti-etal-03}
{\sc Motik, B.}, {\sc Volz, R.}, {\sc and} {\sc Maedche, A.} 2003.
\newblock {Optimizing Query Answering in Description Logics using Disjunctive
  Deductive Databases}.
\newblock In {\em Proceedings of the Tenth International Workshop on Knowledge
  Representation meets Databases {\rm (}KRDB 2003{\rm )}}, {F.~Bry}, {C.~Lutz},
  {U.~Sattler}, {and} {M.~Schoop}, Eds. CEUR Workshop Proceedings, vol.~79.
  RWTH Aachen University, 39--50.
\newblock
  \url{http://sunsite.informatik.rwth-aachen.de/Publications/CEUR-WS/Vol-79/}.

\bibitem[\protect\citeauthoryear{Nodine, Ngu, Cassandra, and Bohrer}{Nodine
  et~al\mbox{.}}{2003}]{nodi-etal-03}
{\sc Nodine, M.}, {\sc Ngu, A.}, {\sc Cassandra, A.}, {\sc and} {\sc Bohrer,
  W.} 2003.
\newblock {Scalable Semantic Brokering over Dynamic Heterogeneous Data Sources
  in {InfoSleuth}}.
\newblock {\em IEEE Transactions\ on Knowledge and Data Engineering\/}~{\em
  15,\/}~5, 1082--1098.

\bibitem[\protect\citeauthoryear{Przymusinski}{Przymusinski}{1988}]{przy-88}
{\sc Przymusinski, T.~C.} 1988.
\newblock {On the Declarative Semantics of Deductive Databases and Logic
  Programs}.
\newblock See \citeN{mink-88}, 193--216.

\bibitem[\protect\citeauthoryear{Sadri and Toni}{Sadri and
  Toni}{2000}]{sadr-toni-99+}
{\sc Sadri, F.} {\sc and} {\sc Toni, F.} 2000.
\newblock {Computational Logic and Multi-Agent Systems: A Roadmap}.
\newblock {\em {C}omputational {L}ogic, {S}pecial {I}ssue on the {F}uture
  {T}echnological {R}oadmap of {C}ompulog-{N}et\/}, 1--31.

\bibitem[\protect\citeauthoryear{Schindlauer}{Schindlauer}{2002}]{schi-02}
{\sc Schindlauer, R.} 2002.
\newblock {Representation of SQL Queries for Declarative Query Analysis}.
\newblock M.S.\ thesis, {Institut f{\"u}r Informationssysteme, Technische
  Universit\"{a}t Wien}, Austria.

\bibitem[\protect\citeauthoryear{Sim and Wong}{Sim and
  Wong}{2001}]{sim-wong-01}
{\sc Sim, K.~M.} {\sc and} {\sc Wong, P.~T.} 2001.
\newblock {Web-Based Information Retrieval using Agent and Ontology}.
\newblock In {\em Proceedings of the First Asia-Pacific Conference on Web
  Intelligence {\rm (}WI 2001\/{\rm )}}, {{N. Zhong et al.}}, Ed. LNCS,
  subseries LNAI, vol. 2198. Springer, 384--388.

\bibitem[\protect\citeauthoryear{Singh, Cannata, Huhns, Jacobs, Ksiezyk, Ong,
  Sheth, Tomlinson, and Woelk}{Singh et~al\mbox{.}}{1997}]{cann-etal-97}
{\sc Singh, M.}, {\sc Cannata, P.}, {\sc Huhns, M.}, {\sc Jacobs, N.}, {\sc
  Ksiezyk, T.}, {\sc Ong, K.}, {\sc Sheth, A.}, {\sc Tomlinson, C.}, {\sc and}
  {\sc Woelk, D.} 1997.
\newblock {The Carnot Heterogeneous Database Project: Implemented
  Applications}.
\newblock {\em {Distributed and Parallel Databases}\/}~{\em 5,\/}~2, 207--225.

\bibitem[\protect\citeauthoryear{Subrahmanian, Bonatti, Dix, Eiter, Kraus,
  Ozcan, and Ross}{Subrahmanian et~al\mbox{.}}{2000}]{subr-etal-00}
{\sc Subrahmanian, V.}, {\sc Bonatti, P.}, {\sc Dix, J.}, {\sc Eiter, T.}, {\sc
  Kraus, S.}, {\sc Ozcan, F.}, {\sc and} {\sc Ross, R.} 2000.
\newblock {\em {Heterogeneous Agent Systems: Theory and Implementation}}.
\newblock MIT Press.

\bibitem[\protect\citeauthoryear{Swift}{Swift}{2004}]{Swift2004}
{\sc Swift, T.} 2004.
\newblock {Deduction in Ontologies via {ASP}}.
\newblock In {\em Proceedings of the Seventh International Conference on Logic
  Programming and Nonmonotonic Reasoning {\rm (}LPNMR 2004{\rm )}},
  {I.~Niemel{\"a}} {and} {V.~Lifschitz}, Eds. LNCS, subseries LNAI, vol. 2923.
  Springer, 275--288.

\bibitem[\protect\citeauthoryear{{Van Nieuwenborgh} and Vermeir}{{Van
  Nieuwenborgh} and Vermeir}{2002}]{vann-verm-2002}
{\sc {Van Nieuwenborgh}, D.} {\sc and} {\sc Vermeir, D.} 2002.
\newblock {Preferred Answer Sets of Ordered Logic Programs}.
\newblock In {\em {Proceedings of the Eighth European Conference on Logics in
  Artificial Intelligence {\rm (}JELIA 2002\/{\rm )}}}, {S.~Flesca},
  {S.~Greco}, {G.~Ianni}, {and} {N.~Leone}, Eds. LNCS, subseries LNAI, vol.
  2424. 432--443.

\bibitem[\protect\citeauthoryear{Wendlandt and Driscoll}{Wendlandt and
  Driscoll}{1991}]{wendlandt-driscoll-91}
{\sc Wendlandt, E.~B.} {\sc and} {\sc Driscoll, J.~R.} 1991.
\newblock {Incorporating a Semantic Analysis into a Document Retrieval
  Strategy}.
\newblock In {\em {Proceedings of the Fourteenth Annual International ACM SIGIR
  Conference on Research and Development in Information Retrieval}},
  {A.~Bookstein}, {Y.~Chiaramella}, {G.~Salton}, {and} {V.~V. Raghavan}, Eds.
  ACM Press, 270--279.

\bibitem[\protect\citeauthoryear{Wiederhold}{Wiederhold}{1993}]{wied-93}
{\sc Wiederhold, G.} 1993.
\newblock {Intelligent Integration of Information}.
\newblock In {\em Proceedings of the ACM SIGMOD Conference on Management of
  Data {\rm (}SIGMOD '93\/{\rm )}}. 434--437.

\end{thebibliography}


\begin{thebibliography}{}

\end{thebibliography}

\else

\fi

\end{document}